\documentclass[letterpaper]{article} % DO NOT CHANGE THIS
\usepackage{aaai2026}  % DO NOT CHANGE THIS
\usepackage{times}  % DO NOT CHANGE THIS
\usepackage{helvet}  % DO NOT CHANGE THIS
\usepackage{courier}  % DO NOT CHANGE THIS
\usepackage[hyphens]{url}  % DO NOT CHANGE THIS
\usepackage{graphicx} % DO NOT CHANGE THIS
\usepackage{natbib}  % DO NOT CHANGE THIS AND DO NOT ADD ANY OPTIONS TO IT
\usepackage{caption} % DO NOT CHANGE THIS AND DO NOT ADD ANY OPTIONS TO IT
\usepackage{algorithm}
\usepackage{algorithmic}
\usepackage{xcolor}
\usepackage{newfloat}
\usepackage{listings}
\DeclareCaptionStyle{ruled}{labelfont=normalfont,labelsep=colon,strut=off} % DO NOT CHANGE THIS
\floatstyle{ruled}
\newfloat{listing}{tb}{lst}{}
\floatname{listing}{Listing}
\usepackage{multirow}
\usepackage{booktabs}
\usepackage{microtype}

\usepackage{inconsolata}

\usepackage{graphicx}
\usepackage{subcaption}
\usepackage{tabularx}
\usepackage{makecell}
\usepackage{colortbl}         % colors of the table

\usepackage{amsmath}

\title{How Much Do Large Language Model Cheat On Evaluation? Benchmarking Overestimation Under The One-Time-Pad-Based Framework}

\author{
{Zi Liang\textsuperscript{\rm 1}}, {Liantong Yu\textsuperscript{\rm 1}},
{Shiyu Zhang\textsuperscript{\rm 1}}, {Qingqing Ye\textsuperscript{\rm 1}},
{Haibo Hu\textsuperscript{\rm 1,2}}\thanks{Corresponding author.}
}

\affiliations {
\textsuperscript{\rm 1}{The Hong Kong Polytechnic University}\\
\textsuperscript{\rm 2}{PolyU Research Centre for Privacy and Security Technologies in Future Smart Systems}\\
{\{zi1415926.liang,liantong2001.yu,shiyu187.zhang\}@connect.polyu.hk,\\
\{qqing.ye,haibo.hu\}@polyu.edu.hk}
}

\begin{document}

\maketitle

\begin{abstract}
Overestimation in evaluating large language models (LLMs) has become an increasing concern.
Due to the contamination of public benchmarks or imbalanced model training, LLMs may achieve unreal evaluation results on public benchmarks, either intentionally or unintentionally, which leads to unfair comparisons among LLMs and undermines their realistic capability assessments.
Existing benchmarks attempt to address these issues by keeping test
cases permanently secret, mitigating contamination through human evaluation, or repeatedly collecting and constructing new samples. However, these approaches fail to ensure reproducibility, transparency, and high efficiency simultaneously.
Moreover, the extent of overestimation in current LLMs remains unquantified.
To address these issues, we propose ArxivRoll, a dynamic evaluation framework inspired by one-time pad encryption in cryptography. ArxivRoll comprises two key components: \emph{i) SCP (Sequencing, Cloze, and Prediction)}, an automated generator for private test cases, and \emph{ii) Rugged Scores (RS)}, metrics that measure the proportion of public benchmark contamination and training bias. Leveraging SCP, ArxivRoll constructs a new benchmark every six months using recent articles from ArXiv and employs them for one-time evaluations of LLM performance.
Extensive experiments demonstrate the high quality of our benchmark, and we provide a systematic evaluation of current LLMs.
\end{abstract}

\section{Introduction}

With the rapid development of large language models (LLMs), their
evaluation has attracted growing attention. Numerous challenging and
widely recognized benchmarks~\cite{mmlu,gsm8k,livebench,arena,swebench,tbench} have been introduced to
assess the knowledge and reasoning capabilities of these models. As a
result, these evaluations have become the primary, and often the only,
standard for comparing the performance of large language models.

Despite their effectiveness, recent research~\cite{overestimation1,
  contamination3,contamination5} increasingly highlights the
shortcomings of current evaluation mechanisms, arguing that the
capabilities of LLMs are often universally \emph{overestimated}. This
occurs mainly due to evaluation leakage, where test samples,
benchmark details or formatting information can be exploited to game
the benchmark. Consequently, it may inflate the
perceived performance of a model, resulting in unreliable evaluations
and unfair comparisons among LLMs. Malicious developers
could further fool benchmarks by incorporating test
samples or benchmark-specific information during training or
fine-tuning. For instance, a previous study~\cite{attack-contamination}
demonstrated that a 13-billion-parameter Llama model can easily achieve
results comparable to GPT-4 on benchmarks like MMLU~\cite{mmlu} through
post-processing-based fine-tuning. Additionally, popular open-source LLMs
such as Llama-4 and Qwen-2.5 have been reported~\cite{llama4-overshadow,rom} to experience 
test-data-contaminated training. Such intentional or unintentional
cheating behaviors distort the true capabilities of LLMs, misleading
subsequent training procedures and corresponding discoveries~\cite{rom}.

Specifically, there are two main types of abuse involving evaluation
benchmarks. The first is \emph{data
  contamination}~\cite{contamination1,contamination2,contamination3,contamination4,contamination5},
where test cases from the benchmarks are included in the training set
of large language models, enabling them to become familiar with or
even memorize these samples, resulting in artificially improved
performance. The second is \emph{biased overtraining}, where models
are claimed to be ``comprehensive'' but actually prioritize improving
their performance in the evaluated domain at the expense of
undertraining in other areas. Both scenarios significantly undermine
the effectiveness, fairness, and reliability of evaluation results.

Unfortunately, existing benchmarks designed to mitigate cheating behaviors have notable
limitations. Private benchmarks maintained by trusted third-party
platforms, such as SEAL\footnote{\url{https://scale.com/leaderboard}},
and Arena-like benchmarks~\cite{arena,olympicarena,arena-hard,arena-hard-auto}, such as Chatbot
Arena~\cite{arena}, lack transparency and reproducibility in their
evaluation processes. Symbolic formatting benchmarks for specific
domains~\cite{dyval,darg,dempa}, such as GSM-Symbolic~\cite{gsm-symbolic} and
LiveBench~\cite{livebench}, are restricted to
narrow fields and therefore fail to provide a comprehensive evaluation
of LLMs. Furthermore, the above benchmarks primarily
focus on assessing the realistic abilities of LLMs, without offering a
clear \emph{quantification} of the extent of overestimation. As a
result, a stable, transparent, reproducible, and human-effort-free
framework and benchmark for evaluating LLMs has yet to be developed.

To address these issues, we propose ArxivRoll, a robust and dynamic
framework designed to evaluate both the realistic performance and the
overestimation of large language models. ArxivRoll consists of two key
components: \emph{1)} SCP (Sequencing, Cloze, and Prediction), a novel
method that automatically generates test cases from newly published
articles on ArXiv to construct private benchmarks; and \emph{2)}
Rugged Scores (RS), indicators that quantify the performance
difference between public and private benchmarks, providing a clear
measure of overestimation. Inspired by the security guarantee of \emph{One-Time
Pad}~\cite{otp,otp2} in cryptography, which uses a unique secret key
for each use, ArxivRoll divides benchmarks into public benchmarks
(existing ones) and private ArxivRollBenches (generated by SCP), and regard the
private benchmarks as the \textbf{one-time-used} secrets to mitigate the
overestimation. After evaluation, the private benchmarks are publicly
released to ensure reproducibility of evaluation but are marked as expired to
prevent future use or reference.
Extensive meta-evaluations on ArxivRollBench demonstrate that SCP consistently produces high-quality
test samples. Besides, the private benchmarks exhibit a strong correlation
with existing private yet non-transparent benchmarks, confirming their reliability and relevance.

Our contributions are summarized as follows:

\begin{itemize}
\item We devise a novel private benchmark construction strategy, SCP (Sequencing, Cloze, and Prediction) based on Arxiv, which automatically generates high-quality, challenging, and fresh test cases tailored for assessing the capabilities of LLMs. Extensive experiments have proved the high quality of our generated private benchmarks.

\item We design \emph{rugged scores (RS)} to quantify the proportion
  of cheating behavior in a given LLM when tasked with specific
  challenges. To the best of our knowledge, this is the first study to
  measure the proportion of overestimation and the biased
  overtraining.

\item Leveraging RS and SCP, we present a novel and one-time-pad-based
  evaluation framework, namely ArxivRoll. This framework not only
  evaluates the performance of current LLMs but also considers their
  overestimation situations. Through ArxivRoll, we conduct a
  systematic evaluation and establish a leaderboard\footnote{\url{https://arxivroll.moreoverai.com}} for popular LLMs, providing a comprehensive evaluation of their capabilities.
\end{itemize}

The source code is available at \url{https://github.com/liangzid/ArxivRoll/}.

\begin{figure*}[t]
  \centering
  \includegraphics[width=0.99\linewidth]
    {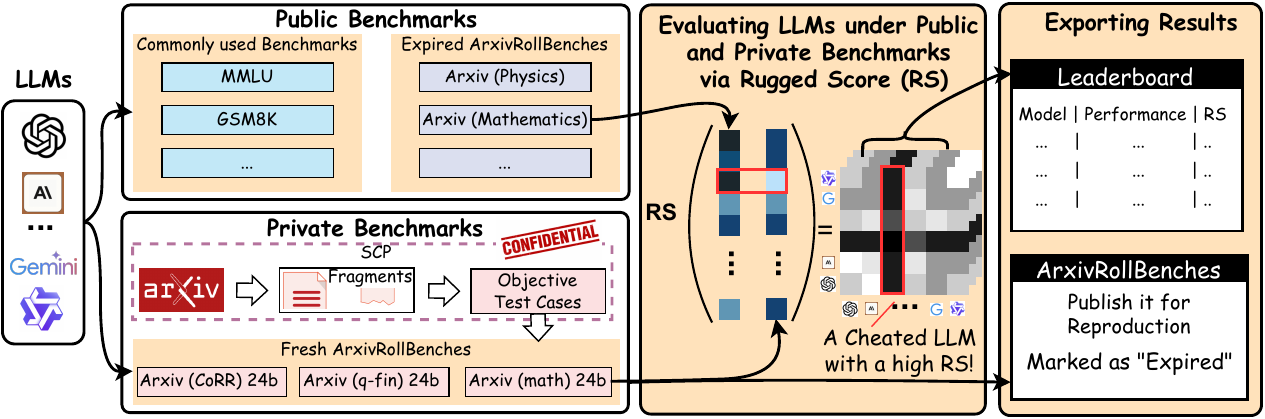}
\caption{\textbf{Framework of ArxivRoll}, which
  categorizes benchmarks into two distinct groups: public benchmarks
  and private benchmarks (i.e., ArxivRollBench). These benchmarks are utilized to estimate
  both the overestimation proportion and performance of large language
  models (LLMs). Notably, ArxivRoll represents a \emph{dynamic}
  benchmarking system, where private benchmarks are utilized
  exclusively once and then expire for subsequent evaluation stages,
  ensuring freshness and reliability in each assessment.}
    \label{fig:robench}
\end{figure*}

%%% Local Variables:
%%% mode: latex
%%% TeX-master: "main"
%%% End:

\section{ArxivRoll}

\begin{figure*}[t]
  \centering
  \includegraphics[width=0.95\linewidth]
    {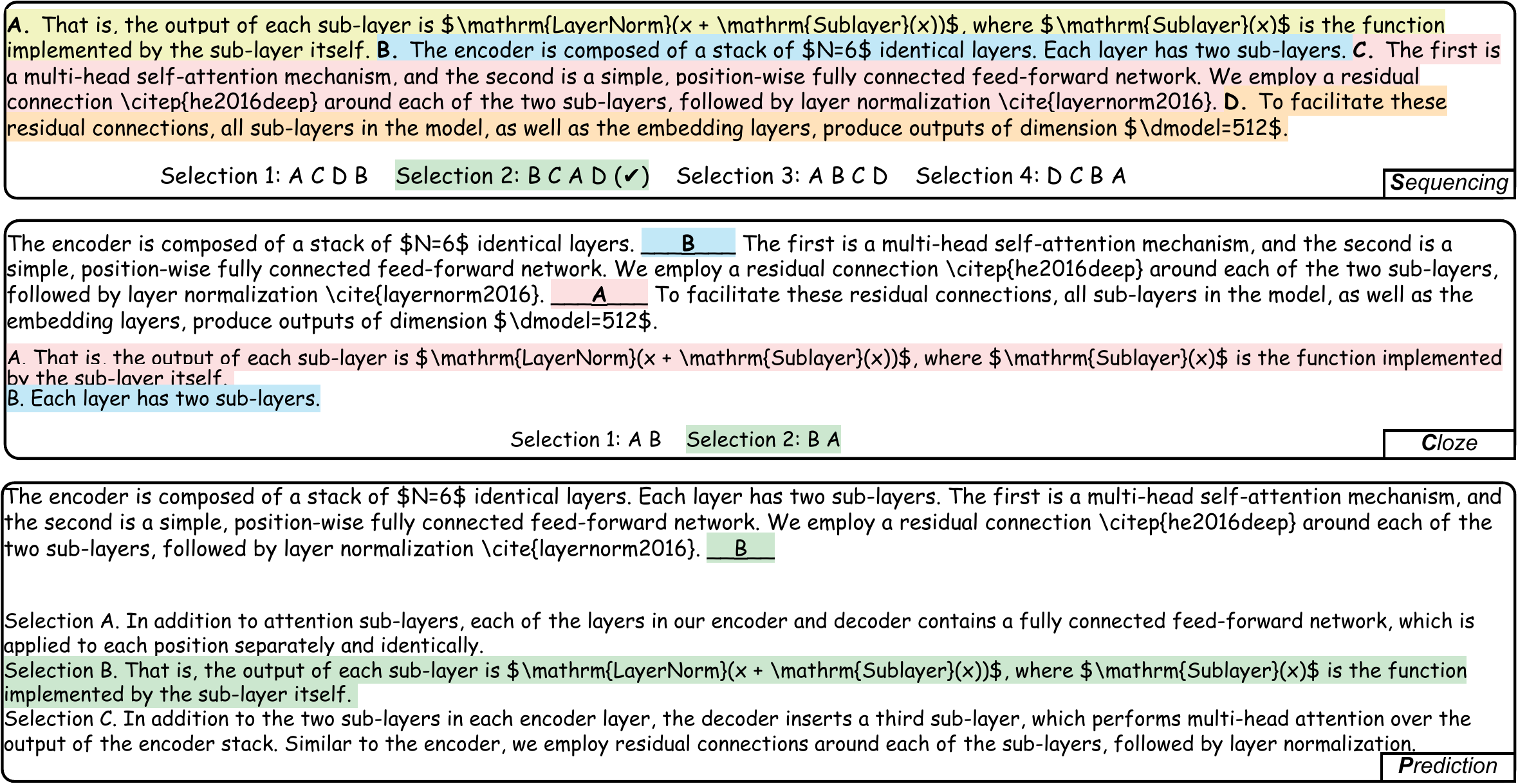}
\caption{\textbf{An illustrative example of symbolic formatting for
    test samples}, encompassing three formats: sequencing, cloze, and
  prediction (SCP).
  % We have reformatted SCP into a four-candidate
  % selection task, as detailed in Figure \ref{fig:details}.
}
    \label{fig:scp}
\end{figure*}

In this section, we will introduce the implementation of ArxivRoll and
explain why it can address limitations that existed in previous benchmarks.
Specifically, we first introduce our dynamic evaluation framework in
Section \ref{sec:overview}, and then respectively detail our test cases generator
technique as well as the metrics in
Section \ref{sec:scp} and \ref{sec:rs}.

\subsection{Overview}\label{sec:overview}

As illustrated in Figure \ref{fig:robench}, ArxivRoll encompasses two
categories of benchmarks: \emph{public} and \emph{private}. Public
benchmarks refer to those publicly available on the Internet, which
may be susceptible to contamination or hacking during the pre-training
of LLMs. Conversely, private benchmarks, namely ArxivRollBench, are created by ArxivRoll and
remain confidential until the evaluation period, thereby ensuring that
they are unseen by LLMs. In addition to assessing performance, ArxivRoll
also computes two key values:
\begin{itemize}
\item The difference in performance for an LLM between the public and private benchmarks within the same domain (e.g., mathematics reasoning). This metric reflects the proportion of contamination in the model's performance on public benchmarks;
\item The difference in performance for an LLM among various private benchmarks. This metric indicates the degree of biased overtraining in the model.
\end{itemize}
We propose the rugged score to quantify these two differences, as shown in
Section \ref{sec:rs}.

After evaluation, we can compile the performances and rugged scores for all LLMs into a leaderboard and make our constructed private benchmarks publicly available on the Internet to ensure the reproducibility and transparency of the evaluation process. These benchmarks will be regarded as public benchmarks in future evaluations.

This outlines the entire procedure of ArxivRoll for one evaluation
period. As a dynamic benchmark, it will regularly publish new
evaluations (e.g., every six months). For each evaluation period, as
shown above, ArxivRoll will incorporate new private benchmarks to
minimize the impact of contamination and biased overtraining, as shown
in Figure \ref{fig:robench}.

Such a framework faces two primary challenges:

\begin{itemize}
\item How do we create confidential benchmarks for each evaluation stage that are both challenging and representative of the domain, while ensuring they remain unseen by LLMs until the evaluation period?
\item How do we formally measure the two differences to provide a rigorous and interpretable evaluation?
\end{itemize}

We will address them in the following two parts.

\subsection{Sequencing, Cloze, and Prediction (SCP): Producing Test Cases}\label{sec:scp}

From Section \ref{sec:overview}, we can discern that private
benchmarks must meet the following four criteria: \emph{i)
  confidentiality}, ensuring that LLMs do not
encounter the test cases during their training process; \emph{ii)
  difficulty}, where test cases should not adhere to fixed patterns but
remain flexible and complex in content, preventing LLMs from easily
solving them through lexical comprehension alone; \emph{iii)
  objectivity}, to minimize the impact of subjective evaluation metrics;
and \emph{iv) comprehensiveness}, for encompassing a wide range of fields or sub-fields rather than being
confined to a narrow task.
Moreover, given the need to introduce new private benchmarks for
each evaluation stage, we aspire to construct test cases
\emph{automatically}.

To this end, we have chosen ArXiv\footnote{\url{https://arxiv.org/}}, a preprint platform, as the
source for our test cases. The timely papers published on ArXiv
fulfill the criteria of confidentiality and difficulty, as they
represent the latest research advancements in their domains
and are often unprecedented in academia. Consequently, these papers
are conceptually unseen by LLMs to date, making them suitable for our
benchmark construction.

Despite the potential of designing test samples based on ArXiv
articles, the process remains time-consuming and challenging,
necessitating expert-level annotators. To tackle this issue, we adopt
the concept of symbolic formatting and propose
an automated test sample generation strategy named SCP. SCP is
inspired by educational quizzes \cite{cloze1,cloze2,cloze3} and
Gestalt psychology \cite{gestalt1,gestalt2},
which comprises three objective tasks:

\begin{itemize}
\item \textbf{Sequencing}: Given a text fragment extracted from an
  article, the input of a test case consists of shuffled sentences from this
  fragment. LLMs are tasked with selecting the correct order of these
  sentences.
\item \textbf{Cloze}: In this task, a text fragment is provided with
  certain sentences masked. LLMs are required to select the
  appropriate sentences to fill in these gaps.
\item \textbf{Prediction}: Given a text fragment, a correct subsequent
  sequence, and three distractors, LLMs must identify and select the
  correct next sequence.
\end{itemize}

Formally, for an article, we first sample a text fragment containing
$N_{p}$ paragraphs, filtering out texts heavy with mathematical
formulas and tables. Then, we utilize one of the strategies
within SCP to generate the test case. Figure \ref{fig:scp} depicts the construction
process of SCP.

% SCP shares similarities with certain pre-training tasks in terms of
% its structure. For instance, the Cloze task mirrors the format of
% BERT's Masked Language Modeling (MLM)~\cite{bert}, Electra's discrimination
% task~\cite{electra}, and T5's recovery mechanism~\cite{t5}, while the
% Prediction task aligns with BERT's Next Sentence Prediction (NSP) and
% GPT's Next Token Generation (NTG). However, SCP distinguishes itself
% by focusing on evaluating reasoning abilities at the "concept" level,
% rather than the "token" or "lexical" levels. This shift in focus aims
% to assess the deeper comprehension and inferential skills of large
% language models (LLMs). In Appendix \ref{sec:detail-robench}, we provide a more nuanced
% analysis of the benchmarks generated using SCP.

\subsection{RS: Quantifying Overestimation}\label{sec:rs}

Given both public and private benchmarks, another challenge arises in
assessing the reliability of performance evaluations conducted on
public benchmarks. Intuitively, within the same domain, if an
LLM demonstrates significantly higher performance on
a public benchmark compared to a private one, we may conclude that the
public benchmark is being "fooled" by the LLM. To quantify this
discrepancy, we introduce a novel metric called the \textbf{rugged score (RS)}. This
metric measures the degree of "ruggedness" in performance between
public and private benchmarks.

Formally, given $N_{p}$ public-private benchmark pairs
$\mathcal{T}=\{(T_{p}^{i},T_{c}^{i})\}_{i=1,2,...,N_{p}}$ in which
the public benchmark $T_{p}^{i}$ and the private benchmark $T_{c}^{i}$
comes from the same domain, and given $N_{p}'$ unmatchable public
benchmarks $\mathcal{T}_{p}=\{T_{p}^{j}\}_{j=1,2,...,N_{p}'}$ and
$N_{c}$ unmatchable private benchmarks $\mathcal{T}_{c}=\{T_{c}^{k}\}_{k=1,2,...,N_{c}}$, we can
define the rugged score of a model $m$ as:
\begin{equation}
\label{eq:rs}\small
\begin{aligned}
&\mathbf{RS_I}(m,\mathcal{T},\mathcal{T}_{p},\mathcal{T}_{c})\\&=\frac{2}{N_{p}}\sum_{i}^{N_{p}}{\left[\frac{\mathcal{M}(m,T_{p}^{i})-\mathcal{M}(m,T_{c}^{i})}{\mathcal{M}(m,T_{p}^{i})+\mathcal{M}(m,T_{c}^{i})}\right]}+2\times\\&\left[\frac{\frac{1}{N_{p}'}\sum_{j}^{N_{p}'}{\mathcal{M}}(m,T_{p}^{j})-\frac{1}{N_{c}}\sum_{k}^{N_{c}}{\mathcal{M}}(m,T_{c}^{k})}{\frac{1}{N_{p}'}\sum_{j}^{N_{p}'}{\mathcal{M}}(m,T_{p}^{j})+\frac{1}{N_{c}}\sum_{k}^{N_{c}}{\mathcal{M}}(m,T_{c}^{k})}\right],
\end{aligned}
\end{equation}
where $\mathcal{M}(m,T)$ denotes the performance evaluation metric for the
model $m$ on task $T$. It can either be an absolute metric such as
the accuracy, or a relative metric like the rank of $m$
among all evaluated models $M$.

In intuition, the higher the $\mathbf{RS_I}$, the rugger the
$m$, demonstrating that the evaluated results of $m$ on
public benchmarks $\{T_{p}^{i}\}_{N_{p}}\cup \{T_{p}^{j}\}_{N_{p}'}$
may be less reliable, and model $m$ may be overfitted to the specific
characteristics of them.

Unfortunately, $\mathbf{RS_I}$ is not a \emph{normalized} metric and is
unavoidably coupled with models and benchmarks used for
evaluation. This means that $\mathbf{RS_I}$ obtained for different sets of models $M$ on
different benchmark triples
$(\mathcal{T},\mathcal{T}_{p},\mathcal{T}_{c})$
are \emph{incomparable}, and we can \textbf{only} decouple one factor between $M$ and
benchmarks triples from $\mathbf{RS_I}$.
Specifically, $\mathbf{RS_I}$ becomes
\emph{model-independent} when an absolute metric is adopted as $\mathcal{M}$,
allowing the free addition of new models under
the same triple $(\mathcal{T},\mathcal{T}_{p},\mathcal{T}_{c})$
without affecting the score's comparability. Conversely, it becomes
\emph{benchmark-independent} when a 
relative metric is used, meaning that it is comparable across
different evaluation periods for the same model set $M$.
In our evaluation, we will use both types of rugged scores.

To investigate the proportion of unbalanced overtraining on LLMs,
we propose $\mathbf{RS_{II}}$, which
can be measured by the standard variance on private benchmarks, i.e.,

\begin{equation}
\label{eq:rs2}
\mathbf{RS_{II}}=\sqrt{\sum_{T_{c}\sim \{\mathcal{T}_{c}\cup\mathcal{T}_{c}^{p}\}}\left[\mathcal{M}(m,T_{c})-\bar{\mathcal{M}}\right]^{2}},
\end{equation}
where $\mathcal{T}_{c}^{p}=\{T_{c}^{i},|i=1,2,...,N_{p}\}$ represents the set
of private benchmarks in $\mathcal{T}$, $\bar{\mathcal{M}}=\sum_{T_{c}\sim
  \{\mathcal{T}_{c}\cup\mathcal{T}_{c}^{p}\}}\mathcal{M}(m,T_{c})$ is the
average performance on private benchmarks. We also propose a normalized version:
\begin{equation*}
\label{eq:rs2}
\mathbf{RS_{II}^{N}}=\mathbf{RS_{II}}/\bar{\mathcal{M}}.
\end{equation*}

%%% Local Variables:
%%% mode: latex
%%% TeX-master: "main"
%%% End:

\section{Meta Evaluation}

In this section, we conduct a meta-evaluation of ArxivRoll. Specifically, Section \ref{sec:stable-robench} examines and assesses the quality of the generated test cases, while Section \ref{sec:corr-robench} investigates the potential correlation between ArxivRollBench's evaluation outcomes and those from other private benchmarks.

\subsection{The Generation Of SCP Is Stable}\label{sec:stable-robench}

\begin{figure}[t]
  \centering
  \includegraphics[width=0.99\linewidth]
    {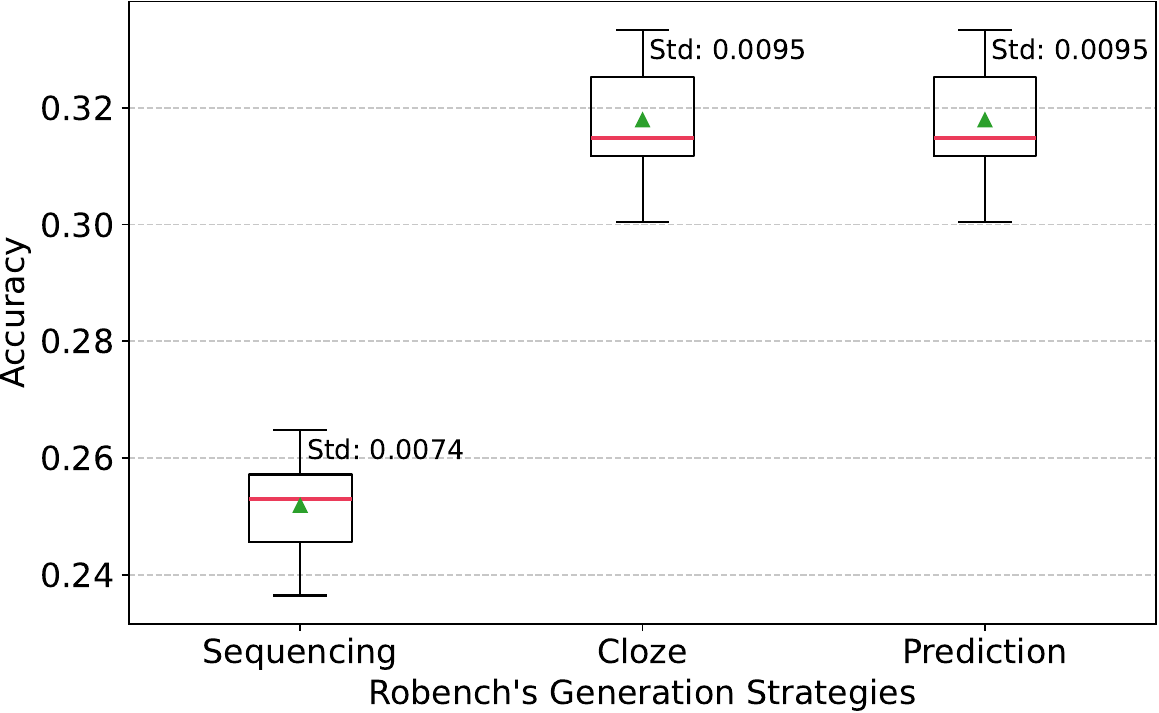}
\caption{Performance of Llama3 (8B) across 32-time-generated
  ArxivRollBench benchmarks. The benchmarks were generated 32 times from
  the same raw article set using SCP. The small standard variance in
  evaluation results indicates that SCP produces stable test cases.}
    \label{fig:stability}
\end{figure}

Our benchmark construction strategy (SCP) relies heavily on randomness, raising whether the evaluation results adequately reflect an LLM's understanding of the articles. To address this, we examine if the performance of an LLM varies significantly when evaluated on the same set of raw articles but generated with different random seeds.

Specifically, we repeated the generation process for ArxivRollBench2024b-CS 32 times using different seeds and collected the evaluation results of the Llama3-8B model. We calculated the performance variations across our three test case generation strategies: sequencing, cloze, and prediction, as illustrated in Figure \ref{fig:stability}.

Although the accuracy variations across all three benchmarks appear noticeable ($\sim 2.5$ points), the standard deviation of the evaluation results is minimal, remaining below 1 point. This demonstrates that our generation strategy is reliable and that the evaluation results are consistently reproducible, even with benchmarks generated under varying random seeds.

\subsection{ArxivRollBench Exhibits High Correlations With Popular Benchmarks}\label{sec:corr-robench}

In this section, we address the second concern of ArxivRoll: whether
the evaluation results from ArxivRollBench meaningfully reflect the
knowledge and reasoning abilities of LLMs within these domains. To
explore this, we compute the correlation between the performance
rankings of ArxivRollBench2024b's private benchmarks and those of widely used
benchmarks which are relatively harder to fool.

As a reference, we select ChatbotArena~\cite{arena}, a crowdsourced
and voting-based benchmark. Despite its limitations in
interpretability, transparency, and reproducibility, ChatbotArena
remains one of the most widely regarded benchmarks for LLM performance evaluation. We employ three standard correlation metrics: Pearson's coefficient, Spearman's rank correlation, and Kendall's rank correlation, all of which are commonly used to assess linear and rank-based relationships.

We first compute the correlations between our benchmark construction strategy, SCP, and the reference benchmark. Additionally, we analyze the internal correlations among the three test case generation strategies of SCP. The results of these analyses are presented in Table \ref{tab:corr}.

% \usepackage{graphicx}
\begin{table}[t]
\centering
\resizebox{\linewidth}{!}{%
\begin{tabular}{c|rrr}
\Xhline{1.5pt}
\multicolumn{1}{c|}{Benchmarks} & \multicolumn{1}{c}{Spear. Corr.} & \multicolumn{1}{c}{Pearson Co.} & \multicolumn{1}{c}{Kendall Corr.} \\ \hline
A.R.Bench (S) - ChatbotArena & 0.76 & 0.71 & 0.6 \\
A.R.Bench (C) - ChatbotArena & 0.61 & 0.51 & 0.55 \\
A.R.Bench (P) - ChatbotArena & 0.73 & 0.69 & 0.55 \\ \hline
A.R.Bench (S) - A.R.Bench (C) & 0.86 & 0.92 & 0.77 \\
A.R.Bench (S) - A.R.Bench (P) & 0.86 & 0.86 & 0.69 \\
A.R.Bench (C) - A.R.Bench (P) & 0.86 & 0.86 & 0.69 \\ \Xhline{1.5pt}
\end{tabular}%
}
\caption{Correlation Experiments among ArxivRollBench (A.R.Bench) and ChatbotArena, where
  Spear. Corr., Pearson Co., and Kendall Corr. denote the Spearman Correlation, Pearson Coefficient, and Kendall Correlation, respectively.}
\label{tab:corr}
\end{table}

As shown in Table \ref{tab:corr}, ArxivRollBench constructed with the
S (equencing) and P (rediction) strategies achieves up to a 0.70
Spearman correlation with ChatbotArena, while ArxivRollBench with the
C (loze) strategy also exhibits a notable correlation with the reference benchmark. This demonstrates that ArxivRollBench's private benchmarks effectively capture the realistic capabilities of LLMs. Moreover, the strong correlations among the three SCP construction strategies indicate their internal consistency. The Pearson coefficients further suggest that their evaluation results exhibit linear relationships, reinforcing the robustness of our proposed approach.

Having established the utility of ArxivRollBench, we proceed to provide an in-depth analysis of the overestimation behavior of current LLMs under ArxivRoll in Section \ref{sec:eva}.

% \usepackage{graphicx}
\begin{table*}[h]
\centering
\resizebox{0.90\textwidth}{!}{%
\begin{tabular}{c|rrrrrrrrrr}
\Xhline{1.5pt}
Model name & \multicolumn{8}{c}{ArxivRollBench-2024b (S)} &  &  \\
\hline
 & CS & Q-Fin. & Math & Phy. & Stat. & Bio. & Econ. & EESS &  &  \\
 \hline
GPT-J-6B & 10.3 $\pm$ 0.6 & 12.2 $\pm$ 1.1 & 8.0 $\pm$ 0.6 & 11.7 $\pm$ 0.7 & 9.7 $\pm$ 0.5 & 12.0 $\pm$ 0.8 & 9.7 $\pm$ 1.0 & 12.5 $\pm$ 0.5 &  &  \\
Phi-1 & 5.6 $\pm$ 0.4 & 6.9 $\pm$ 0.9 & 7.2 $\pm$ 0.6 & 7.5 $\pm$ 0.6 & 7.6 $\pm$ 0.4 & 5.1 $\pm$ 0.6 & 6.8 $\pm$ 0.9 & 6.3 $\pm$ 0.4 &  &  \\
Phi-1.5 & 22.7 $\pm$ 0.8 & 20.9 $\pm$ 1.4 & 25.2 $\pm$ 0.9 & 23.9 $\pm$ 1.0 & 22.5 $\pm$ 0.7 & 23.6 $\pm$ 1.1 & 24.5 $\pm$ 1.5 & 21.4 $\pm$ 0.7 &  &  \\
Phi-2 & 23.2 $\pm$ 0.8 & 22.8 $\pm$ 1.4 & 24.8 $\pm$ 0.9 & 24.4 $\pm$ 1.0 & 23.6 $\pm$ 0.7 & 24.2 $\pm$ 1.1 & 24.8 $\pm$ 1.5 & 23.1 $\pm$ 0.7 &  &  \\
Phi-3-Mini-4K-Instruct & 6.3 $\pm$ 0.4 & 4.8 $\pm$ 0.7 & 5.3 $\pm$ 0.5 & 3.4 $\pm$ 0.4 & 6.7 $\pm$ 0.4 & 5.3 $\pm$ 0.6 & 5.1 $\pm$ 0.7 & 6.4 $\pm$ 0.4 &  &  \\
Phi-3.5-Mini-Instruct & 19.8 $\pm$ 0.7 & 20.3 $\pm$ 1.4 & 19.2 $\pm$ 0.9 & 17.9 $\pm$ 0.9 & 19.1 $\pm$ 0.7 & 18.6 $\pm$ 1.0 & 19.3 $\pm$ 1.3 & 19.4 $\pm$ 0.6 &  &  \\
Phi4-Reasoning & 2.0 $\pm$ 0.3 & 2.2 $\pm$ 0.5 & 3.4 $\pm$ 0.4 & 1.3 $\pm$ 0.3 & 1.9 $\pm$ 0.2 & 1.9 $\pm$ 0.4 & 2.5 $\pm$ 0.5 & 1.4 $\pm$ 0.2 &  &  \\ 
Phi4-Reasoning-Plus & 11.1 $\pm$ 0.6 & 13.1 $\pm$ 1.2 & 10.9 $\pm$ 0.7 & 9.0 $\pm$ 0.6 & 9.8 $\pm$ 0.5 & 10.6 $\pm$ 0.8 & 13.0 $\pm$ 1.1 & 9.2 $\pm$ 0.5 &  &  \\ 
Qwen2-7B-Instruct & 26.6 $\pm$ 0.8 & 27.9 $\pm$ 1.5 & 25.7 $\pm$ 1.0 & 26.4 $\pm$ 1.0 & 28.3 $\pm$ 0.8 & 27.0 $\pm$ 1.2 & 27.9 $\pm$ 1.5 & 27.6 $\pm$ 0.7 &  &  \\
Qwen2.5-7B & 23.7 $\pm$ 0.8 & 24.8 $\pm$ 1.5 & 22.1 $\pm$ 0.9 & 23.9 $\pm$ 1.0 & 23.4 $\pm$ 0.7 & 26.8 $\pm$ 1.1 & 25.3 $\pm$ 1.5 & 24.3 $\pm$ 0.7 &  &  \\ 
Qwen2.5-7B-Instruct & 27.6 $\pm$ 0.8 & 26.5 $\pm$ 1.5 & 28.6 $\pm$ 1.0 & 28.3 $\pm$ 1.0 & 26.7 $\pm$ 0.7 & 28.2 $\pm$ 1.2 & 28.3 $\pm$ 1.5 & 27.4 $\pm$ 0.7 &  &  \\
Qwen2.5-Math-7B & 16.7 $\pm$ 0.7 & 18.4 $\pm$ 1.3 & 18.8 $\pm$ 0.9 & 17.7 $\pm$ 0.9 & 17.5 $\pm$ 0.6 & 17.0 $\pm$ 1.0 & 17.2 $\pm$ 1.3 & 15.6 $\pm$ 0.6 &  &  \\ 
Qwen2.5-Math-7B-Instruct & 5.0 $\pm$ 0.4 & 4.7 $\pm$ 0.7& 3.7 $\pm$ 0.4 & 3.6 $\pm$ 0.4 & 6.7 $\pm$ 0.4 & 4.9 $\pm$ 0.6 & 7.4 $\pm$ 0.9 & 6.0 $\pm$ 0.4 &  &  \\ 
Qwen2.5-72B-Instruct & 20.5 $\pm$ 0.7 & 21.8 $\pm$ 1.4 & 17.8 $\pm$ 0.8 & 18.6 $\pm$ 0.9 & 18.8 $\pm$ 0.7 & 22.1 $\pm$ 1.1 & 18.3 $\pm$ 1.3 & 21.8 $\pm$ 0.7 &  &  \\
Qwen3-8B & 31.0 $\pm$ 0.9 & 31.3 $\pm$ 1.6 & 29.0 $\pm$ 1.0 & 28.7 $\pm$ 1.0 & 30.3 $\pm$ 0.8 & 28.5 $\pm$ 1.2 & 27.5 $\pm$ 1.5 & 29.2 $\pm$ 0.7 &  &  \\ 
Qwen3-14B & 4.7 $\pm$ 0.4 & 6.0 $\pm$ 0.8 & 6.3 $\pm$ 0.5 & 5.0 $\pm$ 0.5 & 5.4 $\pm$ 0.4 & 5.1 $\pm$ 0.6 & 5.1 $\pm$ 0.7 & 4.9 $\pm$ 0.4 &  &  \\ 
Qwen3-32B & 20.2 $\pm$ 0.7 & 22.2 $\pm$ 1.4 & 20.7 $\pm$ 0.9 & 17.8 $\pm$ 0.9 & 20.2 $\pm$ 0.7 & 19.9 $\pm$ 1.0 & 18.3 $\pm$ 1.3 & 20.1 $\pm$ 0.7 &  &  \\ 
Llama2-7B-Chat-HF & 7.5 $\pm$ 0.5 & 8.5$\pm$ 1.0 & 10.0 $\pm$ 0.7 & 6.3 $\pm$ 0.5 & 7.8 $\pm$ 0.5 & 7.3 $\pm$ 0.7 & 10.4 $\pm$ 1.0 & 6.8 $\pm$ 0.4 &  &  \\
Llama3-8B & 22.9 $\pm$ 0.8 & 22.8 $\pm$ 1.4 & 21.7 $\pm$ 0.9 & 23.0 $\pm$ 0.9 & 22.3 $\pm$ 0.7 & 23.6 $\pm$ 1.1 & 20.5 $\pm$ 1.4 & 21.4 $\pm$ 0.7 &  &  \\
Llama3.1-8B & 26.0 $\pm$ 0.8 & 24.2 $\pm$ 1.5 & 24.4 $\pm$ 0.9 & 25.3 $\pm$ 1.0 & 24.7 $\pm$ 0.7 & 25.3 $\pm$ 1.1 & 21.3 $\pm$ 1.4 & 23.0 $\pm$ 0.7 &  &  \\
Llama3.1-8B-Instruct & 28.5 $\pm$ 0.8 & 25.2 $\pm$ 1.5 & 28.6 $\pm$ 1.0 & 27.4 $\pm$ 1.0 & 26.8 $\pm$ 0.8 & 26.1 $\pm$ 1.1 & 24.9 $\pm$ 1.5 & 25.5 $\pm$ 0.7 &  &  \\
Llama3.1-70B-Instruct & 31.4 $\pm$ 0.9 & 34.0 $\pm$ 1.6 & 29.3 $\pm$ 1.0 & 30.9 $\pm$ 1.0 & 30.3 $\pm$ 0.8 & 33.7 $\pm$ 1.2 & 31.9 $\pm$ 1.6 & 32.2 $\pm$ 0.8 &  &  \\
Llama3.1-Nemotron-70B& 33.3 $\pm$ 0.9 &  35.8 $\pm$ 1.6 & 30.1 $\pm$ 1.0 & 32.8 $\pm$ 1.1 & 32.1 $\pm$ 0.8 & 34.4 $\pm$ 1.2 & 33.2 $\pm$ 1.6  & 34.4 $\pm$ 0.8 & &  \\ 
Llama3.2-1B & 24.0 $\pm$ 0.8 & 23.6 $\pm$ 1.5 & 25.8 $\pm$ 1.0 & 25.3 $\pm$ 1.0 & 23.8 $\pm$ 0.7 & 25.0 $\pm$ 1.1 & 26.2 $\pm$ 1.5 & 24.1 $\pm$ 0.7 &  &  \\
Llama3.2-3B & 23.1 $\pm$ 0.8 & 21.1 $\pm$ 1.4 & 19.2 $\pm$ 0.9 & 22.0 $\pm$ 0.9 & 21.6 $\pm$ 0.7 & 23.0 $\pm$ 1.1 & 24.3 $\pm$ 1.4 & 21.4 $\pm$ 0.7 &  &  \\ 
Llama3.3-70B-Instruct & 37.3 $\pm$ 0.9 & 39.0 $\pm$ 1.7 & 34.9 $\pm$ 1.0 & 36.4 $\pm$ 1.1 & 36.0 $\pm$ 0.8 & 37.7 $\pm$ 1.3 & 37.1 $\pm$ 1.6 & 37.4 $\pm$ 0.8 &  &  \\ 
Yi1.5-34B & 28.1 $\pm$ 0.8 & 28.1 $\pm$ 1.5 & 25.9 $\pm$ 1.0 & 26.5 $\pm$ 1.0 & 29.8 $\pm$ 0.8 & 27.1 $\pm$ 1.2 & 25.7 $\pm$ 1.5 & 27.9 $\pm$ 0.7 &  &  \\ 
Kimi-K2 & 35.7 $\pm$ 7.5 & 40.8 $\pm$ 7.1 & 50.0 $\pm$ 8.7 & 40.0 $\pm$ 7.4 & 44.4 $\pm$ 7.5 & 41.9 $\pm$ 7.6 & 41.7 $\pm$ 7.2 & 43.8 $\pm$ 7.2 &  &  \\ 
Deepseek-Chat-V3 & 45.2 $\pm$ 7.8 & 38.8 $\pm$ 7.0 & 50.0 $\pm$ 8.7 & 44.4 $\pm$ 7.5 & 42.2 $\pm$ 7.4 & 44.2 $\pm$ 7.7 & 41.7 $\pm$ 7.2 & 50.0 $\pm$ 7.3 &  &  \\ 
\hline
GPT-3.5-turbo  & 38.1 $\pm$ 7.6 & 28.6 $\pm$ 6.5 & 50.0 $\pm$ 8.7 & 20.0 $\pm$ 6.0 & 26.7 $\pm$ 6.7 & 34.9 $\pm$ 7.4 & 14.6 $\pm$ 5.1 & 31.3 $\pm$ 6.8 &  &  \\   
GPT-4 & 42.9 $\pm$ 7.7 & 42.9 $\pm$ 7.1 & 32.4 $\pm$ 8.1 & 37.8 $\pm$ 7.3 & 40.0 $\pm$ 7.4 & 34.9 $\pm$ 7.4 & 37.5 $\pm$ 7.1 & 41.7 $\pm$ 7.2 &  &  \\ 
GPT-4o & 42.9 $\pm$ 7.7 & 49.0 $\pm$ 7.2 & 35.3 $\pm$  8.3 & 31.1 $\pm$ 7.0 & 46.7 $\pm$ 7.5 & 41.9 $\pm$ 7.6 & 39.6 $\pm$ 7.1 & 41.7 $\pm$ 7.2 &  &  \\ 
Claude-3.5-Sonnet & 38.1 $\pm$ 7.6 & 36.7 $\pm$ 7.0 & 26.5 $\pm$ 7.7 & 37.8 $\pm$ 7.3 & 44.4 $\pm$ 7.5 & 37.2 $\pm$ 7.5 & 35.4 $\pm$ 7.0 & 43.8 $\pm$ 7.2 &  &  \\ 
Claude-3.7-Sonnet & 33.3 $\pm$ 7.4 & 40.8 $\pm$ 7.1 & 20.6 $\pm$ 7.0 & 37.8 $\pm$ 7.3 & 44.4 $\pm$ 7.5 & 30.2 $\pm$ 7.1 & 25.0 $\pm$ 6.3 & 37.5 $\pm$ 7.1 &  &  \\ 
Claude-4-Sonnet & 57.1 $\pm$ 7.7 & 51.0 $\pm$ 7.2 & 35.3 $\pm$ 8.3 & 31.1 $\pm$ 7.0 & 57.8 $\pm$ 7.4 & 41.9 $\pm$ 7.6 & 35.4 $\pm$ 7.0 & 37.5 $\pm$ 7.1 &  &  \\ 
Gemini-2.0-flash-001 & 40.5 $\pm$ 7.7 & 44.9 $\pm$ 7.2 & 41.2 $\pm$ 8.6 & 40.0 $\pm$ 7.4 & 37.8 $\pm$ 7.3 & 41.9 $\pm$ 7.6 & 45.8 $\pm$ 7.3 & 41.7 $\pm$ 7.2 &  &  \\ 
Gemini-2.5-flash & 40.5 $\pm$ 7.7 & 59.2 $\pm$ 7.1 & 35.3 $\pm$ 8.3 & 55.7 $\pm$ 7.5 & 60.0 $\pm$ 7.4 & 46.5 $\pm$ 7.7 & 47.9 $\pm$ 7.3 & 43.8 $\pm$ 7.2 &  &  \\ 
 \Xhline{1.5pt}
\end{tabular}%
}
\caption{Evaluation results of current popular models on
  ArxivRollBench2024b for Sequencing Tasks.
  % Results in Cloze and
  % Prediction are in Table \ref{tab:table_c} and Table
  % \ref{tab:table_p} in Appendix.
}
\label{tab:table_s}
\end{table*}

\section{Evaluation}\label{sec:eva}

In this section, we evaluate current popular LLMs with ArxivRollBench and correspondingly quantify the proportions of the overestimation. Specifically, we first detail the settings of our evaluation in Section \ref{sec:settings}, then introduce the performances of models on our private benchmarks as well as their RS scores in Section \ref{sec:perf-pri} and Section \ref{sec:overestimation}, respectively.

\subsection{Settings}\label{sec:settings}

\subsubsection{Evaluated Models}

We categorize the models benchmarked into two groups: open-source LLMs and close AI models.

\noindent$\bullet$~\textbf{Open-source LLMs}: This category includes
open-source LLMs. We evaluate GPT-J-6B~\cite{gpt-j-6b},
Phi-1~\cite{phi-1}, Phi-1.5~\cite{phi-1.5}, Phi-2~\cite{phi-2},
Phi-3-Mini-4K-Instruct~\cite{phi-3},
Phi-3.5-Mini-Instruct~\cite{phi-3},Phi-4-Reasoning,
Phi-4-Reasoning-Plus, Llama-2-7B-Chat-HF~\cite{llama2},
Llama-3-8B-Instruct~\cite{llama3}, Llama-3.1-8B,
Llama-3.1-8B-Instruct~\cite{llama3},
Llama-3.1-70B-Instruct~\cite{llama3},
Llama-3.1-Nemotron-70B-Instruct-HF~\cite{netron},
Llama-3.2-3B,
Llama-3.3-70B-Instruct,
Qwen2-7B-Instruct~\cite{qwen2}, Qwen2.5-7B~\cite{qwen2},
Qwen2.5-7B-Instruct, Qwen2.5-Math-7B, Qwen2.5-Math-7B-Instruct,
Qwen2.5-72B-Instruct~\cite{qwen2}, Qwen3-8B, Qwen3-14B, Qwen3-32B,
Yi-1.5-34B-Chat, Kimi-K2, and Deepseek-Chat-V3.

\noindent$\bullet$~\textbf{Close AIs}: This group consists of LLMs available as
commercial services, providing API access for integration into various
applications. Our experiments cover GPT-3.5-turbo, GPT-4~\cite{gpt4},
GPT-4o~\cite{gpt4o}, Claude-3.5-Sonnet, Claude-3.7-Sonnet,
Claude-4-Sonnet, Gemini-2.0-Flash-001, and Gemini-2.5-Flash.

\begin{figure*}[h]
  \centering
  \begin{subfigure}[b]{0.70\linewidth}
    \includegraphics[width=\linewidth]{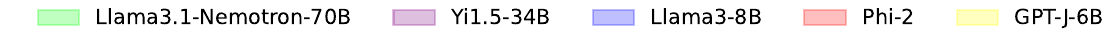}
  \end{subfigure}
  \\
  \begin{subfigure}[b]{0.32\linewidth}
    \includegraphics[width=\linewidth]{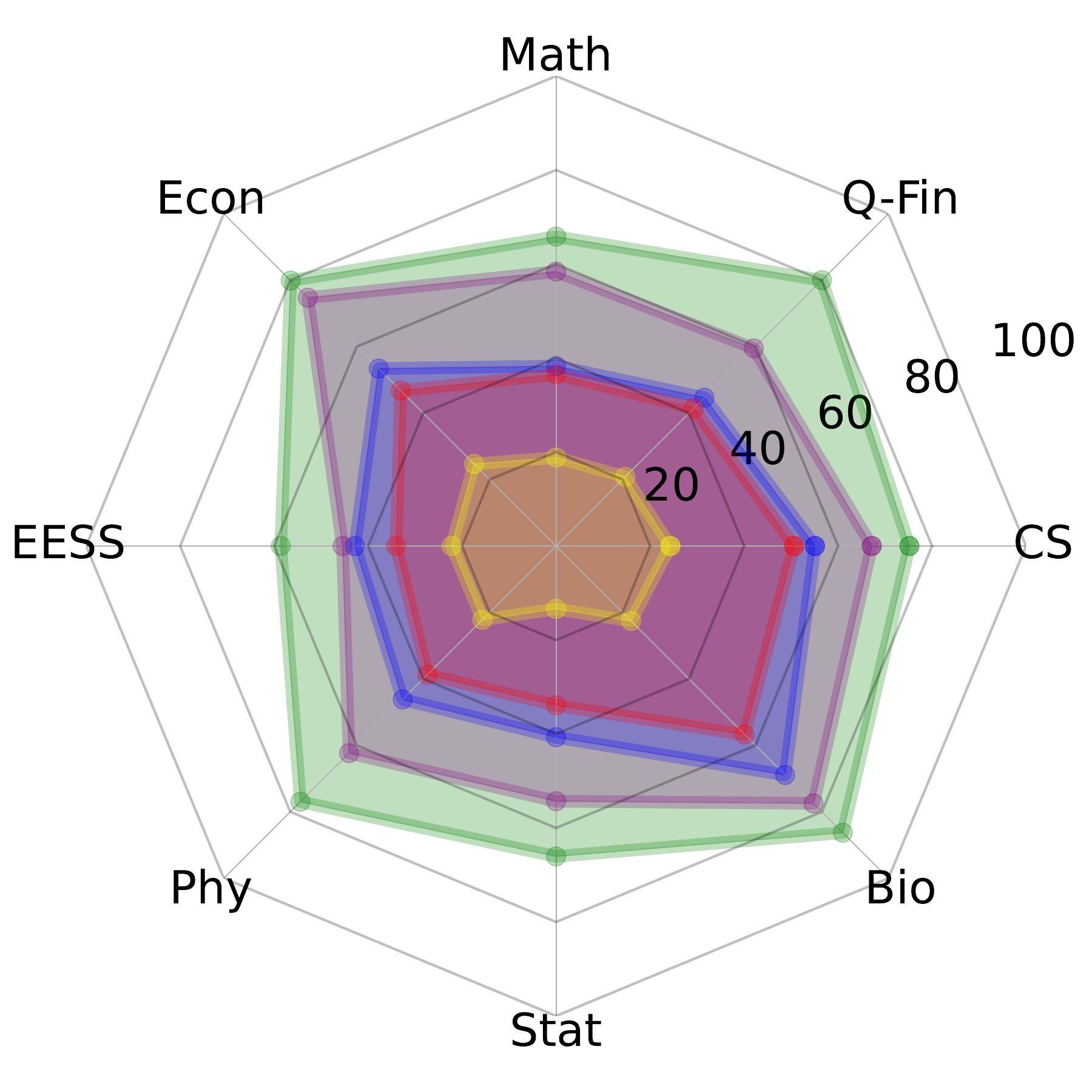}
    \caption{Performance on public benchmarks}
  \end{subfigure}
  \begin{subfigure}[b]{0.32\linewidth}
    \includegraphics[width=\linewidth]{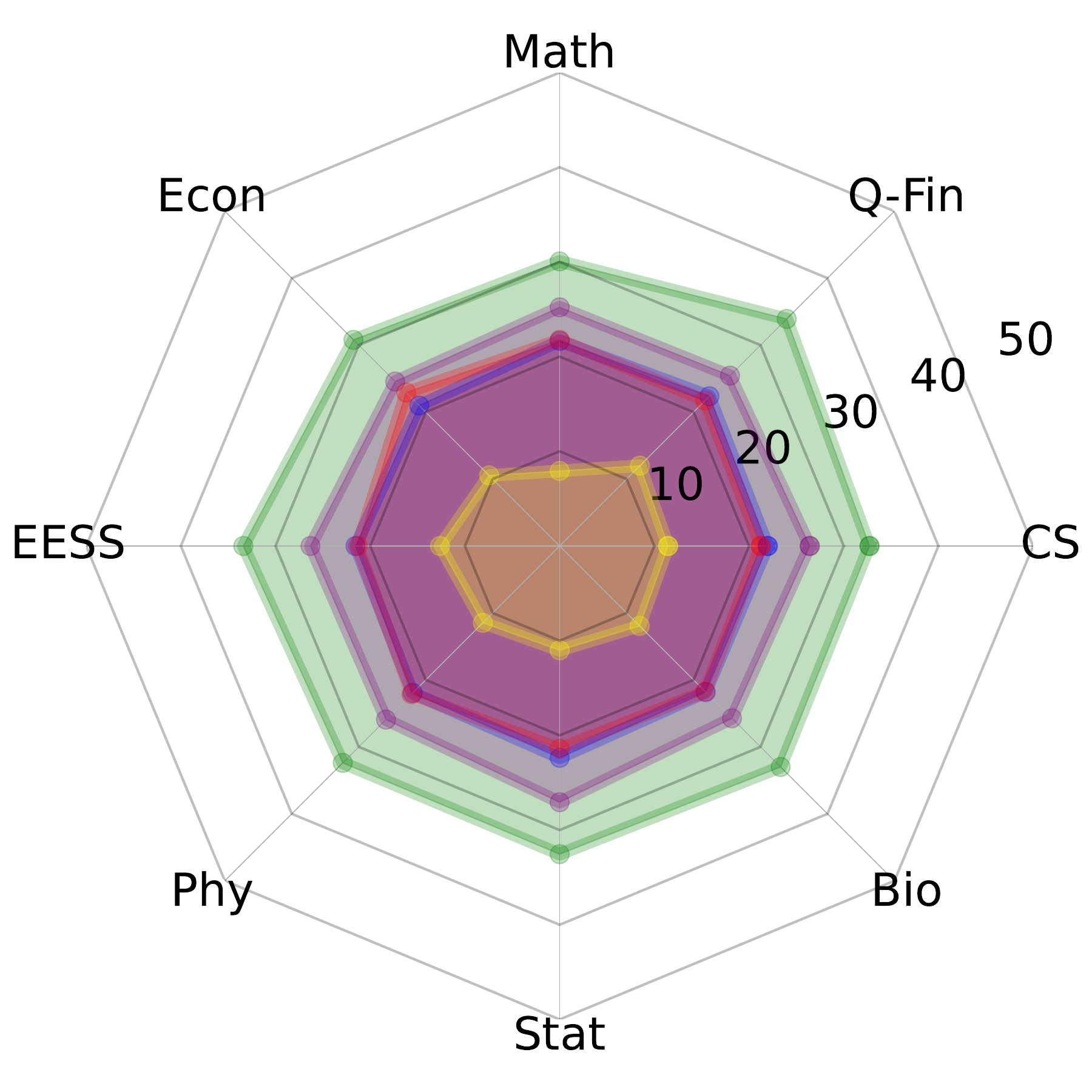}
    \caption{Performance on ArxivRollBench}
  \end{subfigure}
  \begin{subfigure}[b]{0.32\linewidth}
    \includegraphics[width=\linewidth]{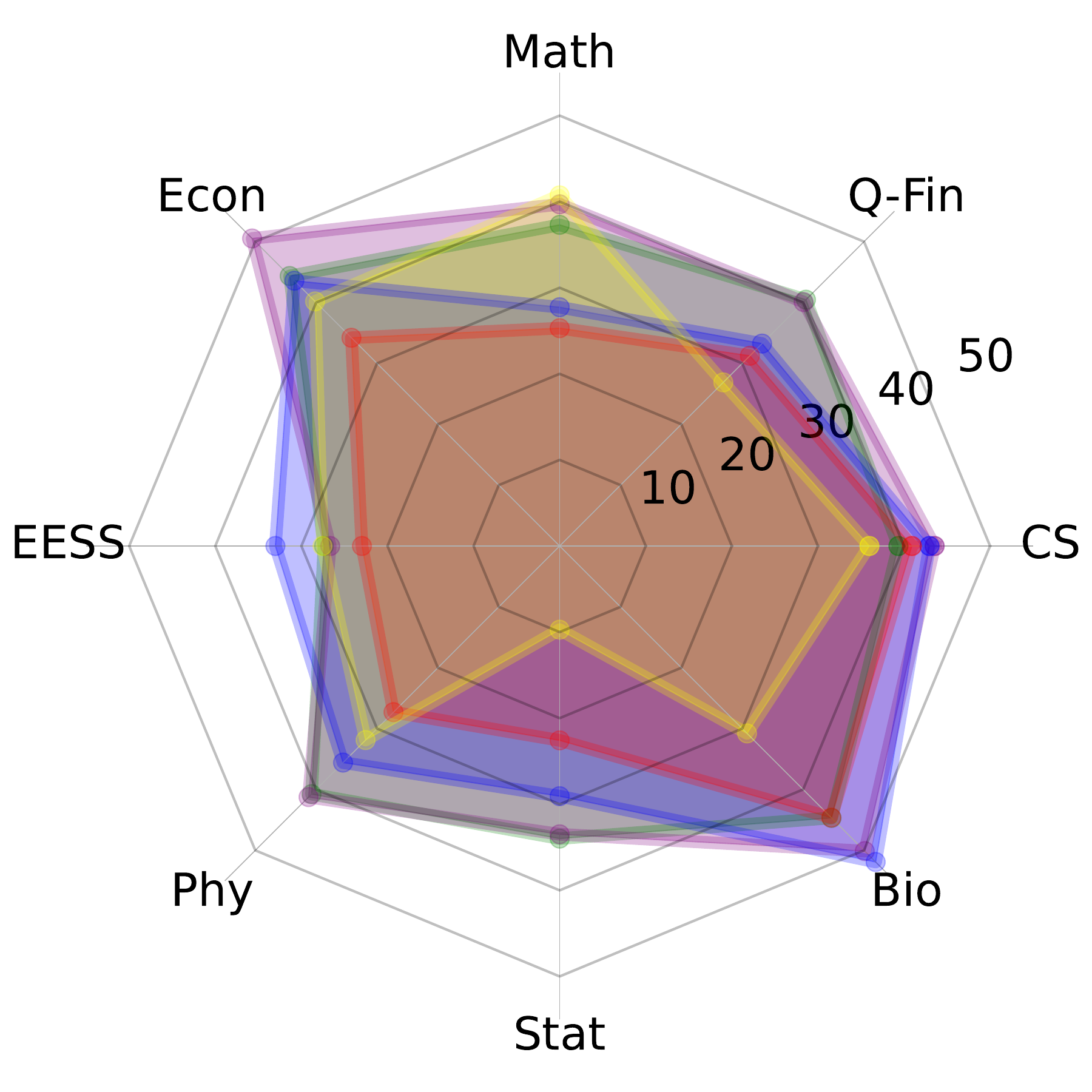}
    \caption{$\mathbf{RS_I}$ performance}
  \end{subfigure}
\caption{Performance of models across different domain benchmarks and the corresponding Absolute $\mathbf{RS_I}$.}
\label{fig:domain_per}
\end{figure*}

\subsubsection{Implementation Details}

We assess the performance of the aforementioned LLMs with LM
Evaluation Harness~\cite{harness}. Following previous studies~\cite{lab1,lab2,lab3,lab4,lab5,lab6}, we use greedy search
in the generation, with the maximized token length of 50. We use the
``exact matching'' for seeking answers and compute the accuracy among
all samples. While the dataset covers eight different domains, the
generation process remains consistent across
them.
% Figure~\ref{fig:details} in Appendix provides an overview of the
% instructions we used.
All open-source LLMs are executed with
$4~\times$ Nvidia H100 GPUs.

\subsection{Evaluating The Performances}\label{sec:perf-pri}

We conduct experiments on our private benchmarks, ArxivRollBench2024b
with Sequencing (S), Cloze (C), and Prediction (P) among 8 domains,
where the results on sequencing are shown in Table \ref{tab:table_s}.
% , Table \ref{tab:table_c}, and Table \ref{tab:table_p}.
We identify several key findings:

\begin{figure}[t]
  \centering
  \includegraphics[width=0.8\linewidth]
    {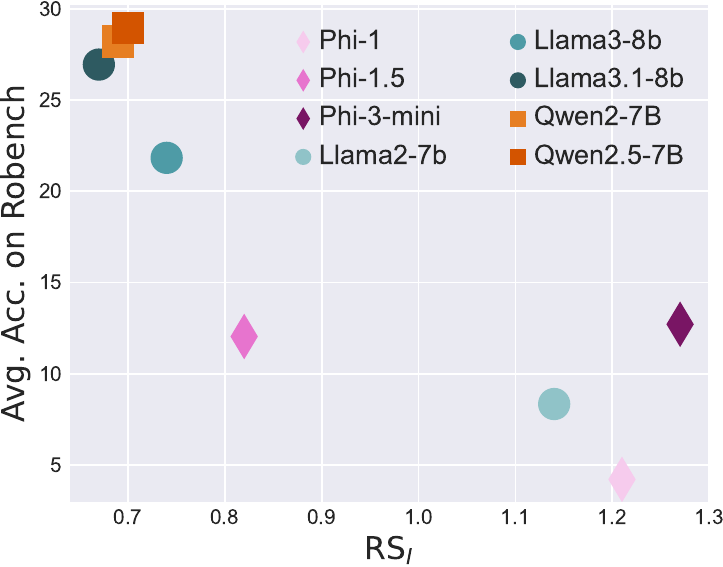}
\caption{Evolution of series models.}
    \label{fig:series}
\end{figure}

\noindent
$\bullet$ \textbf{Open-source LLMs show performance comparable to closeAIs.}
    Open-source LLMs have shown remarkable progress in recent
    years. While the performance of many open-source models remains
    relatively low, certain models rival proprietary counterparts. For
    instance, Kimi-K2, the best-performing open-source model,
    consistently achieves accuracy rates exceeding 40\%, closely
    matching Gemini and Claude and even surpassing it in some tasks.

\noindent
$\bullet$ \textbf{Small Language Models (SLMs) are not consistently comparable to medium-sized models.}
Table \ref{tab:table_s} indicates that Phi-3-mini and Phi-3.5-mini perform poorly on Sequencing and Prediction tasks, respectively, with accuracies not exceeding 10\%. This suggests that, while SLMs can achieve performance comparable to or even exceeding that of 7-billion-scale models on certain tasks, their actual capabilities may sometimes be over-claimed.

\noindent
$\bullet$ \textbf{While newly emerged LLMs \emph{indeed} achieve better performance, the improvements they claim often reflect \emph{growing} overestimation.}
As illustrated in Figure~\ref{fig:series}, it is evident that within
each series, as models evolve, there is an improvement in
accuracy. However, the corresponding $\mathbf{RS_I}$ scores also
increase on some models (e.g., Phi series). This suggests that while the performance of the models is
enhanced through evolution, the degree of overestimation also
escalates.

\subsection{Evaluating The Overestimation}\label{sec:overestimation}

\noindent\textbf{Analysis on $\mathbf{RS_{I}}$ and $\mathbf{RS_{II}}$.} We provide a detailed
comparison of various models in terms of their Absolute
$\mathbf{RS_I}$ and Relative $\mathbf{RS_I}$ in Table~\ref{tab:rs1},
and $\mathbf{RS_{\text{II}}}$ and $\mathbf{N~RS_{\text{II}}}$ in
Table~\ref{tab:rs2}.
Comparing the Absolute $\mathbf{RS_I}$, it is clear that the Qwen
and Phi series exhibit the highest degree of overestimation, which
are even larger than 100\%. Similarly, their corresponding rankings (Relative
$\mathbf{RS_I}$) also show significant changes. As for
$\mathbf{RS_{\text{II}}}$, we observe that models
Llama-3.1-Nemotron-70B and Llama3.1-70B score highly on
$\mathbf{RS_{\text{II}}}$, but their $\mathbf{RS_{\text{II}}^{N}}$ are
relatively lower. This discrepancy is due to their high accuracies
across various domains in ArxivRollBench and the corresponding high absolute
differences. However, after normalization, these differences are not
as pronounced.

\begin{table}[]
\centering
\resizebox{0.48\textwidth}{!}{%
\begin{tabular}{c|cc}
\Xhline{1.5pt}
\multicolumn{1}{c|}{Models} & \multicolumn{1}{c}{Absolute $\mathbf{RS_I}$} & \multicolumn{1}{c}{Relative Rank Changes} \\ \hline
Phi-1 & 1.21 & $\uparrow$ 1.31 \\
Phi-1.5 & 0.82 & $\downarrow$ 0.57 \\
Phi-2 & 0.62 & $\downarrow$ 0.36 \\
Phi-3-mini & 1.27 & $\uparrow$ 0.05 \\
Phi-3.5-mini & 1.07 & $\downarrow$ 0.07 \\
Qwen2-7B & 0.69 & $\downarrow$ 0.42 \\
Qwen2.5-7B & 0.70 & $\downarrow$ 0.37 \\
Qwen2.5-72B & 1.41 & $\downarrow$ 0.32 \\
Yi-1.5-34B & 0.81 & $\downarrow$ 0.55 \\
Llama-3.1-Nemotron-70B & 0.77 & $\uparrow$ 0.14 \\
Llama2-7B & 1.14 & $\uparrow$ 0.11 \\
Llama3-8B & 0.74 & $\uparrow$ 0.74 \\
Llama3.1-8B & 0.67 & $\uparrow$ 0.25 \\
Llama3.1-70B & 0.48 & $\uparrow$ 0.02 \\ \Xhline{1.5pt}
\end{tabular}%
}
\caption{Contamination evaluation with $\mathbf{RS_I}$.}
\label{tab:rs1}
\end{table}

\noindent\textbf{Measuring Biased Overtraining.}
We also select five models (GPT-J-6B, Phi2, Llama3-8B, Yi1.5-34B and
Llama3.1-Nemotron-70B) as references to analyze their performance
across various domain benchmarks and their corresponding Absolute
$\mathbf{RS_I}$, as shown in Figure~\ref{fig:domain_per}. Upon
comparison, it is apparent that model performance on public benchmarks
is inconsistent, with notably better performance in the domains of
Econ, Q-Fin, Bio, and Phy compared to others. However, on ArxivRollBench,
the differences in model performance across various domains are
minimal, indirectly indicating the fairness of ArxivRollBench across these
domains. Besides, it is observed that Absolute $\mathbf{RS_I}$ are
also significantly higher in the Econ, Q-Fin, Bio, and Phy domains,
suggesting that advantages of these models on public benchmarks in these areas might be due to overfitting.

% \usepackage{graphicx}
\begin{table}[]
\centering
\resizebox{0.4\textwidth}{!}{%
\begin{tabular}{c|cc}
\Xhline{1.5pt}
\multicolumn{1}{c|}{Model} & \multicolumn{1}{c}{$\mathbf{RS_{\text{II}}}$} & \multicolumn{1}{c}{$\mathbf{N~RS_{\text{II}}}$} \\ \hline
Phi-1 & 0.22\% & 5.21\% \\
Phi-1.5 & 0.50\%  & 4.02\% \\
Phi-2 & 0.53\% & 2.39\%  \\
Phi-3-mini & 0.76\% & 5.84\%  \\
Phi-3.5-mini & 0.57\% & 3.27\% \\
Qwen2-7B & 0.51\% & 1.76\% \\
Qwen2.5-7B & 0.66\% & 2.21\% \\
Qwen2.5-72B & 0.64\% & 3.84\% \\
Yi-1.5-34B & 0.78\% & 2.91\% \\
Llama-3.1-Nemotron-70B & 1.30\% & 3.88\% \\
Llama2-7B & 0.51\% & 6.20\% \\
Llama3-8B & 0.45\% & 2.00\% \\
Llama3.1-8B & 0.96\% & 3.46\% \\
Llama3.1-70B & 1.19\% & 3.66\% \\ \Xhline{1.5pt}
\end{tabular}%
}
\caption{Biased overtraining evaluation with $\mathbf{RS_{II}}$.}
\label{tab:rs2}
\end{table}

%%% Local Variables:
%%% mode: latex
%%% TeX-master: "main"
%%% End:

\section{Conclusion}
This paper proposes a novel dynamic evaluation framework called
ArxivRoll. It is designed to address the critical issue of
overestimation in evaluating LLMs. The framework introduces SCP
(Sequencing, Cloze, and Prediction), an automated generator of private
test cases, and Rugged Scores (RS), metrics that assess the degree of
public benchmark contamination and training bias. Extensive experiments
conducted demonstrate the high quality and reliability of the
our benchmarks.

\section*{Acknowledgements}
We would like to express our sincere gratitude to the reviewers for
their insightful comments and valuable suggestions. We are also
grateful to Shuxing Fang, Zhenhua Zhou, Yugui Liu, Runshu Wang, Runze
Wang, Kunlong Yang, and Junxiang Zhang from The Hong Kong Polytechnic
University for their suggestions and assistants, to Yueyue Wang from
University of Science and Technology of China for her efforts on this
benchmark, and to Professor Hongxia Yang from
The Hong Kong Polytechnic University for her constructive feedback.
This work was supported by the National Natural Science Foundation of
China (Grant No: 92270123 and 62372122), and the Research Grants
Council (Grant No: 15210023 and 15207725), and the Innovation and
Technology Fund (Grant No: ITS-140-23FP), Hong Kong SAR, China.

% \newpage
% \section*{Limitations}
% While our test cases, extracted from ArXiv articles, cover a
% significant portion of state-of-the-art human knowledge, we
% acknowledge that they are insufficient to fully represent the entirety
% of human frontier knowledge, particularly in fields such as natural
% sciences, social sciences, and beyond. Therefore, we plan to expand
% the range of article sources in the future to encompass a broader
% spectrum of human knowledge.
% Besides, although our framework and the SCP have the potential to
% be extended to a wider scope of evaluations (e.g. multi-modal
% evaluation), currently, we only provide evaluations for language
% models. We may consider evolving RoBench to address broader evaluation
% scenarios in future work. In addition, objective forms of test cases
% beyond the SCP remain a promising direction, though they have not been
% explored in our current studies.

% \newpage

\bibliography{custom}

% \clearpage
% \input{ReproducibilityChecklist}

\appendix
% \clearpage

\begin{figure*}[t]
  \centering
  \includegraphics[width=0.99\linewidth]
    {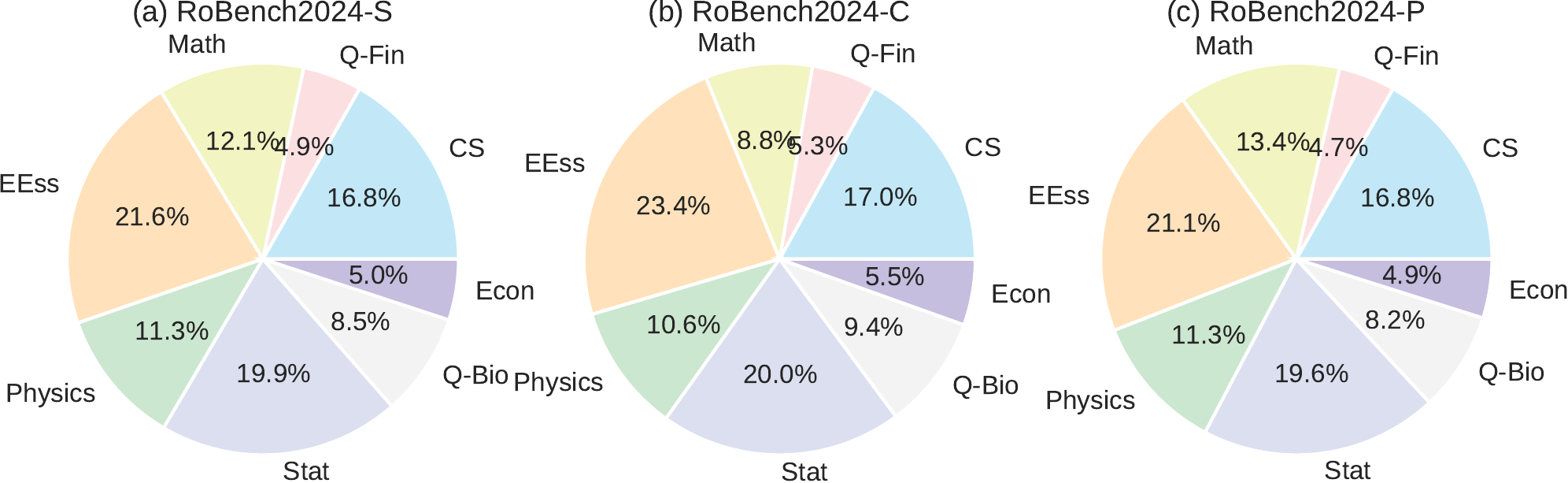}
\caption{Sample numbers distribution of ArxivRollBench2024b among eight
  categories across sequencing (a), cloze (b), and prediction (c).}
    \label{fig:pie}
\end{figure*}

\section{A Detailed Description of Our Private Benchmark: ArxivRollBench2024b}\label{sec:detail-robench}

\noindent
\textbf{Private Benchmark Details.}
As detailed in Section \ref{sec:scp}, we constructed our private benchmarks using preprint papers from Arxiv. Specifically, we downloaded papers uploaded between April 2024 and September 2024 across the following eight domains: Computer Science (CS), Economics (Econ), Electrical Engineering (EESS), Mathematics (Math), Physics (Phy), Biology (Bio), Finance (Fin), and Statistics (Stat). The distribution of collected papers is uneven, with domains such as CS, Math, and Stat containing significantly more articles, while Econ and Fin have fewer. However, even the domains with the smallest number of articles include at least 1,000 papers, ensuring that the constructed benchmark remains robust and suitable for evaluation purposes.

Based on the collected articles, we generated test cases using SCP. Specifically, we began by splitting each article at the ``\textbackslash n'' delimiter and then randomly selecting $N$ consensus phrases to construct a text fragment. To ensure quality, fragments were filtered by checking whether their length exceeded a predefined minimum word count $N_f$, eliminating candidates that were too short. Following this, test cases were generated based on SCP as follows:
\begin{itemize}
    \item Sequencing. The fragment was divided into four parts, permuted, and concatenated.
    \item Cloze. Four sentences within the fragment were randomly masked.
    \item Prediction. The last sentence of the fragment was removed, and three similar candidates were retrieved from the article using TF-IDF similarity.
\end{itemize} 
To improve quality, we identified low-quality text samples and
developed specific rules to exclude them. Additionally, two annotators
manually reviewed the benchmark to further reduce low-quality
samples. For the generation process, we set $N=1$ and $N_f=80$.

% \usepackage{graphicx}
\begin{table*}[htbp]
\centering
\resizebox{0.9\textwidth}{!}{%
\begin{tabular}{crrrrr}
\Xhline{1.5pt}
\multicolumn{6}{c}{ArxivRollBench-2024b-S (CS)} \\ \hline
\multicolumn{1}{c|}{Data Type} & \multicolumn{1}{c}{\# Samples} & \multicolumn{1}{c}{\# Avg. Words} & \multicolumn{1}{c}{\# Median Word Num.} & \multicolumn{1}{c}{Max Word Num.} & \multicolumn{1}{c}{Min Word Num.} \\ \hline
\multicolumn{1}{c|}{Shuffled Text} & 2,931 & 94.89 & 83 & 612 & 20 \\ \hline
\multicolumn{6}{c}{ArxivRollBench-2024b-C (CS)} \\ \hline
\multicolumn{1}{c|}{Data Type} & \multicolumn{1}{c}{\# Samples} & \multicolumn{1}{c}{\# Avg. Words} & \multicolumn{1}{c}{\# Median Word Num.} & \multicolumn{1}{c}{Max Word Num.} & \multicolumn{1}{c}{Min Word Num.} \\ \hline
\multicolumn{1}{c|}{Question} & 2377 & 117.90 & 102 & 571 & 15 \\
\multicolumn{1}{c|}{Candidates} & 2377 & 69.15 & 67 & 342 & 10 \\ \hline
\multicolumn{6}{c}{ArxivRollBench-2024b-P (CS)} \\ \hline
\multicolumn{1}{c|}{Data Type} & \multicolumn{1}{c}{\# Samples} & \multicolumn{1}{c}{\# Avg. Words} & \multicolumn{1}{c}{\# Median Word Num.} & \multicolumn{1}{c}{Max Word Num.} & \multicolumn{1}{c}{Min Word Num.} \\ \hline
  \multicolumn{1}{c|}{Selections} & 3166 & 60.76 & 42 & 542 & 1 \\
  \Xhline{1.1pt}
\end{tabular}%
}
\caption{Basic statistical information of ArxivRollBench2024b's computer
  science category.}
\label{tab:num-stat}
\end{table*}

After construction, the final distributions of ArxivRollBench's private
benchmarks are illustrated in Figure \ref{fig:pie}. Domains such as
Econ, Fin, and Bio account for a smaller proportion of samples
compared to EESS, CS, Phy, Stat, and Math. A detailed statistical
analysis of the private benchmarks is presented in Table
\ref{tab:num-stat}, revealing that most benchmarks exhibit high
diversity in sequence length. Furthermore, the settings of $N$ and
$N_f$ result in an average context length of approximately 100 words,
making the benchmarks suitable for evaluating most current LLMs.

\noindent
\textbf{Public-Private Benchmark Pairs.}
As illustrated in Figure \ref{fig:robench}, ArxivRollBench leverages both private and public benchmarks to estimate the extent of overestimation. For public benchmarks, we employed two widely used and comprehensive datasets: MMLU and MMLU Pro. Additionally, we supplemented the public benchmarks with domain-specific datasets, such as those commonly used in Math. In the future, expired private benchmarks will also be incorporated into the public benchmarks to further enhance their coverage and utility.

\section{External Experiment Results}

% \usepackage{graphicx}
\begin{table*}[h]
\centering
\resizebox{\textwidth}{!}{%
\begin{tabular}{c|rrrrrrrrrr}
\Xhline{1.5pt}
Model name & \multicolumn{8}{c}{ArxivRollBench-2024b (C)} &  &  \\ \hline
 & CS & Q-Fin. & Math & Phy. & Stat. & Bio. & Econ. & EESS &  &  \\ \hline
GPT-J-6B & 2.6 $\pm$ 0.3 & 3.3$\pm$ 0.7 & 2.1$\pm$ 0.4 & 4.6  $\pm$ 0.5 & 3.5$\pm$ 0.3 & 2.9$\pm$ 0.5 & 2.4$\pm$0.5 & 3.5 $\pm$ 0.3 &  &  \\
Phi-1 & 1.9 $\pm$ 0.3 & 1.9 $\pm$ 0.5 & 1.7 $\pm$ 0.4 & 1.6  $\pm$ 0.3 & 1.9$\pm$ 0.3 & 1.2 $\pm$ 0.3 & 1.8 $\pm$0.5 & 1.6 $\pm$ 0.2 &  &  \\
Phi-1.5 & 10.0 $\pm$ 0.6 & 11.6 $\pm$ 1.2 & 10.4 $\pm$ 0.9 & 9.0 $\pm$ 0.7 & 13.5 $\pm$ 0.6 & 10.7$\pm$ 0.9 & 11.8 $\pm$ 1.2 & 10.8 $\pm$ 0.5 &  &  \\
Phi-2 & 16.9 $\pm$ 0.8 & 19.1$\pm$ 1.4 & 19.0 $\pm$ 1.1 & 19.1 $\pm$ 1.0 & 18.2$\pm$ 0.7 & 15.7 $\pm$ 1.0 & 19.4 $\pm$ 1.4 & 18.1 $\pm$ 0.7 &  &  \\
Phi-3-Mini-4K-Instruct & 21.9 $\pm$ 0.8 & 22.0 $\pm$ 1.5 & 17.3$\pm$ 1.1 & 19.0 $\pm$ 1.0 & 20.6 $\pm$ 0.8 & 19.5 $\pm$ 1.1 & 19.5 $\pm$ 1.4 & 22.3 $\pm$ 0.7 &  &  \\
Phi-3.5-Mini-Instruct & 27.6 $\pm$ 0.9 & 27.8 $\pm$ 1.6 & 27.1 $\pm$ 1.3 & 30.0 $\pm$ 1.2 & 27.3 $\pm$ 0.8 & 26.3 $\pm$ 1.2 & 24.5 $\pm$ 1.6 & 28.8 $\pm$ 0.8 &  &  \\
Phi4-Reasoning & 10.9 $\pm$ 0.6 & 15.1 $\pm$ 1.3 & 15.3 $\pm$ 1.0 & 11.7 $\pm$ 0.8 & 13.5 $\pm$ 0.6 & 11.2 $\pm$ 0.9 & 15.6 $\pm$ 1.3 & 12.1 $\pm$ 0.6 &  &  \\ 
Phi4-Reasoning-Plus & 23.1 $\pm$ 0.9 & 24.1 $\pm$ 1.6 & 21.2 $\pm$ 1.2 & 19.9 $\pm$ 1.0 & 23.4 $\pm$ 0.8 & 21.9 $\pm$ 1.1 & 21.7 $\pm$ 1.5 & 24.0 $\pm$ 0.7 &  &  \\ 
Qwen2-7B-Instruct & 24.3 $\pm$ 0.9 & 22.8 $\pm$ 1.5 & 27.4 $\pm$ 1.3 & 25.4 $\pm$ 1.1 & 25.2 $\pm$ 0.8 & 23.8 $\pm$ 1.2 & 23.7 $\pm$ 1.5 & 25.7 $\pm$ 0.8 &  &  \\
Qwen2.5-7B & 25.0 $\pm$ 0.9 & 24.8 $\pm$ 1.6 & 28.2 $\pm$ 1.3 & 24.8 $\pm$ 1.1 & 28.2 $\pm$ 0.9 & 25.9 $\pm$ 1.2 & 26.2 $\pm$ 1.6 & 27.3 $\pm$ 0.8 &  &  \\
Qwen2.5-7B-Instruct & 22.5 $\pm$ 0.9 & 22.0 $\pm$ 1.5 & 24.8 $\pm$ 1.2 & 20.4 $\pm$ 1.0 & 23.5 $\pm$ 0.8 & 21.2 $\pm$ 1.1 & 22.3 $\pm$ 1.5 & 24.3 $\pm$ 0.7 &  &  \\
Qwen2.5-Math-7B & 24.2 $\pm$ 0.9 & 22.4 $\pm$ 1.5 & 23.7 $\pm$ 1.2 & 24.1 $\pm$ 1.1 & 24.8 $\pm$ 0.8 & 24.6 $\pm$ 1.2 & 24.2 $\pm$ 1.6 & 24.9 $\pm$ 0.8 &  &  \\
Qwen2.5-Math-7B-Instruct & 1.5 $\pm$ 0.3 & 1.2 $\pm$ 0.4 & 2.3 $\pm$ 0.4 & 1.8 $\pm$ 0.3 & 1.5 $\pm$ 0.2 & 1.2 $\pm$ 0.3 & 2.4 $\pm$ 0.5 & 1.4 $\pm$ 0.2 &  &  \\
Qwen2.5-72B-Instruct& 14.7 $\pm$ 0.7 & 15.1 $\pm$ 1.3 & 16.3 $\pm$ 1.1 & 11.9 $\pm$ 0.8 & 16.4 $\pm$ 0.7 & 11.2 $\pm$ 0.9 & 15.6 $\pm$ 1.3 & 12.4 $\pm$ 0.6 &  &  \\
Qwen3-8B & 24.0 $\pm$ 0.9 & 23.7 $\pm$ 1.6 & 25.3 $\pm$ 1.2 & 24.4 $\pm$ 1.1 & 24.3 $\pm$ 0.8 & 23.9 $\pm$ 1.2 & 26.4 $\pm$ 1.6 & 24.4 $\pm$ 0.8 &  &  \\ 
Qwen3-14B & 4.3 $\pm$ 0.4 & 5.9 $\pm$ 0.9 & 7.6 $\pm$ 0.8 & 5.7 $\pm$ 0.6 & 5.1 $\pm$ 0.4 & 4.2 $\pm$ 0.6 & 5.9 $\pm$ 0.9 & 2.8 $\pm$ 0.3 &  &  \\ 
Qwen3-32B & 4.7 $\pm$ 0.4 & 3.7 $\pm$ 0.7 & 4.8 $\pm$ 0.6 & 3.1 $\pm$ 0.5 & 5.5 $\pm$ 0.4 & 3.0 $\pm$ 0.5 & 3.9 $\pm$ 0.7 & 3.7 $\pm$ 0.3 &  &  \\ 
Llama2-7B-Chat-HF & 3.2 $\pm$ 0.4 & 1.9 $\pm$ 0.5 & 5.4 $\pm$ 0.6 & 2.0 $\pm$ 0.4 & 2.1 $\pm$ 0.3 & 1.2 $\pm$ 0.3 & 1.3 $\pm$ 0.4 & 1.8 $\pm$ 0.2 &  &  \\
Llama3-8B & 18.2 $\pm$ 0.8 & 18.2 $\pm$ 1.4 & 19.2 $\pm$ 1.1 & 18.2 $\pm$ 1.0 & 20.4 $\pm$ 0.8 & 19.0 $\pm$ 1.1 & 17.5 $\pm$ 1.4 & 18.6 $\pm$ 0.7 &  &  \\
Llama3.1-8B& 14.3$\pm$ 0.7 & 14.2 $\pm$ 1.3 & 17.3 $\pm$ 1.1 & 15.9 $\pm$ 1.0 & 15.6 $\pm$ 0.7 & 13.8 $\pm$ 1.0 & 13.0 $\pm$ 1.2 & 14.9 $\pm$ 0.6 &  &  \\
Llama3.1-8B-Instruct& 26.2 $\pm$ 0.9 & 23.6 $\pm$ 1.6 & 25.2 $\pm$ 1.2 & 25.5 $\pm$ 1.1 & 25.0 $\pm$ 0.8 & 24.5 $\pm$ 1.2 & 22.9 $\pm$ 1.5 & 27.4 $\pm$ 0.8 &  &  \\
Llama3.1-70B-Instruct& 27.2 $\pm$ 0.9 & 26.5 $\pm$ 1.6 & 25.9 $\pm$ 1.2 & 27.1 $\pm$ 1.2 & 28.1 $\pm$ 0.8 & 27.2 $\pm$ 1.2 & 25.9 $\pm$ 1.6 & 27.3 $\pm$ 0.8 &  &  \\
Llama3.1-Nemotron-70B& 26.9 $\pm$ 0.9 & 26.8 $\pm$ 1.6 & 25.4 $\pm$ 1.2 & 27.5$\pm$ 1.2 & 28.0 $\pm$ 0.8   & 27.2 $\pm$ 1.2 & 25.5 $\pm$ 1.6  & 27.5 $\pm$ 0.8 & &  \\ 
Llama3.2-1B & 15.4 $\pm$ 0.7 & 13.3 $\pm$ 1.2 & 12.8 $\pm$ 0.9 & 11.5 $\pm$ 0.8 & 17.8 $\pm$ 0.7 & 13.7 $\pm$ 0.9 & 15.4 $\pm$ 1.3 & 16.0 $\pm$ 0.6 &  &  \\ 
Llama3.2-3B & 23.8 $\pm$ 0.9 & 26.5 $\pm$ 1.6 & 24.2 $\pm$ 1.2 & 25.4 $\pm$ 1.1 & 25.8 $\pm$ 0.8 & 23.9 $\pm$ 1.2 & 25.1 $\pm$ 1.6 & 25.9 $\pm$ 0.8 &  &  \\ 
Llama3.3-70B-Instruct & 13.5 $\pm$ 0.7 & 13.3 $\pm$ 1.2 & 15.5 $\pm$ 1.0 & 14.7 $\pm$ 0.9 & 15.1 $\pm$ 0.7 & 13.2 $\pm$ 0.9 & 11.4 $\pm$ 1.2 & 14.1 $\pm$ 0.6 &  &  \\ 
Yi1.5-34B & 19.8 $\pm$ 0.8 & 21.2 $\pm$ 1.5 & 21.0 $\pm$ 1.2 & 20.3 $\pm$ 1.0 & 19.9 $\pm$ 0.8 & 19.8 $\pm$1.1 & 18.6 $\pm$ 1.4 & 20.6 $\pm$ 0.7 &  &  \\
Kimi-K2 & 25.8 $\pm$ 8.0 & 27.3 $\pm$ 6.8 & 26.7 $\pm$ 11.8 & 28.6 $\pm$ 8.7 & 24.2 $\pm$ 7.6 & 32.4 $\pm$ 8.1 & 34.9 $\pm$ 7.4 & 40.5 $\pm$ 7.7 &  &  \\ 
Deepseek-Chat-V3 & 19.4 $\pm$ 7.2 & 15.9 $\pm$ 5.6 & 33.3 $\pm$ 12.6 & 17.9 $\pm$ 7.4 & 24.2 $\pm$ 7.6 & 14.7 $\pm$ 6.2 & 18.6 $\pm$ 6.0 & 19.0 $\pm$ 6.1 &  &  \\
\hline
GPT-3.5-turbo  & 29.0 $\pm$ 8.3 & 22.72 $\pm$ 6.4 & 40.0 $\pm$ 13.1 & 21.4 $\pm$ 7.9 & 21.2 $\pm$ 7.2 & 20.6 $\pm$ 7.0 & 34.9 $\pm$ 7.4 & 31.0 $\pm$ 7.2 &  &  \\   
GPT-4 & 16.1 $\pm$ 6.7 & 18.2 $\pm$ 5.9 & 33.3 $\pm$ 12.6 & 14.3 $\pm$ 6.7 & 21.2 $\pm$ 7.2 & 23.5 $\pm$ 7.4 & 27.9 $\pm$ 6.9 & 23.8 $\pm$ 6.7 &  &  \\ 
GPT-4o & 25.8 $\pm$ 8.0 & 15.9 $\pm$ 5.6 & 26.7 $\pm$ 11.8 & 10.7 $\pm$ 6.0 & 30.3 $\pm$ 8.1 & 23.5 $\pm$ 7.4 & 23.3 $\pm$ 6.5 & 16.7 $\pm$ 5.8 &  &  \\
Claude-3.5-Sonnet & 25.8 $\pm$ 8.0 & 22.7 $\pm$ 6.4 & 33.3 $\pm$ 12.6 & 35.7 $\pm$ 9.2 & 33.3 $\pm$ 8.3 & 32.4 $\pm$ 8.1 & 34.9 $\pm$ 7.4 & 28.6 $\pm$ 7.1 &  &  \\ 
Claude-3.7-Sonnet & 19.4 $\pm$ 7.2 & 20.5 $\pm$ 6.2 & 40.0 $\pm$ 13.1 & 17.9 $\pm$ 7.4 & 21.2 $\pm$ 7.2 & 20.6 $\pm$ 7.0 & 34.9 $\pm$ 7.4 & 23.8 $\pm$ 6.7 &  &  \\ 
Claude-4-Sonnet & 12.9 $\pm$ 6.1 & 9.1 $\pm$ 4.4 & 20.0 $\pm$ 10.7 & 7.1 $\pm$ 5.0 & 27.3 $\pm$ 7.9 & 5.9 $\pm$ 4.1 & 11.6 $\pm$ 4.9 & 14.3 $\pm$ 5.5 &  &  \\
Claude-4-Opus & 6.5 $\pm$ 4.5 & 2.3 $\pm$ 2.3 & 0.0 $\pm$ 0.0 & 0.0 $\pm$ 0.0 & 12.1 $\pm$ 5.8 & 8.8 $\pm$ 4.9 & 14.0 $\pm$ 5.3 & 4.8 $\pm$ 3.3 &  &  \\
Gemini-2.0-flash-001 & 19.4 $\pm$ 7.2 & 25.0 $\pm$ 6.6 & 40.0 $\pm$ 13.1 & 28.6 $\pm$ 8.7 & 33.3 $\pm$ 8.3 & 32.4 $\pm$ 8.1 & 23.3 $\pm$ 6.5 & 23.8 $\pm$ 6.7 &  &  \\ 
Gemini-2.5-flash & 22.6 $\pm$ 7.6 & 13.6 $\pm$ 5.2 & 33.3 $\pm$ 12.6 & 14.3 $\pm$ 6.7 & 33.3 $\pm$ 8.3 & 29.4 $\pm$ 7.9 & 23.3 $\pm$ 6.5 & 31.0 $\pm$ 7.2 &  &  \\ 
  \Xhline{1.5pt}
\end{tabular}%
}
\caption{Evaluation results of current popular models on ArxivRollBench for Cloze tasks.}
\label{tab:table_c}
\end{table*}

% \usepackage{graphicx}
\begin{table*}[h]
\centering
\resizebox{\textwidth}{!}{%
\begin{tabular}{c|rrrrrrrrrr}
\Xhline{1.5pt}
Model name & \multicolumn{8}{c}{ArxivRollBench-2024b (P)} &  &  \\ \hline
 & CS & Q-Fin. & Math & Phy. & Stat. & Bio. & Econ. & EESS &  &  \\ \hline
%%%%%
GPT-J-6B & 21.5 $\pm$ 0.7   & 20.4 $\pm$ 1.4   & 13.7 $\pm$ 0.7   & 18.1 $\pm$ 0.8  & 19.9 $\pm$ 0.7   & 21.0 $\pm$ 1.0   & 19.6 $\pm$1.3    & 21.9 $\pm$ 0.7  &  &  \\

Phi-1 & 4.9 $\pm$ 0.4 & 4.4 $\pm$ 0.7 & 3.2 $\pm$ 0.4 & 3.2 $\pm$ 0.4 & 4.1 $\pm$ 0.3 & 5.4 $\pm$ 0.6 & 4.8 $\pm$ 0.7 & 4.5 $\pm$ 0.3 &  &  \\

Phi-1.5 & 2.0 $\pm$ 0.3 & 2.0 $\pm$ 0.5 & 2.1 $\pm$ 0.3 & 2.6 $\pm$ 0.3 & 1.6 $\pm$ 0.2 & 2.8 $\pm$ 0.4 & 1.4 $\pm$ 0.4 & 2.1 $\pm$ 0.2 &  &  \\

Phi-2 & 23.7 $\pm$ 0.8 & 23.3 $\pm$ 1.4 & 21.6 $\pm$ 0.8 & 22.9 $\pm$ 0.9 & 22.4 $\pm$ 0.7 & 25.2 $\pm$ 1.1 & 24.6 $\pm$ 1.4 & 22.8 $\pm$ 0.7 &  &  \\

Phi-3-Mini-4K-Instruct  & 11.9 $\pm$ 0.6 & 13.3 $\pm$ 1.1 & 12.3 $\pm$ 0.7 & 12.8 $\pm$ 0.7 & 12.4 $\pm$ 0.5 & 13.0 $\pm$ 0.9 & 12.4 $\pm$ 1.1 & 11.9 $\pm$ 0.5 &  &  \\

Phi-3.5-Mini-Instruct & 4.0 $\pm$ 0.4 & 4.3 $\pm$ 0.7 & 5.6 $\pm$ 0.5 & 5.3 $\pm$ 0.5 & 4.1 $\pm$ 0.3 & 4.6 $\pm$ 0.5 & 4.0 $\pm$ 0.7 & 3.9 $\pm$ 0.3 &  &  \\

Phi4-Reasoning & 0.4 $\pm$ 0.1 & 0.8 $\pm$ 0.3 & 0.4 $\pm$ 0.1 & 0.8 $\pm$ 0.2 & 0.5 $\pm$ 0.1 & 0.5 $\pm$ 0.2 & 0.5 $\pm$ 0.2 & 0.6 $\pm$ 0.1 &  &  \\ 

Phi4-Reasoning-Plus & 21.9 $\pm$ 0.7 & 22.5 $\pm$ 1.4 & 18.6 $\pm$ 0.8 & 21.3 $\pm$ 0.9 & 22.6 $\pm$ 0.7 & 25.5 $\pm$ 1.1 & 26.2 $\pm$ 1.5 & 24.4 $\pm$ 0.7 &  &  \\ 

Qwen2-7B-Instruct & 34.2 $\pm$ 0.8 & 32.0 $\pm$ 1.6 & 30.2 $\pm$ 0.9 & 32.5 $\pm$ 1.0 & 32.9 $\pm$ 0.8 & 32.2 $\pm$ 1.2 & 33.8 $\pm$ 1.6 & 33.4 $\pm$ 0.8 &  &  \\

Qwen2.5-7B & 32.9 $\pm$ 0.8 & 30.3 $\pm$ 1.5 & 30.5 $\pm$ 0.9 & 33.1 $\pm$ 1.0 & 32.9 $\pm$ 0.8 & 32.3 $\pm$ 1.2 & 33.0 $\pm$ 1.6 & 33.7 $\pm$ 0.7 &  &  \\

Qwen2.5-7B-Instruct & 37.8 $\pm$ 0.9 & 34.4 $\pm$ 1.6 & 34.1 $\pm$ 0.9 & 37.7 $\pm$ 1.1 & 37.0 $\pm$ 0.8 & 39.4 $\pm$ 1.2 & 34.7 $\pm$ 1.6 & 37.3 $\pm$ 0.8 &  &  \\

Qwen2.5-Math-7B & 22.0 $\pm$ 0.7 & 22.9 $\pm$ 1.4 & 20.5 $\pm$ 0.8 & 21.5 $\pm$ 0.9 & 21.4 $\pm$ 0.7 & 22.6 $\pm$ 1.1 & 22.3 $\pm$ 1.4 & 21.6 $\pm$ 0.7 &  &  \\

Qwen2.5-Math-7B-Instruct & 13.6 $\pm$ 0.6 & 18.3 $\pm$ 1.3 & 11.1 $\pm$ 0.6 & 14.1 $\pm$ 0.8 & 14.8 $\pm$ 0.6 & 15.9 $\pm$ 0.9 & 17.7 $\pm$ 1.3 & 13.9 $\pm$ 0.5 &  &  \\

Qwen2.5-72B-Instruct & 3.4 $\pm$ 0.3 & 5.3 $\pm$ 0.7 & 6.3 $\pm$ 0.5 & 5.6 $\pm$ 0.5 & 3.6 $\pm$ 0.3 & 4.3 $\pm$ 0.5 & 6.4 $\pm$ 0.8 & 3.6 $\pm$ 0.3 &  &  \\ 

Qwen3-8B & 31.3 $\pm$ 0.8 & 30.6 $\pm$ 1.6 & 26.6 $\pm$ 0.9 & 30.0 $\pm$ 1.0 & 29.8 $\pm$ 0.8 & 31.2 $\pm$ 1.2 & 28.9 $\pm$ 1.5 & 31.1 $\pm$ 0.7 &  &  \\ 

Qwen3-14B & 11.8 $\pm$ 0.6 & 11.5 $\pm$ 1.1 & 14.3 $\pm$ 0.7 & 13.0 $\pm$ 0.7 & 11.6 $\pm$ 0.5 & 8.4 $\pm$ 0.7 & 10.7 $\pm$ 1.0 & 10.3 $\pm$ 0.5 &  &  \\ 

Qwen3-32B & 3.2 $\pm$ 0.3 & 2.7 $\pm$ 0.5 & 3.7 $\pm$ 0.4 & 3.5 $\pm$ 0.4 & 3.1 $\pm$ 0.3 & 2.5 $\pm$ 0.4 & 3.0 $\pm$ 0.6 & 2.6 $\pm$ 0.3 &  &  \\ 

Llama2-7B-Chat-HF & 13.7 $\pm$ 0.6 & 16.0 $\pm$ 1.2 & 10.8 $\pm$ 0.6 & 14.5 $\pm$ 0.8 & 15.2 $\pm$ 0.6 & 16.7 $\pm$ 1.0 & 15.7 $\pm$ 1.2 & 14.4 $\pm$ 0.6 &  &  \\

Llama3-8B & 24.9 $\pm$ 0.8 & 26.1 $\pm$ 1.5 & 24.1 $\pm$ 0.9 & 24.6 $\pm$ 0.9 & 24.3 $\pm$ 0.7 & 22.8 $\pm$ 1.1 & 24.9 $\pm$ 1.4 & 24.7 $\pm$ 0.7 &  &  \\

Llama3.1-8B & 24.2 $\pm$ 0.8 & 25.1 $\pm$ 1.5 & 23.5 $\pm$ 0.8 & 25.1 $\pm$ 0.9 & 24.9 $\pm$ 0.7 & 23.5 $\pm$ 1.1 & 26.3 $\pm$ 1.5 & 23.5 $\pm$ 0.7 &  &  \\

Llama3.1-8B-Instruct & 29.5 $\pm$ 0.8 & 29.3 $\pm$ 1.5 & 26.6 $\pm$ 0.9 & 31.5 $\pm$ 1.0 & 29.9 $\pm$ 0.8 & 28.6 $\pm$ 1.1 & 28.4 $\pm$ 1.5 & 29.7 $\pm$ 0.7 &  &  \\

Llama3.1-70B-Instruct & 36.3 $\pm$ 0.9 & 37.7 $\pm$ 1.6 & 33.4 $\pm$ 0.9 & 36.0 $\pm$ 1.0 & 37.1 $\pm$ 0.8 & 36.9 $\pm$ 1.2 & 32.0 $\pm$ 1.5 & 37.4 $\pm$ 0.8 &  &  \\

Llama3.1-Nemotron-70B & 37.9 $\pm$ 0.9 & 39.2 $\pm$ 1.6 & 34.6 $\pm$ 1.0 & 36.9 $\pm$ 1.0 & 37.6 $\pm$ 0.8  & 37.4 $\pm$ 1.2 & 33.6 $\pm$ 1.6 & 38.3 $\pm$ 0.8 & &  \\ 

Llama3.2-1B & 26.3 $\pm$ 0.8 & 26.2 $\pm$ 1.5 & 24.9 $\pm$ 0.9 & 25.3 $\pm$ 0.9 & 25.4 $\pm$ 0.7 & 25.6 $\pm$ 1.1 & 25.0 $\pm$ 1.4 & 24.2 $\pm$ 0.7 &  &  \\ 

Llama3.2-3B & 22.5 $\pm$ 0.7 & 23.3 $\pm$ 1.4 & 23.1 $\pm$ 0.8 & 23.2 $\pm$ 0.9 & 22.7 $\pm$ 0.7 & 20.7 $\pm$ 1.0 & 20.8 $\pm$ 1.3 & 21.4 $\pm$ 0.7 &  &  \\ 

Llama3.3-70B-Instruct & 38.9 $\pm$ 0.9 & 38.5 $\pm$ 1.6 & 34.3 $\pm$ 0.9 & 37.5 $\pm$ 1.0 & 38.4 $\pm$ 0.8 & 38.1 $\pm$ 1.2 & 36.2 $\pm$ 1.6 & 38.0 $\pm$ 0.8 &  &  \\ 

Yi1.5-34B & 31.3 $\pm$ 0.8 & 27.1 $\pm$ 1.5 & 28.8 $\pm$ 0.9 & 31.0 $\pm$ 1.0 & 31.5 $\pm$ 0.8 & 30.3 $\pm$ 1.2 & 29.4 $\pm$ 1.5 & 30.5 $\pm$ 0.7 &  &  \\

Kimi-K2 & 48.0 $\pm$ 7.1 & 52.0 $\pm$ 7.1 & 39.2 $\pm$ 6.9 & 31.4 $\pm$ 6.6 & 44.0 $\pm$ 7.1 & 44.9 $\pm$ 7.2 & 52.0 $\pm$ 7.1 & 47.1 $\pm$ 7.1 &  &  \\ 

Deepseek-Chat-V3 & 42.0 $\pm$ 7.1 & 56.0 $\pm$ 7.1 & 39.2 $\pm$ 6.9 & 33.3 $\pm$ 6.7 & 52.0 $\pm$ 7.1 & 49.0 $\pm$ 7.2 & 44.0 $\pm$ 7.1 & 47.1 $\pm$ 7.1 &  &  \\ 

 \hline
 
GPT-3.5-turbo & 24.0 $\pm$ 6.1 & 40.0 $\pm$ 7.0 & 31.4 $\pm$ 6.6 & 33.3 $\pm$ 6.7 & 38.0 $\pm$ 6.9 & 38.8 $\pm$ 7.0 & 32.0 $\pm$ 6.7 & 25.5 $\pm$ 6.2 &  &  \\

GPT-4 & 36.0 $\pm$ 6.9 & 52.0 $\pm$ 7.1 & 43.1 $\pm$ 7.0 & 31.4 $\pm$ 6.6 & 60.0 $\pm$ 7.0 & 51.0 $\pm$ 7.2 & 50.0 $\pm$ 7.1 & 37.3 $\pm$ 6.8 &  &  \\

GPT-4o & 40.0 $\pm$ 7.0 & 52.0 $\pm$ 7.1 & 39.2 $\pm$ 6.9 & 23.5 $\pm$ 6.0 & 58.0 $\pm$ 7.1 & 51.0 $\pm$ 7.2 & 46.0 $\pm$ 7.1 & 39.2 $\pm$ 6.9 &  &  \\  

Claude-3.5-Sonnet & 48.0 $\pm$ 7.1 & 58.0 $\pm$ 7.1 & 49.0 $\pm$ 7.1 & 43.1 $\pm$ 7.0 & 54.0 $\pm$ 7.1 & 57.1 $\pm$ 7.1 & 52.0 $\pm$ 7.1 & 58.8 $\pm$ 7.0 &  &  \\ 

Claude-3.7-Sonnet & 52.0 $\pm$ 7.1 & 60.0 $\pm$ 7.0 & 45.1 $\pm$ 7.0 & 37.3 $\pm$ 6.8 & 52.0 $\pm$ 7.1 & 57.1 $\pm$ 7.1 & 60.0 $\pm$ 7.0 & 58.8 $\pm$ 7.0 &  &  \\ 

Claude-4-Sonnet & 58.0 $\pm$ 7.1 & 66.0 $\pm$ 6.8 & 47.1 $\pm$ 7.1 & 41.2 $\pm$ 7.0 & 62.0 $\pm$ 6.9 & 55.1 $\pm$ 7.2 & 60.0 $\pm$ 7.0 & 58.8 $\pm$ 7.0 &  &  \\ 

Gemini-2.0-flash-001 & 38.0 $\pm$ 6.9 & 56.0 $\pm$ 7.1 & 35.3 $\pm$ 6.8 & 37.3 $\pm$ 6.8 & 50.0 $\pm$ 7.1 & 57.1 $\pm$ 7.1 & 42.0 $\pm$ 7.1 & 49.0 $\pm$ 7.1 &  &  \\ 

Gemini-2.5-flash & 54.0 $\pm$ 7.1 & 52.0 $\pm$ 7.1 & 43.1 $\pm$ 7.0 & 37.3 $\pm$ 6.8 & 58.0 $\pm$ 7.1 & 59.2 $\pm$ 7.1 & 60.0 $\pm$ 7.0 & 51.0 $\pm$ 7.1 &  &  \\ 

\Xhline{1.5pt}
\end{tabular}%
}
\caption{Evaluation results of current popular models on ArxivRollBench for Prediction tasks.}
\label{tab:table_p}
\end{table*}

\begin{figure*}[h]
  \centering
  \includegraphics[width=0.99\linewidth]
    {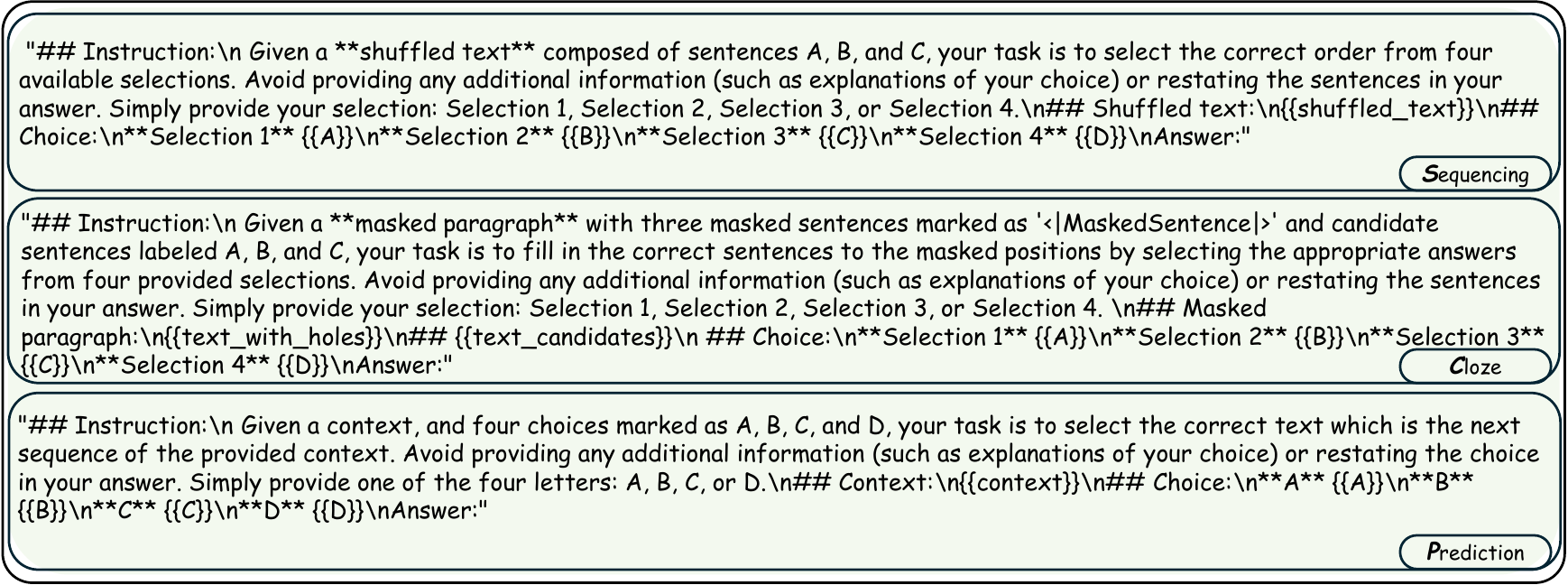}
\caption{\text{The details of private datasets generation}, encompassing three formats: sequencing, cloze, and prediction (SCP).}
    \label{fig:details}
\end{figure*}

\section{A Detailed Description of New Private Benchmarks: ArxivRollBench2025a and ArxivRollBench2026a}
\label{sec:detail-robench-new}

\noindent
\textbf{Private Benchmark Details.}
Following the construction of ArxivRollBench2024b, we further instantiate two new private benchmarks, ArxivRollBench2025a and ArxivRollBench2026a. Each benchmark is built from a fresh six-month snapshot of recent arXiv papers and covers the same eight domains: Computer Science (CS), Economics (Econ), Electrical Engineering (EESS), Mathematics (Math), Physics (Phy), Biology (Bio), Finance (Fin), and Statistics (Stat). The construction pipeline is unchanged from Appendix \ref{sec:detail-robench}: articles are first segmented into text fragments, low-quality fragments are filtered, and the remaining fragments are converted into sequencing, cloze, and prediction questions through SCP.

For these two releases, we use a compact ``-50'' split to support cost-controlled evaluation of both open-source and API models. The split keeps the benchmark broad across all eight domains and all three SCP formats, while limiting the total number of API calls. ArxivRollBench2025a contains 1,107 examples in this split, and ArxivRollBench2026a contains 1,121 examples. Table \ref{tab:new-release-stat} reports the per-domain distribution.

\begin{table*}[t]
\centering
\resizebox{0.95\textwidth}{!}{%
\begin{tabular}{c|lrrrr}
\Xhline{1.5pt}
Benchmark & Domain & Sequencing & Cloze & Prediction & Total \\ \hline

ArxivRollBench2025a & CS & 50 & 44 & 51 & 145 \\

ArxivRollBench2025a & Q-Fin. & 50 & 43 & 51 & 144 \\

ArxivRollBench2025a & Math & 42 & 28 & 51 & 121 \\

ArxivRollBench2025a & Phy. & 44 & 34 & 49 & 127 \\

ArxivRollBench2025a & Stat. & 48 & 42 & 50 & 140 \\

ArxivRollBench2025a & Bio. & 49 & 49 & 50 & 148 \\

ArxivRollBench2025a & Econ. & 48 & 45 & 51 & 144 \\

ArxivRollBench2025a & EESS & 46 & 42 & 50 & 138 \\

\hline

ArxivRollBench2026a & CS & 51 & 38 & 51 & 140 \\

ArxivRollBench2026a & Q-Fin. & 48 & 45 & 51 & 144 \\

ArxivRollBench2026a & Math & 44 & 30 & 51 & 125 \\

ArxivRollBench2026a & Phy. & 47 & 41 & 50 & 138 \\

ArxivRollBench2026a & Stat. & 49 & 46 & 51 & 146 \\

ArxivRollBench2026a & Bio. & 47 & 40 & 48 & 135 \\

ArxivRollBench2026a & Econ. & 48 & 47 & 51 & 146 \\

ArxivRollBench2026a & EESS & 51 & 45 & 51 & 147 \\

\hline

\Xhline{1.5pt}
\end{tabular}%
}
\caption{Sample-number distribution of the compact evaluation split for ArxivRollBench2025a and ArxivRollBench2026a.}
\label{tab:new-release-stat}
\end{table*}

\noindent
\textbf{Scoring Protocol.}
For the supplemental results, we report pooled valid accuracy as the main score. Empty or null responses are excluded from the denominator of the main score, while non-empty invalid-format answers are counted as incorrect. We additionally report raw accuracy, which uses all examples as the denominator, and coverage, which is the fraction of examples receiving a non-empty response. Models with high null-response rates are retained in the ranking but explicitly marked. This protocol makes capability comparisons less sensitive to API transport failures or thinking-mode null outputs, while still exposing low-coverage behavior.

\section{External Experiment Results on ArxivRollBench2025a and ArxivRollBench2026a}
\label{sec:appendix-new-release-results}

Tables \ref{tab:appendix-2025a-overall} and \ref{tab:appendix-2026a-overall} present the overall rankings. Tables \ref{tab:appendix-2025a-s}--\ref{tab:appendix-2026a-p} further provide detailed domain-level results for sequencing, cloze, and prediction tasks.

\begin{table*}[t]
\centering
\resizebox{0.92\textwidth}{!}{%
\begin{tabular}{c|lrrrr}
\Xhline{1.5pt}
Rank & Model & Valid Acc. & Raw Acc. & Coverage & Warning \\ \hline
1 & google/gemini-2.5-flash & 44.3 $\pm$ 1.5 & 44.3 & 100.0 & -- \\
2 & anthropic/claude-3.5-sonnet & 43.1 $\pm$ 1.5 & 43.1 & 100.0 & -- \\
3 & anthropic/claude-3.7-sonnet & 42.9 $\pm$ 1.5 & 42.9 & 100.0 & -- \\
4 & anthropic/claude-opus-4 & 42.7 $\pm$ 1.5 & 42.7 & 100.0 & -- \\
5 & anthropic/claude-sonnet-4 & 41.2 $\pm$ 1.5 & 41.2 & 100.0 & -- \\
6 & deepseek/deepseek-chat-v3-0324 & 40.5 $\pm$ 1.5 & 40.5 & 100.0 & -- \\
7 & google/gemini-2.0-flash-001 & 39.5 $\pm$ 1.5 & 39.5 & 100.0 & -- \\
8 & moonshotai/kimi-k2 & 37.0 $\pm$ 1.5 & 37.0 & 100.0 & -- \\
9 & openai/gpt-4o & 36.2 $\pm$ 1.5 & 36.2 & 100.0 & -- \\
10 & openai/gpt-4 & 35.3 $\pm$ 1.5 & 35.3 & 100.0 & -- \\
11 & nvidia/Llama-3.1-Nemotron-70B-Instruct-HF & 33.0 $\pm$ 0.2 & 33.0 & 100.0 & -- \\
12 & meta-llama/Llama-3.1-70B-Instruct & 32.2 $\pm$ 0.2 & 32.2 & 100.0 & -- \\
13 & meta-llama/Llama-3.3-70B-Instruct & 30.1 $\pm$ 0.2 & 30.1 & 100.0 & -- \\
14 & Qwen/Qwen2.5-7B-Instruct & 29.6 $\pm$ 0.2 & 29.6 & 100.0 & -- \\
15 & Qwen/Qwen2-7B-Instruct & 29.0 $\pm$ 0.2 & 29.0 & 100.0 & -- \\
16 & Qwen/Qwen3-8B & 28.4 $\pm$ 0.2 & 28.4 & 100.0 & -- \\
17 & deepseek/deepseek-chat-v3.1 & 28.4 $\pm$ 1.4 & 28.4 & 100.0 & -- \\
18 & Qwen/Qwen2.5-7B & 28.0 $\pm$ 0.2 & 28.0 & 100.0 & -- \\
19 & meta-llama/Llama-3.1-8B-Instruct & 27.5 $\pm$ 0.2 & 27.5 & 100.0 & -- \\
20 & meta-llama/Llama-3.2-3B & 23.4 $\pm$ 0.1 & 23.4 & 100.0 & -- \\
21 & meta-llama/Meta-Llama-3-8B & 22.6 $\pm$ 0.1 & 22.6 & 100.0 & -- \\
22 & microsoft/phi-2 & 22.3 $\pm$ 0.1 & 22.3 & 100.0 & -- \\
23 & meta-llama/Llama-3.1-8B & 21.8 $\pm$ 0.1 & 21.8 & 100.0 & -- \\
24 & meta-llama/Llama-3.2-1B & 21.7 $\pm$ 0.1 & 21.7 & 100.0 & -- \\
25 & Qwen/Qwen2.5-Math-7B & 21.4 $\pm$ 0.1 & 21.4 & 100.0 & -- \\
26 & microsoft/Phi-4-reasoning-plus & 18.3 $\pm$ 0.1 & 18.3 & 100.0 & -- \\
27 & microsoft/Phi-3.5-mini-instruct & 16.9 $\pm$ 0.1 & 16.9 & 100.0 & -- \\
28 & Qwen/Qwen2.5-72B-Instruct & 13.4 $\pm$ 0.1 & 13.4 & 100.0 & -- \\
29 & microsoft/Phi-3-mini-4k-instruct & 12.8 $\pm$ 0.1 & 12.8 & 100.0 & -- \\
30 & microsoft/phi-1\_5 & 12.5 $\pm$ 0.1 & 12.5 & 100.0 & -- \\
31 & Qwen/Qwen3-32B & 9.2 $\pm$ 0.1 & 9.2 & 100.0 & -- \\
32 & meta-llama/Llama-2-7b-chat-hf & 8.4 $\pm$ 0.1 & 8.4 & 100.0 & -- \\
33 & Qwen/Qwen2.5-Math-7B-Instruct & 7.3 $\pm$ 0.1 & 7.3 & 100.0 & -- \\
34 & Qwen/Qwen3-14B & 7.2 $\pm$ 0.1 & 7.2 & 100.0 & -- \\
35 & microsoft/Phi-4-reasoning & 4.8 $\pm$ 0.1 & 4.8 & 100.0 & -- \\
36 & microsoft/phi-1 & 4.4 $\pm$ 0.1 & 4.4 & 100.0 & -- \\
37 & google/gemini-2.5-pro & 0.2 $\pm$ 0.1 & 0.2 & 100.0 & -- \\
38 & deepseek/deepseek-r1-0528 & 0.0 $\pm$ 0.0 & 0.0 & 100.0 & -- \\
\Xhline{1.5pt}
\end{tabular}%
}
\caption{Overall supplemental evaluation results on ArxivRollBench2025a. Scores are percentages. Valid Acc. is the pooled valid accuracy and is used for ranking.}
\label{tab:appendix-2025a-overall}
\end{table*}

\begin{table*}[t]
\centering
\resizebox{\textwidth}{!}{%
\begin{tabular}{c|rrrrrrrr}
\Xhline{1.5pt}
Model name & \multicolumn{8}{c}{ArxivRollBench-2025a (S)} \\ \hline
 & CS & Q-Fin. & Math & Phy. & Stat. & Bio. & Econ. & EESS \\ \hline
google/gemini-2.5-flash & 50.0 $\pm$ 7.1 & 42.0 $\pm$ 7.1 & 40.5 $\pm$ 7.7 & 54.5 $\pm$ 7.6 & 50.0 $\pm$ 7.3 & 53.1 $\pm$ 7.2 & 45.8 $\pm$ 7.3 & 30.4 $\pm$ 6.9 \\
anthropic/claude-3.5-sonnet & 44.0 $\pm$ 7.1 & 34.0 $\pm$ 6.8 & 47.6 $\pm$ 7.8 & 38.6 $\pm$ 7.4 & 43.8 $\pm$ 7.2 & 44.9 $\pm$ 7.2 & 50.0 $\pm$ 7.3 & 41.3 $\pm$ 7.3 \\
anthropic/claude-3.7-sonnet & 40.0 $\pm$ 7.0 & 40.0 $\pm$ 7.0 & 45.2 $\pm$ 7.8 & 36.4 $\pm$ 7.3 & 43.8 $\pm$ 7.2 & 42.9 $\pm$ 7.1 & 45.8 $\pm$ 7.3 & 37.0 $\pm$ 7.2 \\
anthropic/claude-opus-4 & 52.0 $\pm$ 7.1 & 58.0 $\pm$ 7.1 & 57.1 $\pm$ 7.7 & 59.1 $\pm$ 7.5 & 50.0 $\pm$ 7.3 & 67.3 $\pm$ 6.8 & 64.6 $\pm$ 7.0 & 39.1 $\pm$ 7.3 \\
anthropic/claude-sonnet-4 & 48.0 $\pm$ 7.1 & 50.0 $\pm$ 7.1 & 33.3 $\pm$ 7.4 & 54.5 $\pm$ 7.6 & 39.6 $\pm$ 7.1 & 57.1 $\pm$ 7.1 & 50.0 $\pm$ 7.3 & 32.6 $\pm$ 7.0 \\
deepseek/deepseek-chat-v3-0324 & 46.0 $\pm$ 7.1 & 44.0 $\pm$ 7.1 & 47.6 $\pm$ 7.8 & 50.0 $\pm$ 7.6 & 47.9 $\pm$ 7.3 & 46.9 $\pm$ 7.2 & 45.8 $\pm$ 7.3 & 30.4 $\pm$ 6.9 \\
google/gemini-2.0-flash-001 & 34.0 $\pm$ 6.8 & 28.0 $\pm$ 6.4 & 42.9 $\pm$ 7.7 & 38.6 $\pm$ 7.4 & 41.7 $\pm$ 7.2 & 44.9 $\pm$ 7.2 & 43.8 $\pm$ 7.2 & 26.1 $\pm$ 6.5 \\
moonshotai/kimi-k2 & 36.0 $\pm$ 6.9 & 42.0 $\pm$ 7.1 & 28.6 $\pm$ 7.1 & 38.6 $\pm$ 7.4 & 43.8 $\pm$ 7.2 & 36.7 $\pm$ 7.0 & 41.7 $\pm$ 7.2 & 34.8 $\pm$ 7.1 \\
openai/gpt-4o & 44.0 $\pm$ 7.1 & 42.0 $\pm$ 7.1 & 40.5 $\pm$ 7.7 & 50.0 $\pm$ 7.6 & 54.2 $\pm$ 7.3 & 34.7 $\pm$ 6.9 & 43.8 $\pm$ 7.2 & 23.9 $\pm$ 6.4 \\
openai/gpt-4 & 44.0 $\pm$ 7.1 & 34.0 $\pm$ 6.8 & 38.1 $\pm$ 7.6 & 45.5 $\pm$ 7.6 & 41.7 $\pm$ 7.2 & 32.7 $\pm$ 6.8 & 41.7 $\pm$ 7.2 & 26.1 $\pm$ 6.5 \\
nvidia/Llama-3.1-Nemotron-70B-Instruct-HF & 33.0 $\pm$ 0.7 & 34.2 $\pm$ 1.6 & 31.2 $\pm$ 0.7 & 32.8 $\pm$ 0.7 & 33.6 $\pm$ 0.7 & 33.9 $\pm$ 1.2 & 34.0 $\pm$ 1.5 & 34.2 $\pm$ 0.7 \\
meta-llama/Llama-3.1-70B-Instruct & 31.3 $\pm$ 0.7 & 33.4 $\pm$ 1.6 & 29.7 $\pm$ 0.6 & 31.3 $\pm$ 0.7 & 31.6 $\pm$ 0.7 & 32.9 $\pm$ 1.2 & 32.2 $\pm$ 1.5 & 32.7 $\pm$ 0.7 \\
meta-llama/Llama-3.3-70B-Instruct & 36.9 $\pm$ 0.7 & 36.9 $\pm$ 1.7 & 35.7 $\pm$ 0.7 & 37.1 $\pm$ 0.7 & 37.7 $\pm$ 0.7 & 38.8 $\pm$ 1.2 & 37.1 $\pm$ 1.6 & 38.2 $\pm$ 0.7 \\
Qwen/Qwen2.5-7B-Instruct & 27.7 $\pm$ 0.6 & 28.0 $\pm$ 1.5 & 27.5 $\pm$ 0.6 & 27.9 $\pm$ 0.6 & 28.5 $\pm$ 0.6 & 26.3 $\pm$ 1.1 & 30.5 $\pm$ 1.5 & 29.7 $\pm$ 0.6 \\
Qwen/Qwen2-7B-Instruct & 27.3 $\pm$ 0.6 & 25.5 $\pm$ 1.5 & 27.6 $\pm$ 0.6 & 27.8 $\pm$ 0.6 & 27.2 $\pm$ 0.6 & 29.0 $\pm$ 1.1 & 27.8 $\pm$ 1.5 & 27.4 $\pm$ 0.6 \\
Qwen/Qwen3-8B & 30.5 $\pm$ 0.7 & 32.3 $\pm$ 1.6 & 30.3 $\pm$ 0.6 & 31.1 $\pm$ 0.7 & 29.6 $\pm$ 0.6 & 31.2 $\pm$ 1.2 & 29.1 $\pm$ 1.5 & 30.4 $\pm$ 0.7 \\
deepseek/deepseek-chat-v3.1 & 26.0 $\pm$ 6.3 & 22.0 $\pm$ 5.9 & 38.1 $\pm$ 7.6 & 29.5 $\pm$ 7.0 & 31.2 $\pm$ 6.8 & 34.7 $\pm$ 6.9 & 41.7 $\pm$ 7.2 & 15.2 $\pm$ 5.4 \\
Qwen/Qwen2.5-7B & 25.1 $\pm$ 0.6 & 24.0 $\pm$ 1.5 & 23.6 $\pm$ 0.6 & 23.9 $\pm$ 0.6 & 24.7 $\pm$ 0.6 & 23.7 $\pm$ 1.1 & 25.3 $\pm$ 1.4 & 25.6 $\pm$ 0.6 \\
meta-llama/Llama-3.1-8B-Instruct & 27.3 $\pm$ 0.6 & 26.8 $\pm$ 1.5 & 28.5 $\pm$ 0.6 & 27.9 $\pm$ 0.6 & 26.5 $\pm$ 0.6 & 28.4 $\pm$ 1.1 & 26.0 $\pm$ 1.4 & 26.2 $\pm$ 0.6 \\
meta-llama/Llama-3.2-3B & 22.7 $\pm$ 0.6 & 20.3 $\pm$ 1.4 & 21.6 $\pm$ 0.6 & 22.6 $\pm$ 0.6 & 22.0 $\pm$ 0.6 & 22.8 $\pm$ 1.1 & 24.1 $\pm$ 1.4 & 22.9 $\pm$ 0.6 \\
meta-llama/Meta-Llama-3-8B & 23.0 $\pm$ 0.6 & 24.9 $\pm$ 1.5 & 23.5 $\pm$ 0.6 & 24.3 $\pm$ 0.6 & 23.0 $\pm$ 0.6 & 24.5 $\pm$ 1.1 & 21.7 $\pm$ 1.3 & 23.0 $\pm$ 0.6 \\
microsoft/phi-2 & 24.2 $\pm$ 0.6 & 22.2 $\pm$ 1.4 & 24.5 $\pm$ 0.6 & 24.0 $\pm$ 0.6 & 24.6 $\pm$ 0.6 & 22.5 $\pm$ 1.0 & 26.3 $\pm$ 1.4 & 24.8 $\pm$ 0.6 \\
meta-llama/Llama-3.1-8B & 24.9 $\pm$ 0.6 & 24.1 $\pm$ 1.5 & 25.6 $\pm$ 0.6 & 25.5 $\pm$ 0.6 & 24.6 $\pm$ 0.6 & 25.5 $\pm$ 1.1 & 26.3 $\pm$ 1.4 & 24.5 $\pm$ 0.6 \\
meta-llama/Llama-3.2-1B & 24.9 $\pm$ 0.6 & 23.7 $\pm$ 1.5 & 25.1 $\pm$ 0.6 & 25.0 $\pm$ 0.6 & 25.6 $\pm$ 0.6 & 23.6 $\pm$ 1.1 & 28.5 $\pm$ 1.5 & 26.0 $\pm$ 0.6 \\
Qwen/Qwen2.5-Math-7B & 16.8 $\pm$ 0.5 & 17.4 $\pm$ 1.3 & 18.9 $\pm$ 0.6 & 17.4 $\pm$ 0.5 & 17.6 $\pm$ 0.5 & 17.0 $\pm$ 0.9 & 18.4 $\pm$ 1.3 & 16.4 $\pm$ 0.5 \\
microsoft/Phi-4-reasoning-plus & 8.7 $\pm$ 0.4 & 12.7 $\pm$ 1.1 & 11.9 $\pm$ 0.5 & 9.0 $\pm$ 0.4 & 9.6 $\pm$ 0.4 & 11.1 $\pm$ 0.8 & 10.0 $\pm$ 1.0 & 9.4 $\pm$ 0.4 \\
microsoft/Phi-3.5-mini-instruct & 19.2 $\pm$ 0.6 & 20.1 $\pm$ 1.4 & 18.8 $\pm$ 0.6 & 18.9 $\pm$ 0.6 & 18.9 $\pm$ 0.6 & 21.7 $\pm$ 1.0 & 19.7 $\pm$ 1.3 & 19.2 $\pm$ 0.6 \\
Qwen/Qwen2.5-72B-Instruct & 21.7 $\pm$ 0.6 & 24.1 $\pm$ 1.5 & 19.5 $\pm$ 0.6 & 20.5 $\pm$ 0.6 & 20.8 $\pm$ 0.6 & 24.0 $\pm$ 1.1 & 20.9 $\pm$ 1.3 & 21.7 $\pm$ 0.6 \\
microsoft/Phi-3-mini-4k-instruct & 6.6 $\pm$ 0.4 & 4.9 $\pm$ 0.7 & 5.7 $\pm$ 0.3 & 4.3 $\pm$ 0.3 & 6.5 $\pm$ 0.3 & 5.9 $\pm$ 0.6 & 5.6 $\pm$ 0.7 & 7.0 $\pm$ 0.4 \\
microsoft/phi-1\_5 & 22.6 $\pm$ 0.6 & 21.5 $\pm$ 1.4 & 24.2 $\pm$ 0.6 & 22.9 $\pm$ 0.6 & 24.2 $\pm$ 0.6 & 21.8 $\pm$ 1.0 & 26.2 $\pm$ 1.4 & 23.3 $\pm$ 0.6 \\
Qwen/Qwen3-32B & 19.8 $\pm$ 0.6 & 23.6 $\pm$ 1.5 & 23.1 $\pm$ 0.6 & 18.5 $\pm$ 0.5 & 19.9 $\pm$ 0.6 & 18.6 $\pm$ 1.0 & 19.3 $\pm$ 1.3 & 21.2 $\pm$ 0.6 \\
meta-llama/Llama-2-7b-chat-hf & 8.0 $\pm$ 0.4 & 8.7 $\pm$ 1.0 & 8.1 $\pm$ 0.4 & 6.4 $\pm$ 0.3 & 7.3 $\pm$ 0.4 & 6.4 $\pm$ 0.6 & 11.0 $\pm$ 1.0 & 7.6 $\pm$ 0.4 \\
Qwen/Qwen2.5-Math-7B-Instruct & 5.5 $\pm$ 0.3 & 6.1 $\pm$ 0.8 & 4.2 $\pm$ 0.3 & 4.5 $\pm$ 0.3 & 6.9 $\pm$ 0.4 & 5.4 $\pm$ 0.6 & 6.9 $\pm$ 0.8 & 5.7 $\pm$ 0.3 \\
Qwen/Qwen3-14B & 4.8 $\pm$ 0.3 & 5.6 $\pm$ 0.8 & 6.6 $\pm$ 0.3 & 5.8 $\pm$ 0.3 & 5.0 $\pm$ 0.3 & 6.8 $\pm$ 0.6 & 5.6 $\pm$ 0.7 & 5.5 $\pm$ 0.3 \\
microsoft/Phi-4-reasoning & 1.4 $\pm$ 0.2 & 1.8 $\pm$ 0.5 & 3.1 $\pm$ 0.2 & 1.1 $\pm$ 0.1 & 2.0 $\pm$ 0.2 & 1.6 $\pm$ 0.3 & 2.5 $\pm$ 0.5 & 1.4 $\pm$ 0.2 \\
microsoft/phi-1 & 6.2 $\pm$ 0.3 & 5.8 $\pm$ 0.8 & 7.0 $\pm$ 0.4 & 6.8 $\pm$ 0.4 & 8.2 $\pm$ 0.4 & 6.1 $\pm$ 0.6 & 6.6 $\pm$ 0.8 & 7.4 $\pm$ 0.4 \\
google/gemini-2.5-pro & 0.0 $\pm$ 0.0 & 0.0 $\pm$ 0.0 & 0.0 $\pm$ 0.0 & 0.0 $\pm$ 0.0 & 0.0 $\pm$ 0.0 & 0.0 $\pm$ 0.0 & 0.0 $\pm$ 0.0 & 0.0 $\pm$ 0.0 \\
deepseek/deepseek-r1-0528 & 0.0 $\pm$ 0.0 & 0.0 $\pm$ 0.0 & 0.0 $\pm$ 0.0 & 0.0 $\pm$ 0.0 & 0.0 $\pm$ 0.0 & 0.0 $\pm$ 0.0 & 0.0 $\pm$ 0.0 & 0.0 $\pm$ 0.0 \\
\Xhline{1.5pt}
\end{tabular}%
}
\caption{Detailed evaluation results of current popular models on ArxivRollBench2025a for sequencing tasks. Scores are percentages and use the same valid-response scoring protocol as the overall ranking.}
\label{tab:appendix-2025a-s}
\end{table*}

\begin{table*}[t]
\centering
\resizebox{\textwidth}{!}{%
\begin{tabular}{c|rrrrrrrr}
\Xhline{1.5pt}
Model name & \multicolumn{8}{c}{ArxivRollBench-2025a (C)} \\ \hline
 & CS & Q-Fin. & Math & Phy. & Stat. & Bio. & Econ. & EESS \\ \hline
google/gemini-2.5-flash & 31.8 $\pm$ 7.1 & 32.6 $\pm$ 7.2 & 28.6 $\pm$ 8.7 & 23.5 $\pm$ 7.4 & 14.3 $\pm$ 5.5 & 28.6 $\pm$ 6.5 & 26.7 $\pm$ 6.7 & 35.7 $\pm$ 7.5 \\
anthropic/claude-3.5-sonnet & 40.9 $\pm$ 7.5 & 37.2 $\pm$ 7.5 & 28.6 $\pm$ 8.7 & 23.5 $\pm$ 7.4 & 21.4 $\pm$ 6.4 & 22.4 $\pm$ 6.0 & 20.0 $\pm$ 6.0 & 28.6 $\pm$ 7.1 \\
anthropic/claude-3.7-sonnet & 31.8 $\pm$ 7.1 & 27.9 $\pm$ 6.9 & 32.1 $\pm$ 9.0 & 38.2 $\pm$ 8.5 & 26.2 $\pm$ 6.9 & 24.5 $\pm$ 6.2 & 22.2 $\pm$ 6.3 & 21.4 $\pm$ 6.4 \\
anthropic/claude-opus-4 & 9.1 $\pm$ 4.4 & 7.0 $\pm$ 3.9 & 0.0 $\pm$ 0.0 & 2.9 $\pm$ 2.9 & 2.4 $\pm$ 2.4 & 0.0 $\pm$ 0.0 & 8.9 $\pm$ 4.3 & 4.8 $\pm$ 3.3 \\
anthropic/claude-sonnet-4 & 15.9 $\pm$ 5.6 & 18.6 $\pm$ 6.0 & 7.1 $\pm$ 5.0 & 8.8 $\pm$ 4.9 & 9.5 $\pm$ 4.6 & 20.4 $\pm$ 5.8 & 11.1 $\pm$ 4.7 & 11.9 $\pm$ 5.1 \\
deepseek/deepseek-chat-v3-0324 & 29.5 $\pm$ 7.0 & 30.2 $\pm$ 7.1 & 32.1 $\pm$ 9.0 & 20.6 $\pm$ 7.0 & 23.8 $\pm$ 6.7 & 10.2 $\pm$ 4.4 & 31.1 $\pm$ 7.0 & 23.8 $\pm$ 6.7 \\
google/gemini-2.0-flash-001 & 34.1 $\pm$ 7.2 & 34.9 $\pm$ 7.4 & 17.9 $\pm$ 7.4 & 17.6 $\pm$ 6.6 & 26.2 $\pm$ 6.9 & 20.4 $\pm$ 5.8 & 17.8 $\pm$ 5.8 & 31.0 $\pm$ 7.2 \\
moonshotai/kimi-k2 & 31.8 $\pm$ 7.1 & 27.9 $\pm$ 6.9 & 21.4 $\pm$ 7.9 & 20.6 $\pm$ 7.0 & 23.8 $\pm$ 6.7 & 26.5 $\pm$ 6.4 & 31.1 $\pm$ 7.0 & 42.9 $\pm$ 7.7 \\
openai/gpt-4o & 20.5 $\pm$ 6.2 & 20.9 $\pm$ 6.3 & 21.4 $\pm$ 7.9 & 8.8 $\pm$ 4.9 & 7.1 $\pm$ 4.0 & 22.4 $\pm$ 6.0 & 13.3 $\pm$ 5.1 & 19.0 $\pm$ 6.1 \\
openai/gpt-4 & 25.0 $\pm$ 6.6 & 20.9 $\pm$ 6.3 & 17.9 $\pm$ 7.4 & 20.6 $\pm$ 7.0 & 11.9 $\pm$ 5.1 & 22.4 $\pm$ 6.0 & 28.9 $\pm$ 6.8 & 21.4 $\pm$ 6.4 \\
nvidia/Llama-3.1-Nemotron-70B-Instruct-HF & 28.0 $\pm$ 0.6 & 27.0 $\pm$ 1.6 & 27.4 $\pm$ 0.6 & 28.2 $\pm$ 0.6 & 28.3 $\pm$ 0.7 & 28.5 $\pm$ 1.2 & 24.5 $\pm$ 1.5 & 28.2 $\pm$ 0.6 \\
meta-llama/Llama-3.1-70B-Instruct & 28.2 $\pm$ 0.6 & 28.9 $\pm$ 1.6 & 27.4 $\pm$ 0.6 & 28.0 $\pm$ 0.6 & 28.2 $\pm$ 0.7 & 28.7 $\pm$ 1.2 & 24.1 $\pm$ 1.5 & 28.3 $\pm$ 0.6 \\
meta-llama/Llama-3.3-70B-Instruct & 14.2 $\pm$ 0.5 & 13.9 $\pm$ 1.3 & 15.5 $\pm$ 0.5 & 14.3 $\pm$ 0.5 & 14.0 $\pm$ 0.5 & 14.8 $\pm$ 0.9 & 10.3 $\pm$ 1.1 & 14.5 $\pm$ 0.5 \\
Qwen/Qwen2.5-7B-Instruct & 22.5 $\pm$ 0.6 & 25.7 $\pm$ 1.6 & 24.4 $\pm$ 0.6 & 23.9 $\pm$ 0.6 & 23.1 $\pm$ 0.6 & 24.3 $\pm$ 1.1 & 24.7 $\pm$ 1.5 & 23.0 $\pm$ 0.6 \\
Qwen/Qwen2-7B-Instruct & 27.2 $\pm$ 0.6 & 25.7 $\pm$ 1.6 & 26.6 $\pm$ 0.6 & 26.3 $\pm$ 0.6 & 26.3 $\pm$ 0.7 & 24.3 $\pm$ 1.1 & 24.4 $\pm$ 1.5 & 25.3 $\pm$ 0.6 \\
Qwen/Qwen3-8B & 24.4 $\pm$ 0.6 & 23.2 $\pm$ 1.5 & 24.9 $\pm$ 0.6 & 25.7 $\pm$ 0.6 & 24.6 $\pm$ 0.7 & 24.3 $\pm$ 1.1 & 25.4 $\pm$ 1.5 & 25.0 $\pm$ 0.6 \\
deepseek/deepseek-chat-v3.1 & 18.2 $\pm$ 5.9 & 16.3 $\pm$ 5.7 & 28.6 $\pm$ 8.7 & 14.7 $\pm$ 6.2 & 19.0 $\pm$ 6.1 & 6.1 $\pm$ 3.5 & 20.0 $\pm$ 6.0 & 16.7 $\pm$ 5.8 \\
Qwen/Qwen2.5-7B & 26.8 $\pm$ 0.6 & 26.6 $\pm$ 1.6 & 28.2 $\pm$ 0.6 & 27.9 $\pm$ 0.6 & 27.1 $\pm$ 0.7 & 28.6 $\pm$ 1.2 & 26.8 $\pm$ 1.5 & 27.5 $\pm$ 0.6 \\
meta-llama/Llama-3.1-8B-Instruct & 27.1 $\pm$ 0.6 & 23.1 $\pm$ 1.5 & 25.3 $\pm$ 0.6 & 26.7 $\pm$ 0.6 & 26.1 $\pm$ 0.7 & 26.3 $\pm$ 1.2 & 22.1 $\pm$ 1.4 & 27.2 $\pm$ 0.6 \\
meta-llama/Llama-3.2-3B & 25.3 $\pm$ 0.6 & 23.1 $\pm$ 1.5 & 24.9 $\pm$ 0.6 & 25.6 $\pm$ 0.6 & 26.5 $\pm$ 0.7 & 25.2 $\pm$ 1.1 & 25.4 $\pm$ 1.5 & 26.2 $\pm$ 0.6 \\
meta-llama/Meta-Llama-3-8B & 18.8 $\pm$ 0.6 & 18.7 $\pm$ 1.4 & 20.1 $\pm$ 0.6 & 20.1 $\pm$ 0.6 & 20.6 $\pm$ 0.6 & 19.2 $\pm$ 1.0 & 16.3 $\pm$ 1.3 & 20.5 $\pm$ 0.6 \\
microsoft/phi-2 & 17.3 $\pm$ 0.5 & 20.8 $\pm$ 1.5 & 20.7 $\pm$ 0.6 & 20.1 $\pm$ 0.6 & 18.4 $\pm$ 0.6 & 19.1 $\pm$ 1.0 & 18.7 $\pm$ 1.4 & 18.0 $\pm$ 0.5 \\
meta-llama/Llama-3.1-8B & 15.1 $\pm$ 0.5 & 14.1 $\pm$ 1.3 & 18.1 $\pm$ 0.5 & 17.2 $\pm$ 0.5 & 17.0 $\pm$ 0.6 & 15.9 $\pm$ 1.0 & 12.3 $\pm$ 1.1 & 15.9 $\pm$ 0.5 \\
meta-llama/Llama-3.2-1B & 15.4 $\pm$ 0.5 & 12.0 $\pm$ 1.2 & 13.6 $\pm$ 0.5 & 11.2 $\pm$ 0.4 & 17.2 $\pm$ 0.6 & 14.1 $\pm$ 0.9 & 15.0 $\pm$ 1.2 & 16.2 $\pm$ 0.5 \\
Qwen/Qwen2.5-Math-7B & 24.9 $\pm$ 0.6 & 22.0 $\pm$ 1.5 & 24.1 $\pm$ 0.6 & 25.0 $\pm$ 0.6 & 25.0 $\pm$ 0.7 & 24.7 $\pm$ 1.1 & 21.9 $\pm$ 1.4 & 25.4 $\pm$ 0.6 \\
microsoft/Phi-4-reasoning-plus & 23.2 $\pm$ 0.6 & 24.4 $\pm$ 1.6 & 22.6 $\pm$ 0.6 & 22.9 $\pm$ 0.6 & 23.2 $\pm$ 0.6 & 21.8 $\pm$ 1.1 & 21.6 $\pm$ 1.4 & 24.1 $\pm$ 0.6 \\
microsoft/Phi-3.5-mini-instruct & 28.7 $\pm$ 0.6 & 23.5 $\pm$ 1.5 & 26.5 $\pm$ 0.6 & 28.1 $\pm$ 0.6 & 27.1 $\pm$ 0.7 & 26.5 $\pm$ 1.2 & 25.4 $\pm$ 1.5 & 28.8 $\pm$ 0.6 \\
Qwen/Qwen2.5-72B-Instruct & 14.7 $\pm$ 0.5 & 17.0 $\pm$ 1.4 & 15.2 $\pm$ 0.5 & 12.6 $\pm$ 0.5 & 16.8 $\pm$ 0.6 & 13.6 $\pm$ 0.9 & 16.2 $\pm$ 1.3 & 13.8 $\pm$ 0.5 \\
microsoft/Phi-3-mini-4k-instruct & 22.1 $\pm$ 0.6 & 19.3 $\pm$ 1.4 & 19.7 $\pm$ 0.6 & 18.6 $\pm$ 0.5 & 21.4 $\pm$ 0.6 & 21.8 $\pm$ 1.1 & 20.9 $\pm$ 1.4 & 22.2 $\pm$ 0.6 \\
microsoft/phi-1\_5 & 11.6 $\pm$ 0.5 & 10.9 $\pm$ 1.1 & 12.6 $\pm$ 0.5 & 8.8 $\pm$ 0.4 & 14.5 $\pm$ 0.5 & 12.5 $\pm$ 0.9 & 12.2 $\pm$ 1.1 & 11.6 $\pm$ 0.5 \\
Qwen/Qwen3-32B & 3.5 $\pm$ 0.3 & 4.2 $\pm$ 0.7 & 5.0 $\pm$ 0.3 & 3.0 $\pm$ 0.2 & 4.5 $\pm$ 0.3 & 3.0 $\pm$ 0.5 & 2.9 $\pm$ 0.6 & 3.8 $\pm$ 0.3 \\
meta-llama/Llama-2-7b-chat-hf & 2.5 $\pm$ 0.2 & 2.2 $\pm$ 0.5 & 5.2 $\pm$ 0.3 & 1.3 $\pm$ 0.2 & 2.5 $\pm$ 0.2 & 1.8 $\pm$ 0.4 & 1.6 $\pm$ 0.4 & 2.7 $\pm$ 0.2 \\
Qwen/Qwen2.5-Math-7B-Instruct & 1.4 $\pm$ 0.2 & 1.6 $\pm$ 0.5 & 1.8 $\pm$ 0.2 & 1.9 $\pm$ 0.2 & 1.4 $\pm$ 0.2 & 1.1 $\pm$ 0.3 & 1.6 $\pm$ 0.4 & 1.4 $\pm$ 0.2 \\
Qwen/Qwen3-14B & 3.4 $\pm$ 0.3 & 4.5 $\pm$ 0.8 & 7.4 $\pm$ 0.4 & 6.3 $\pm$ 0.3 & 4.4 $\pm$ 0.3 & 5.8 $\pm$ 0.6 & 4.8 $\pm$ 0.7 & 3.1 $\pm$ 0.2 \\
microsoft/Phi-4-reasoning & 9.5 $\pm$ 0.4 & 14.9 $\pm$ 1.3 & 15.5 $\pm$ 0.5 & 11.6 $\pm$ 0.5 & 12.9 $\pm$ 0.5 & 11.6 $\pm$ 0.8 & 14.5 $\pm$ 1.2 & 12.2 $\pm$ 0.5 \\
microsoft/phi-1 & 1.7 $\pm$ 0.2 & 2.2 $\pm$ 0.5 & 1.8 $\pm$ 0.2 & 1.8 $\pm$ 0.2 & 1.9 $\pm$ 0.2 & 1.7 $\pm$ 0.3 & 1.5 $\pm$ 0.4 & 1.7 $\pm$ 0.2 \\
google/gemini-2.5-pro & 0.0 $\pm$ 0.0 & 0.0 $\pm$ 0.0 & 0.0 $\pm$ 0.0 & 0.0 $\pm$ 0.0 & 0.0 $\pm$ 0.0 & 0.0 $\pm$ 0.0 & 0.0 $\pm$ 0.0 & 0.0 $\pm$ 0.0 \\
deepseek/deepseek-r1-0528 & 0.0 $\pm$ 0.0 & 0.0 $\pm$ 0.0 & 0.0 $\pm$ 0.0 & 0.0 $\pm$ 0.0 & 0.0 $\pm$ 0.0 & 0.0 $\pm$ 0.0 & 0.0 $\pm$ 0.0 & 0.0 $\pm$ 0.0 \\
\Xhline{1.5pt}
\end{tabular}%
}
\caption{Detailed evaluation results of current popular models on ArxivRollBench2025a for cloze tasks. Scores are percentages and use the same valid-response scoring protocol as the overall ranking.}
\label{tab:appendix-2025a-c}
\end{table*}

\begin{table*}[t]
\centering
\resizebox{\textwidth}{!}{%
\begin{tabular}{c|rrrrrrrr}
\Xhline{1.5pt}
Model name & \multicolumn{8}{c}{ArxivRollBench-2025a (P)} \\ \hline
 & CS & Q-Fin. & Math & Phy. & Stat. & Bio. & Econ. & EESS \\ \hline
google/gemini-2.5-flash & 43.1 $\pm$ 7.0 & 64.7 $\pm$ 6.8 & 54.9 $\pm$ 7.0 & 55.1 $\pm$ 7.2 & 62.0 $\pm$ 6.9 & 56.0 $\pm$ 7.1 & 62.7 $\pm$ 6.8 & 50.0 $\pm$ 7.1 \\
anthropic/claude-3.5-sonnet & 47.1 $\pm$ 7.1 & 66.7 $\pm$ 6.7 & 52.9 $\pm$ 7.1 & 42.9 $\pm$ 7.1 & 60.0 $\pm$ 7.0 & 54.0 $\pm$ 7.1 & 66.7 $\pm$ 6.7 & 54.0 $\pm$ 7.1 \\
anthropic/claude-3.7-sonnet & 49.0 $\pm$ 7.1 & 66.7 $\pm$ 6.7 & 52.9 $\pm$ 7.1 & 38.8 $\pm$ 7.0 & 66.0 $\pm$ 6.8 & 62.0 $\pm$ 6.9 & 64.7 $\pm$ 6.8 & 54.0 $\pm$ 7.1 \\
anthropic/claude-opus-4 & 62.7 $\pm$ 6.8 & 62.7 $\pm$ 6.8 & 56.9 $\pm$ 7.0 & 59.2 $\pm$ 7.1 & 72.0 $\pm$ 6.4 & 74.0 $\pm$ 6.3 & 60.8 $\pm$ 6.9 & 62.0 $\pm$ 6.9 \\
anthropic/claude-sonnet-4 & 56.9 $\pm$ 7.0 & 58.8 $\pm$ 7.0 & 49.0 $\pm$ 7.1 & 53.1 $\pm$ 7.2 & 64.0 $\pm$ 6.9 & 70.0 $\pm$ 6.5 & 64.7 $\pm$ 6.8 & 58.0 $\pm$ 7.1 \\
deepseek/deepseek-chat-v3-0324 & 47.1 $\pm$ 7.1 & 49.0 $\pm$ 7.1 & 45.1 $\pm$ 7.0 & 57.1 $\pm$ 7.1 & 54.0 $\pm$ 7.1 & 44.0 $\pm$ 7.1 & 54.9 $\pm$ 7.0 & 42.0 $\pm$ 7.1 \\
google/gemini-2.0-flash-001 & 51.0 $\pm$ 7.1 & 56.9 $\pm$ 7.0 & 52.9 $\pm$ 7.1 & 57.1 $\pm$ 7.1 & 48.0 $\pm$ 7.1 & 62.0 $\pm$ 6.9 & 54.9 $\pm$ 7.0 & 40.0 $\pm$ 7.0 \\
moonshotai/kimi-k2 & 39.2 $\pm$ 6.9 & 43.1 $\pm$ 7.0 & 41.2 $\pm$ 7.0 & 53.1 $\pm$ 7.2 & 36.0 $\pm$ 6.9 & 44.0 $\pm$ 7.1 & 45.1 $\pm$ 7.0 & 40.0 $\pm$ 7.0 \\
openai/gpt-4o & 37.3 $\pm$ 6.8 & 47.1 $\pm$ 7.1 & 45.1 $\pm$ 7.0 & 57.1 $\pm$ 7.1 & 52.0 $\pm$ 7.1 & 50.0 $\pm$ 7.1 & 47.1 $\pm$ 7.1 & 38.0 $\pm$ 6.9 \\
openai/gpt-4 & 33.3 $\pm$ 6.7 & 43.1 $\pm$ 7.0 & 41.2 $\pm$ 7.0 & 57.1 $\pm$ 7.1 & 46.0 $\pm$ 7.1 & 42.0 $\pm$ 7.1 & 52.9 $\pm$ 7.1 & 38.0 $\pm$ 6.9 \\
nvidia/Llama-3.1-Nemotron-70B-Instruct-HF & 38.6 $\pm$ 0.7 & 39.1 $\pm$ 1.6 & 35.0 $\pm$ 0.7 & 37.2 $\pm$ 0.7 & 38.1 $\pm$ 0.7 & 37.4 $\pm$ 1.2 & 40.2 $\pm$ 1.6 & 39.7 $\pm$ 0.7 \\
meta-llama/Llama-3.1-70B-Instruct & 38.4 $\pm$ 0.7 & 37.7 $\pm$ 1.6 & 34.1 $\pm$ 0.7 & 36.0 $\pm$ 0.7 & 37.4 $\pm$ 0.7 & 37.3 $\pm$ 1.2 & 38.4 $\pm$ 1.6 & 38.1 $\pm$ 0.7 \\
meta-llama/Llama-3.3-70B-Instruct & 39.1 $\pm$ 0.7 & 40.5 $\pm$ 1.7 & 34.9 $\pm$ 0.7 & 37.8 $\pm$ 0.7 & 37.9 $\pm$ 0.7 & 37.0 $\pm$ 1.2 & 41.2 $\pm$ 1.6 & 39.7 $\pm$ 0.7 \\
Qwen/Qwen2.5-7B-Instruct & 37.0 $\pm$ 0.7 & 38.0 $\pm$ 1.6 & 36.0 $\pm$ 0.7 & 36.7 $\pm$ 0.7 & 37.9 $\pm$ 0.7 & 36.5 $\pm$ 1.2 & 36.5 $\pm$ 1.5 & 37.0 $\pm$ 0.7 \\
Qwen/Qwen2-7B-Instruct & 33.6 $\pm$ 0.7 & 35.0 $\pm$ 1.6 & 31.1 $\pm$ 0.7 & 32.5 $\pm$ 0.7 & 34.5 $\pm$ 0.7 & 32.7 $\pm$ 1.1 & 35.5 $\pm$ 1.5 & 33.4 $\pm$ 0.7 \\
Qwen/Qwen3-8B & 30.8 $\pm$ 0.7 & 32.9 $\pm$ 1.6 & 27.0 $\pm$ 0.6 & 28.7 $\pm$ 0.6 & 29.5 $\pm$ 0.6 & 30.8 $\pm$ 1.1 & 32.0 $\pm$ 1.5 & 30.9 $\pm$ 0.7 \\
deepseek/deepseek-chat-v3.1 & 27.5 $\pm$ 6.3 & 33.3 $\pm$ 6.7 & 35.3 $\pm$ 6.8 & 40.8 $\pm$ 7.1 & 44.0 $\pm$ 7.1 & 38.0 $\pm$ 6.9 & 45.1 $\pm$ 7.0 & 28.0 $\pm$ 6.4 \\
Qwen/Qwen2.5-7B & 32.2 $\pm$ 0.7 & 32.6 $\pm$ 1.6 & 31.0 $\pm$ 0.7 & 31.7 $\pm$ 0.7 & 32.8 $\pm$ 0.7 & 31.0 $\pm$ 1.1 & 32.4 $\pm$ 1.5 & 31.7 $\pm$ 0.7 \\
meta-llama/Llama-3.1-8B-Instruct & 29.3 $\pm$ 0.6 & 32.6 $\pm$ 1.6 & 26.8 $\pm$ 0.6 & 29.1 $\pm$ 0.6 & 29.3 $\pm$ 0.6 & 30.3 $\pm$ 1.1 & 30.5 $\pm$ 1.5 & 29.5 $\pm$ 0.6 \\
meta-llama/Llama-3.2-3B & 23.0 $\pm$ 0.6 & 22.3 $\pm$ 1.4 & 22.8 $\pm$ 0.6 & 22.7 $\pm$ 0.6 & 21.8 $\pm$ 0.6 & 23.8 $\pm$ 1.0 & 21.2 $\pm$ 1.3 & 21.7 $\pm$ 0.6 \\
meta-llama/Meta-Llama-3-8B & 24.0 $\pm$ 0.6 & 25.0 $\pm$ 1.5 & 24.8 $\pm$ 0.6 & 24.3 $\pm$ 0.6 & 24.9 $\pm$ 0.6 & 24.0 $\pm$ 1.0 & 22.7 $\pm$ 1.3 & 24.2 $\pm$ 0.6 \\
microsoft/phi-2 & 23.8 $\pm$ 0.6 & 23.5 $\pm$ 1.4 & 22.1 $\pm$ 0.6 & 23.5 $\pm$ 0.6 & 23.9 $\pm$ 0.6 & 24.4 $\pm$ 1.1 & 23.8 $\pm$ 1.4 & 23.3 $\pm$ 0.6 \\
meta-llama/Llama-3.1-8B & 23.3 $\pm$ 0.6 & 22.4 $\pm$ 1.4 & 24.6 $\pm$ 0.6 & 23.7 $\pm$ 0.6 & 24.2 $\pm$ 0.6 & 23.1 $\pm$ 1.0 & 24.2 $\pm$ 1.4 & 23.1 $\pm$ 0.6 \\
meta-llama/Llama-3.2-1B & 24.9 $\pm$ 0.6 & 24.8 $\pm$ 1.5 & 24.5 $\pm$ 0.6 & 25.3 $\pm$ 0.6 & 25.4 $\pm$ 0.6 & 24.7 $\pm$ 1.1 & 26.7 $\pm$ 1.4 & 24.5 $\pm$ 0.6 \\
Qwen/Qwen2.5-Math-7B & 22.5 $\pm$ 0.6 & 21.3 $\pm$ 1.4 & 21.3 $\pm$ 0.6 & 22.2 $\pm$ 0.6 & 22.8 $\pm$ 0.6 & 22.5 $\pm$ 1.0 & 22.6 $\pm$ 1.3 & 22.9 $\pm$ 0.6 \\
microsoft/Phi-4-reasoning-plus & 21.6 $\pm$ 0.6 & 25.1 $\pm$ 1.5 & 19.0 $\pm$ 0.6 & 21.9 $\pm$ 0.6 & 22.2 $\pm$ 0.6 & 24.3 $\pm$ 1.1 & 24.6 $\pm$ 1.4 & 24.3 $\pm$ 0.6 \\
microsoft/Phi-3.5-mini-instruct & 3.5 $\pm$ 0.3 & 2.7 $\pm$ 0.5 & 6.3 $\pm$ 0.3 & 5.2 $\pm$ 0.3 & 4.1 $\pm$ 0.3 & 3.0 $\pm$ 0.4 & 3.5 $\pm$ 0.6 & 3.7 $\pm$ 0.3 \\
Qwen/Qwen2.5-72B-Instruct & 3.8 $\pm$ 0.3 & 5.5 $\pm$ 0.8 & 5.4 $\pm$ 0.3 & 5.5 $\pm$ 0.3 & 4.2 $\pm$ 0.3 & 5.0 $\pm$ 0.5 & 5.7 $\pm$ 0.7 & 3.0 $\pm$ 0.2 \\
microsoft/Phi-3-mini-4k-instruct & 11.8 $\pm$ 0.5 & 12.4 $\pm$ 1.1 & 12.8 $\pm$ 0.5 & 12.0 $\pm$ 0.5 & 12.4 $\pm$ 0.5 & 12.0 $\pm$ 0.8 & 13.2 $\pm$ 1.1 & 11.4 $\pm$ 0.4 \\
microsoft/phi-1\_5 & 2.2 $\pm$ 0.2 & 1.4 $\pm$ 0.4 & 1.8 $\pm$ 0.2 & 2.9 $\pm$ 0.2 & 1.9 $\pm$ 0.2 & 3.4 $\pm$ 0.4 & 2.4 $\pm$ 0.5 & 2.3 $\pm$ 0.2 \\
Qwen/Qwen3-32B & 2.6 $\pm$ 0.2 & 3.4 $\pm$ 0.6 & 3.7 $\pm$ 0.3 & 3.0 $\pm$ 0.2 & 3.4 $\pm$ 0.3 & 2.5 $\pm$ 0.4 & 3.1 $\pm$ 0.5 & 2.8 $\pm$ 0.2 \\
meta-llama/Llama-2-7b-chat-hf & 13.8 $\pm$ 0.5 & 14.6 $\pm$ 1.2 & 11.5 $\pm$ 0.5 & 16.0 $\pm$ 0.5 & 15.7 $\pm$ 0.5 & 15.2 $\pm$ 0.9 & 16.6 $\pm$ 1.2 & 15.2 $\pm$ 0.5 \\
Qwen/Qwen2.5-Math-7B-Instruct & 14.9 $\pm$ 0.5 & 16.7 $\pm$ 1.3 & 12.4 $\pm$ 0.5 & 15.0 $\pm$ 0.5 & 15.2 $\pm$ 0.5 & 14.3 $\pm$ 0.9 & 17.3 $\pm$ 1.2 & 14.7 $\pm$ 0.5 \\
Qwen/Qwen3-14B & 9.0 $\pm$ 0.4 & 11.0 $\pm$ 1.1 & 14.1 $\pm$ 0.5 & 11.3 $\pm$ 0.4 & 11.6 $\pm$ 0.5 & 9.4 $\pm$ 0.7 & 12.2 $\pm$ 1.0 & 9.5 $\pm$ 0.4 \\
microsoft/Phi-4-reasoning & 0.3 $\pm$ 0.1 & 1.0 $\pm$ 0.3 & 0.4 $\pm$ 0.1 & 0.8 $\pm$ 0.1 & 0.6 $\pm$ 0.1 & 1.6 $\pm$ 0.3 & 0.8 $\pm$ 0.3 & 0.5 $\pm$ 0.1 \\
microsoft/phi-1 & 6.0 $\pm$ 0.3 & 4.8 $\pm$ 0.7 & 2.3 $\pm$ 0.2 & 4.5 $\pm$ 0.3 & 4.0 $\pm$ 0.3 & 4.8 $\pm$ 0.5 & 4.9 $\pm$ 0.7 & 5.0 $\pm$ 0.3 \\
google/gemini-2.5-pro & 0.0 $\pm$ 0.0 & 0.0 $\pm$ 0.0 & 2.0 $\pm$ 2.0 & 0.0 $\pm$ 0.0 & 0.0 $\pm$ 0.0 & 0.0 $\pm$ 0.0 & 0.0 $\pm$ 0.0 & 4.0 $\pm$ 2.8 \\
deepseek/deepseek-r1-0528 & 0.0 $\pm$ 0.0 & 0.0 $\pm$ 0.0 & 0.0 $\pm$ 0.0 & 0.0 $\pm$ 0.0 & 0.0 $\pm$ 0.0 & 0.0 $\pm$ 0.0 & 0.0 $\pm$ 0.0 & 0.0 $\pm$ 0.0 \\
\Xhline{1.5pt}
\end{tabular}%
}
\caption{Detailed evaluation results of current popular models on ArxivRollBench2025a for prediction tasks. Scores are percentages and use the same valid-response scoring protocol as the overall ranking.}
\label{tab:appendix-2025a-p}
\end{table*}

\begin{table*}[t]
\centering
\resizebox{0.92\textwidth}{!}{%
\begin{tabular}{c|lrrrr}
\Xhline{1.5pt}
Rank & Model & Valid Acc. & Raw Acc. & Coverage & Warning \\ \hline
1 & moonshotai/kimi-k2.6 & 70.8 $\pm$ 4.8 & 5.6 & 7.9 & high-null \\
2 & gemini-3.5-flash & 55.2 $\pm$ 1.5 & 55.2 & 100.0 & -- \\
3 & gpt-5.5 & 54.4 $\pm$ 1.5 & 54.4 & 100.0 & -- \\
4 & claude-opus-4-7 & 53.4 $\pm$ 1.5 & 52.9 & 99.0 & -- \\
5 & deepseek/deepseek-v4-pro & 53.3 $\pm$ 2.2 & 25.4 & 47.7 & high-null \\
6 & x-ai/grok-4.3 & 52.0 $\pm$ 1.5 & 51.8 & 99.7 & -- \\
7 & claude-sonnet-4-6 & 48.3 $\pm$ 1.5 & 48.3 & 100.0 & -- \\
8 & moonshotai/kimi-k2.5 & 46.0 $\pm$ 2.9 & 12.3 & 26.8 & high-null \\
9 & deepseek/deepseek-v4-flash & 45.3 $\pm$ 1.9 & 28.0 & 61.8 & high-null \\
10 & deepseek/deepseek-r1-0528 & 43.2 $\pm$ 2.6 & 14.3 & 33.0 & high-null \\
11 & claude-opus-4-6 & 42.9 $\pm$ 1.9 & 26.6 & 61.9 & high-null \\
12 & gpt-5.3-codex & 42.1 $\pm$ 1.5 & 40.0 & 95.0 & -- \\
13 & claude-haiku-4-5-20251001 & 40.8 $\pm$ 1.5 & 40.2 & 98.7 & -- \\
14 & gpt-5.4 & 39.3 $\pm$ 1.5 & 39.3 & 100.0 & -- \\
15 & Qwen/Qwen3-14B & 38.5 $\pm$ 0.5 & 7.4 & 19.1 & high-null \\
16 & deepseek/deepseek-chat-v3-0324 & 37.6 $\pm$ 1.4 & 37.6 & 100.0 & -- \\
17 & gpt-5.4-mini & 35.5 $\pm$ 1.4 & 35.4 & 99.8 & -- \\
18 & x-ai/grok-4.20 & 34.5 $\pm$ 1.4 & 34.5 & 100.0 & -- \\
19 & deepseek/deepseek-chat-v3.1 & 33.9 $\pm$ 1.4 & 33.9 & 100.0 & -- \\
20 & deepseek/deepseek-v3.2 & 31.8 $\pm$ 1.4 & 31.8 & 100.0 & -- \\
21 & Qwen/Qwen2.5-7B-Instruct & 30.0 $\pm$ 0.2 & 30.0 & 100.0 & -- \\
22 & Qwen/Qwen2-7B-Instruct & 29.5 $\pm$ 0.2 & 29.4 & 99.8 & -- \\
23 & Qwen/Qwen3-8B & 29.3 $\pm$ 0.2 & 29.3 & 99.8 & -- \\
24 & Qwen/Qwen2.5-7B & 28.2 $\pm$ 0.2 & 28.2 & 100.0 & -- \\
25 & meta-llama/Llama-3.1-8B-Instruct & 27.7 $\pm$ 0.2 & 27.7 & 100.0 & -- \\
26 & Qwen/Qwen3-4B & 27.4 $\pm$ 0.2 & 19.7 & 71.8 & high-null \\
27 & mistralai/Mistral-7B-Instruct-v0.3 & 26.2 $\pm$ 0.2 & 26.2 & 100.0 & -- \\
28 & mistralai/Mistral-7B-Instruct-v0.1 & 25.8 $\pm$ 0.2 & 25.8 & 100.0 & -- \\
29 & tiiuae/Falcon3-10B-Instruct & 23.8 $\pm$ 0.2 & 23.8 & 99.9 & -- \\
30 & mistralai/Mistral-7B-Instruct-v0.2 & 23.7 $\pm$ 0.2 & 22.7 & 95.6 & -- \\
31 & meta-llama/Llama-3.2-3B & 23.4 $\pm$ 0.2 & 23.4 & 100.0 & -- \\
32 & meta-llama/Llama-3.2-3B-Instruct & 23.2 $\pm$ 0.2 & 23.2 & 100.0 & -- \\
33 & meta-llama/Llama-3.2-1B & 22.1 $\pm$ 0.2 & 22.1 & 100.0 & -- \\
34 & meta-llama/Llama-3.1-8B & 21.9 $\pm$ 0.2 & 21.9 & 100.0 & -- \\
35 & Qwen/Qwen2.5-Math-7B & 21.6 $\pm$ 0.2 & 21.2 & 98.2 & -- \\
36 & microsoft/Phi-4-reasoning-plus & 21.2 $\pm$ 0.2 & 20.3 & 95.6 & -- \\
37 & meta-llama/Llama-3.2-1B-Instruct & 12.6 $\pm$ 0.1 & 12.6 & 100.0 & -- \\
38 & microsoft/Phi-4-reasoning & 8.3 $\pm$ 0.2 & 4.8 & 58.5 & high-null \\
39 & Qwen/Qwen2.5-Math-7B-Instruct & 7.5 $\pm$ 0.1 & 7.5 & 100.0 & -- \\
40 & deepseek/deepseek-v3.2-speciale & 0.0 $\pm$ 0.0 & 0.0 & 0.0 & missing \\
41 & gemini-3-flash-preview & 0.0 $\pm$ 0.0 & 0.0 & 100.0 & -- \\
42 & gemini-3.1-pro-preview & 0.0 $\pm$ 0.0 & 0.0 & 100.0 & -- \\
\Xhline{1.5pt}
\end{tabular}%
}
\caption{Overall supplemental evaluation results on ArxivRollBench2026a. Scores are percentages. Valid Acc. is the pooled valid accuracy and is used for ranking.}
\label{tab:appendix-2026a-overall}
\end{table*}

\begin{table*}[t]
\centering
\resizebox{\textwidth}{!}{%
\begin{tabular}{c|rrrrrrrr}
\Xhline{1.5pt}
Model name & \multicolumn{8}{c}{ArxivRollBench-2026a (S)} \\ \hline
 & CS & Q-Fin. & Math & Phy. & Stat. & Bio. & Econ. & EESS \\ \hline
moonshotai/kimi-k2.6 & 40.0 $\pm$ 21.9 & 50.0 $\pm$ 35.4 & 66.7 $\pm$ 27.2 & 75.0 $\pm$ 21.7 & 50.0 $\pm$ 25.0 & 100.0 $\pm$ 0.0 & 100.0 $\pm$ 0.0 & 57.1 $\pm$ 18.7 \\
gemini-3.5-flash & 56.9 $\pm$ 6.9 & 60.4 $\pm$ 7.1 & 65.9 $\pm$ 7.1 & 51.1 $\pm$ 7.3 & 53.1 $\pm$ 7.1 & 74.5 $\pm$ 6.4 & 62.5 $\pm$ 7.0 & 49.0 $\pm$ 7.0 \\
gpt-5.5 & 54.9 $\pm$ 7.0 & 56.2 $\pm$ 7.2 & 59.1 $\pm$ 7.4 & 55.3 $\pm$ 7.3 & 46.9 $\pm$ 7.1 & 63.8 $\pm$ 7.0 & 58.3 $\pm$ 7.1 & 54.9 $\pm$ 7.0 \\
claude-opus-4-7 & 52.9 $\pm$ 7.0 & 58.3 $\pm$ 7.1 & 65.9 $\pm$ 7.1 & 55.3 $\pm$ 7.3 & 50.0 $\pm$ 7.2 & 76.1 $\pm$ 6.3 & 70.8 $\pm$ 6.6 & 54.9 $\pm$ 7.0 \\
deepseek/deepseek-v4-pro & 51.4 $\pm$ 8.4 & 83.3 $\pm$ 8.8 & 64.7 $\pm$ 11.6 & 46.2 $\pm$ 9.8 & 54.5 $\pm$ 10.6 & 72.0 $\pm$ 9.0 & 66.7 $\pm$ 9.6 & 55.6 $\pm$ 9.6 \\
x-ai/grok-4.3 & 58.0 $\pm$ 7.0 & 66.7 $\pm$ 6.8 & 56.8 $\pm$ 7.5 & 44.7 $\pm$ 7.3 & 53.1 $\pm$ 7.1 & 72.3 $\pm$ 6.5 & 60.4 $\pm$ 7.1 & 51.0 $\pm$ 7.0 \\
claude-sonnet-4-6 & 52.9 $\pm$ 7.0 & 54.2 $\pm$ 7.2 & 47.7 $\pm$ 7.5 & 38.3 $\pm$ 7.1 & 42.9 $\pm$ 7.1 & 61.7 $\pm$ 7.1 & 62.5 $\pm$ 7.0 & 41.2 $\pm$ 6.9 \\
moonshotai/kimi-k2.5 & 42.9 $\pm$ 18.7 & 33.3 $\pm$ 13.6 & 76.9 $\pm$ 11.7 & 54.5 $\pm$ 15.0 & 50.0 $\pm$ 17.7 & 66.7 $\pm$ 13.6 & 73.3 $\pm$ 11.4 & 35.3 $\pm$ 11.6 \\
deepseek/deepseek-v4-flash & 54.1 $\pm$ 8.2 & 70.4 $\pm$ 8.8 & 64.0 $\pm$ 9.6 & 37.0 $\pm$ 9.3 & 38.7 $\pm$ 8.7 & 58.6 $\pm$ 9.1 & 63.6 $\pm$ 8.4 & 47.1 $\pm$ 8.6 \\
deepseek/deepseek-r1-0528 & 63.6 $\pm$ 10.3 & 58.3 $\pm$ 14.2 & 62.5 $\pm$ 12.1 & 31.8 $\pm$ 9.9 & 20.0 $\pm$ 12.6 & 61.5 $\pm$ 13.5 & 57.1 $\pm$ 13.2 & 55.6 $\pm$ 11.7 \\
claude-opus-4-6 & 0.0 $\pm$ 0.0 & 60.4 $\pm$ 7.1 & 63.6 $\pm$ 7.3 & 55.3 $\pm$ 7.3 & 46.9 $\pm$ 7.1 & 70.2 $\pm$ 6.7 & 56.2 $\pm$ 8.8 & 49.0 $\pm$ 7.0 \\
gpt-5.3-codex & 47.1 $\pm$ 7.0 & 50.0 $\pm$ 7.4 & 47.7 $\pm$ 7.5 & 25.0 $\pm$ 7.2 & 43.2 $\pm$ 7.5 & 78.1 $\pm$ 7.3 & 48.9 $\pm$ 7.5 & 43.1 $\pm$ 6.9 \\
claude-haiku-4-5-20251001 & 48.0 $\pm$ 7.1 & 52.1 $\pm$ 7.2 & 43.2 $\pm$ 7.5 & 36.2 $\pm$ 7.0 & 36.7 $\pm$ 6.9 & 55.3 $\pm$ 7.3 & 39.6 $\pm$ 7.1 & 39.2 $\pm$ 6.8 \\
gpt-5.4 & 39.2 $\pm$ 6.8 & 33.3 $\pm$ 6.8 & 54.5 $\pm$ 7.5 & 36.2 $\pm$ 7.0 & 44.9 $\pm$ 7.1 & 53.2 $\pm$ 7.3 & 52.1 $\pm$ 7.2 & 51.0 $\pm$ 7.0 \\
Qwen/Qwen3-14B & 38.1 $\pm$ 2.9 & 37.5 $\pm$ 3.7 & 31.5 $\pm$ 2.2 & 34.5 $\pm$ 2.8 & 34.9 $\pm$ 2.1 & 31.2 $\pm$ 2.9 & 34.0 $\pm$ 3.8 & 35.8 $\pm$ 2.0 \\
deepseek/deepseek-chat-v3-0324 & 39.2 $\pm$ 6.8 & 39.6 $\pm$ 7.1 & 40.9 $\pm$ 7.4 & 36.2 $\pm$ 7.0 & 42.9 $\pm$ 7.1 & 55.3 $\pm$ 7.3 & 54.2 $\pm$ 7.2 & 43.1 $\pm$ 6.9 \\
gpt-5.4-mini & 45.1 $\pm$ 7.0 & 37.5 $\pm$ 7.0 & 61.4 $\pm$ 7.3 & 36.2 $\pm$ 7.0 & 36.7 $\pm$ 6.9 & 48.9 $\pm$ 7.3 & 39.6 $\pm$ 7.1 & 49.0 $\pm$ 7.0 \\
x-ai/grok-4.20 & 39.2 $\pm$ 6.8 & 47.9 $\pm$ 7.2 & 38.6 $\pm$ 7.3 & 38.3 $\pm$ 7.1 & 38.8 $\pm$ 7.0 & 51.1 $\pm$ 7.3 & 47.9 $\pm$ 7.2 & 39.2 $\pm$ 6.8 \\
deepseek/deepseek-chat-v3.1 & 39.2 $\pm$ 6.8 & 33.3 $\pm$ 6.8 & 36.4 $\pm$ 7.3 & 34.0 $\pm$ 6.9 & 42.9 $\pm$ 7.1 & 48.9 $\pm$ 7.3 & 35.4 $\pm$ 6.9 & 49.0 $\pm$ 7.0 \\
deepseek/deepseek-v3.2 & 33.3 $\pm$ 6.6 & 39.6 $\pm$ 7.1 & 36.4 $\pm$ 7.3 & 38.3 $\pm$ 7.1 & 44.9 $\pm$ 7.1 & 48.9 $\pm$ 7.3 & 39.6 $\pm$ 7.1 & 47.1 $\pm$ 7.0 \\
Qwen/Qwen2.5-7B-Instruct & 27.0 $\pm$ 1.0 & 29.7 $\pm$ 1.5 & 26.8 $\pm$ 0.9 & 27.4 $\pm$ 1.0 & 27.4 $\pm$ 0.8 & 28.6 $\pm$ 1.2 & 27.5 $\pm$ 1.5 & 28.8 $\pm$ 0.7 \\
Qwen/Qwen2-7B-Instruct & 26.5 $\pm$ 1.0 & 27.1 $\pm$ 1.4 & 27.0 $\pm$ 0.9 & 26.2 $\pm$ 1.0 & 28.1 $\pm$ 0.8 & 27.8 $\pm$ 1.2 & 25.4 $\pm$ 1.4 & 28.2 $\pm$ 0.7 \\
Qwen/Qwen3-8B & 32.0 $\pm$ 1.1 & 28.7 $\pm$ 1.4 & 32.6 $\pm$ 0.9 & 31.1 $\pm$ 1.1 & 30.1 $\pm$ 0.8 & 30.6 $\pm$ 1.2 & 29.1 $\pm$ 1.5 & 29.8 $\pm$ 0.7 \\
Qwen/Qwen2.5-7B & 24.3 $\pm$ 1.0 & 25.7 $\pm$ 1.4 & 23.4 $\pm$ 0.9 & 24.9 $\pm$ 1.0 & 24.3 $\pm$ 0.8 & 24.0 $\pm$ 1.1 & 20.9 $\pm$ 1.3 & 24.2 $\pm$ 0.7 \\
meta-llama/Llama-3.1-8B-Instruct & 28.4 $\pm$ 1.0 & 26.4 $\pm$ 1.4 & 26.7 $\pm$ 0.9 & 27.3 $\pm$ 1.0 & 26.9 $\pm$ 0.8 & 29.2 $\pm$ 1.2 & 25.6 $\pm$ 1.4 & 27.4 $\pm$ 0.7 \\
Qwen/Qwen3-4B & 27.3 $\pm$ 1.1 & 27.7 $\pm$ 1.5 & 26.3 $\pm$ 0.9 & 29.0 $\pm$ 1.1 & 28.0 $\pm$ 0.8 & 28.6 $\pm$ 1.2 & 26.8 $\pm$ 1.5 & 28.5 $\pm$ 0.7 \\
mistralai/Mistral-7B-Instruct-v0.3 & 22.6 $\pm$ 1.0 & 26.4 $\pm$ 1.4 & 24.1 $\pm$ 0.9 & 24.5 $\pm$ 1.0 & 24.5 $\pm$ 0.8 & 24.2 $\pm$ 1.1 & 24.5 $\pm$ 1.4 & 25.1 $\pm$ 0.7 \\
mistralai/Mistral-7B-Instruct-v0.1 & 23.6 $\pm$ 1.0 & 27.7 $\pm$ 1.4 & 24.6 $\pm$ 0.9 & 24.7 $\pm$ 1.0 & 25.1 $\pm$ 0.8 & 24.6 $\pm$ 1.1 & 25.3 $\pm$ 1.4 & 26.1 $\pm$ 0.7 \\
tiiuae/Falcon3-10B-Instruct & 16.2 $\pm$ 0.9 & 16.8 $\pm$ 1.2 & 15.0 $\pm$ 0.7 & 15.4 $\pm$ 0.8 & 14.6 $\pm$ 0.6 & 16.4 $\pm$ 1.0 & 12.7 $\pm$ 1.1 & 17.0 $\pm$ 0.6 \\
mistralai/Mistral-7B-Instruct-v0.2 & 22.7 $\pm$ 1.0 & 26.4 $\pm$ 1.4 & 24.4 $\pm$ 0.9 & 24.3 $\pm$ 1.0 & 24.5 $\pm$ 0.8 & 24.1 $\pm$ 1.1 & 24.9 $\pm$ 1.4 & 25.0 $\pm$ 0.7 \\
meta-llama/Llama-3.2-3B & 22.4 $\pm$ 1.0 & 23.7 $\pm$ 1.4 & 22.3 $\pm$ 0.8 & 22.1 $\pm$ 1.0 & 21.8 $\pm$ 0.7 & 21.9 $\pm$ 1.1 & 23.2 $\pm$ 1.4 & 23.2 $\pm$ 0.7 \\
meta-llama/Llama-3.2-3B-Instruct & 16.0 $\pm$ 0.9 & 15.1 $\pm$ 1.1 & 16.4 $\pm$ 0.8 & 16.9 $\pm$ 0.9 & 16.3 $\pm$ 0.7 & 15.3 $\pm$ 0.9 & 15.4 $\pm$ 1.2 & 16.6 $\pm$ 0.6 \\
meta-llama/Llama-3.2-1B & 23.3 $\pm$ 1.0 & 27.5 $\pm$ 1.4 & 24.5 $\pm$ 0.9 & 24.5 $\pm$ 1.0 & 24.9 $\pm$ 0.8 & 24.7 $\pm$ 1.1 & 25.1 $\pm$ 1.4 & 25.8 $\pm$ 0.7 \\
meta-llama/Llama-3.1-8B & 25.7 $\pm$ 1.0 & 23.8 $\pm$ 1.4 & 25.2 $\pm$ 0.9 & 26.6 $\pm$ 1.0 & 25.8 $\pm$ 0.8 & 24.2 $\pm$ 1.1 & 25.6 $\pm$ 1.4 & 25.2 $\pm$ 0.7 \\
Qwen/Qwen2.5-Math-7B & 15.9 $\pm$ 0.8 & 17.2 $\pm$ 1.2 & 18.3 $\pm$ 0.8 & 17.6 $\pm$ 0.9 & 17.3 $\pm$ 0.7 & 16.9 $\pm$ 1.0 & 15.3 $\pm$ 1.2 & 17.1 $\pm$ 0.6 \\
microsoft/Phi-4-reasoning-plus & 8.6 $\pm$ 0.7 & 12.8 $\pm$ 1.1 & 12.7 $\pm$ 0.7 & 9.7 $\pm$ 0.7 & 11.2 $\pm$ 0.6 & 9.1 $\pm$ 0.7 & 14.0 $\pm$ 1.1 & 9.1 $\pm$ 0.4 \\
meta-llama/Llama-3.2-1B-Instruct & 8.2 $\pm$ 0.6 & 9.5 $\pm$ 0.9 & 6.0 $\pm$ 0.5 & 6.6 $\pm$ 0.6 & 8.0 $\pm$ 0.5 & 7.7 $\pm$ 0.7 & 7.8 $\pm$ 0.9 & 7.1 $\pm$ 0.4 \\
microsoft/Phi-4-reasoning & 1.5 $\pm$ 0.3 & 2.0 $\pm$ 0.4 & 3.4 $\pm$ 0.4 & 1.1 $\pm$ 0.2 & 2.3 $\pm$ 0.3 & 1.7 $\pm$ 0.3 & 3.0 $\pm$ 0.6 & 1.4 $\pm$ 0.2 \\
Qwen/Qwen2.5-Math-7B-Instruct & 4.6 $\pm$ 0.5 & 4.9 $\pm$ 0.7 & 5.0 $\pm$ 0.4 & 5.4 $\pm$ 0.5 & 6.6 $\pm$ 0.4 & 5.9 $\pm$ 0.6 & 7.2 $\pm$ 0.9 & 5.8 $\pm$ 0.4 \\
deepseek/deepseek-v3.2-speciale & 0.0 $\pm$ 0.0 & 0.0 $\pm$ 0.0 & 0.0 $\pm$ 0.0 & 0.0 $\pm$ 0.0 & 0.0 $\pm$ 0.0 & 0.0 $\pm$ 0.0 & 0.0 $\pm$ 0.0 & 0.0 $\pm$ 0.0 \\
gemini-3-flash-preview & -- & -- & -- & -- & -- & -- & -- & -- \\
gemini-3.1-pro-preview & -- & -- & -- & -- & -- & -- & -- & -- \\
\Xhline{1.5pt}
\end{tabular}%
}
\caption{Detailed evaluation results of current popular models on ArxivRollBench2026a for sequencing tasks. Scores are percentages and use the same valid-response scoring protocol as the overall ranking.}
\label{tab:appendix-2026a-s}
\end{table*}

\begin{table*}[t]
\centering
\resizebox{\textwidth}{!}{%
\begin{tabular}{c|rrrrrrrr}
\Xhline{1.5pt}
Model name & \multicolumn{8}{c}{ArxivRollBench-2026a (C)} \\ \hline
 & CS & Q-Fin. & Math & Phy. & Stat. & Bio. & Econ. & EESS \\ \hline
moonshotai/kimi-k2.6 & 50.0 $\pm$ 35.4 & 66.7 $\pm$ 27.2 & 0.0 $\pm$ 0.0 & 50.0 $\pm$ 35.4 & 100.0 $\pm$ 0.0 & 0.0 $\pm$ 0.0 & 0.0 $\pm$ 0.0 & 50.0 $\pm$ 25.0 \\
gemini-3.5-flash & 18.4 $\pm$ 6.3 & 24.4 $\pm$ 6.4 & 30.0 $\pm$ 8.4 & 34.1 $\pm$ 7.4 & 28.3 $\pm$ 6.6 & 40.0 $\pm$ 7.7 & 29.8 $\pm$ 6.7 & 22.2 $\pm$ 6.2 \\
gpt-5.5 & 21.1 $\pm$ 6.6 & 22.2 $\pm$ 6.2 & 30.0 $\pm$ 8.4 & 31.7 $\pm$ 7.3 & 28.3 $\pm$ 6.6 & 40.0 $\pm$ 7.7 & 34.0 $\pm$ 6.9 & 26.7 $\pm$ 6.6 \\
claude-opus-4-7 & 21.1 $\pm$ 6.6 & 24.4 $\pm$ 6.4 & 30.0 $\pm$ 8.4 & 30.0 $\pm$ 7.2 & 23.9 $\pm$ 6.3 & 37.8 $\pm$ 8.0 & 29.8 $\pm$ 6.7 & 26.7 $\pm$ 6.6 \\
deepseek/deepseek-v4-pro & 13.3 $\pm$ 8.8 & 37.5 $\pm$ 12.1 & 38.5 $\pm$ 13.5 & 14.3 $\pm$ 9.4 & 42.1 $\pm$ 11.3 & 20.8 $\pm$ 8.3 & 27.8 $\pm$ 10.6 & 6.2 $\pm$ 6.1 \\
x-ai/grok-4.3 & 23.7 $\pm$ 6.9 & 22.2 $\pm$ 6.2 & 30.0 $\pm$ 8.4 & 34.1 $\pm$ 7.4 & 28.3 $\pm$ 6.6 & 37.5 $\pm$ 7.7 & 34.0 $\pm$ 6.9 & 26.7 $\pm$ 6.6 \\
claude-sonnet-4-6 & 15.8 $\pm$ 5.9 & 22.2 $\pm$ 6.2 & 26.7 $\pm$ 8.1 & 22.0 $\pm$ 6.5 & 23.9 $\pm$ 6.3 & 35.0 $\pm$ 7.5 & 23.4 $\pm$ 6.2 & 22.2 $\pm$ 6.2 \\
moonshotai/kimi-k2.5 & 10.0 $\pm$ 9.5 & 13.3 $\pm$ 8.8 & 16.7 $\pm$ 15.2 & 33.3 $\pm$ 12.2 & 41.7 $\pm$ 14.2 & 45.5 $\pm$ 15.0 & 35.7 $\pm$ 12.8 & 12.5 $\pm$ 8.3 \\
deepseek/deepseek-v4-flash & 25.0 $\pm$ 9.7 & 18.2 $\pm$ 8.2 & 18.2 $\pm$ 11.6 & 0.0 $\pm$ 0.0 & 37.5 $\pm$ 9.9 & 6.7 $\pm$ 6.4 & 26.7 $\pm$ 8.1 & 33.3 $\pm$ 10.3 \\
deepseek/deepseek-r1-0528 & 25.0 $\pm$ 12.5 & 12.5 $\pm$ 8.3 & 30.0 $\pm$ 14.5 & 25.0 $\pm$ 10.8 & 13.3 $\pm$ 8.8 & 33.3 $\pm$ 15.7 & 30.8 $\pm$ 12.8 & 29.4 $\pm$ 11.1 \\
claude-opus-4-6 & 15.8 $\pm$ 5.9 & 24.4 $\pm$ 6.4 & 33.3 $\pm$ 8.6 & 29.3 $\pm$ 7.1 & 28.3 $\pm$ 6.6 & 35.0 $\pm$ 7.5 & 31.9 $\pm$ 6.8 & 17.8 $\pm$ 5.7 \\
gpt-5.3-codex & 2.6 $\pm$ 2.6 & 8.9 $\pm$ 4.2 & 12.5 $\pm$ 6.8 & 22.0 $\pm$ 6.5 & 8.7 $\pm$ 4.2 & 7.5 $\pm$ 4.2 & 13.3 $\pm$ 5.1 & 18.2 $\pm$ 5.8 \\
claude-haiku-4-5-20251001 & 18.4 $\pm$ 6.3 & 22.2 $\pm$ 6.2 & 23.3 $\pm$ 7.7 & 26.8 $\pm$ 6.9 & 26.1 $\pm$ 6.5 & 30.0 $\pm$ 7.2 & 23.4 $\pm$ 6.2 & 26.7 $\pm$ 6.6 \\
gpt-5.4 & 7.9 $\pm$ 4.4 & 15.6 $\pm$ 5.4 & 23.3 $\pm$ 7.7 & 17.1 $\pm$ 5.9 & 17.4 $\pm$ 5.6 & 15.0 $\pm$ 5.6 & 17.0 $\pm$ 5.5 & 22.2 $\pm$ 6.2 \\
Qwen/Qwen3-14B & 24.2 $\pm$ 3.1 & 24.6 $\pm$ 3.3 & 28.7 $\pm$ 2.6 & 25.8 $\pm$ 2.4 & 25.2 $\pm$ 2.0 & 26.2 $\pm$ 2.8 & 24.7 $\pm$ 2.9 & 24.9 $\pm$ 2.0 \\
deepseek/deepseek-chat-v3-0324 & 10.5 $\pm$ 5.0 & 11.1 $\pm$ 4.7 & 26.7 $\pm$ 8.1 & 26.8 $\pm$ 6.9 & 17.4 $\pm$ 5.6 & 22.5 $\pm$ 6.6 & 21.3 $\pm$ 6.0 & 22.2 $\pm$ 6.2 \\
gpt-5.4-mini & 5.3 $\pm$ 3.6 & 13.6 $\pm$ 5.2 & 13.8 $\pm$ 6.4 & 19.5 $\pm$ 6.2 & 19.6 $\pm$ 5.8 & 22.5 $\pm$ 6.6 & 17.0 $\pm$ 5.5 & 13.3 $\pm$ 5.1 \\
x-ai/grok-4.20 & 2.6 $\pm$ 2.6 & 6.7 $\pm$ 3.7 & 20.0 $\pm$ 7.3 & 14.6 $\pm$ 5.5 & 6.5 $\pm$ 3.6 & 7.5 $\pm$ 4.2 & 12.8 $\pm$ 4.9 & 15.6 $\pm$ 5.4 \\
deepseek/deepseek-chat-v3.1 & 5.3 $\pm$ 3.6 & 13.3 $\pm$ 5.1 & 20.0 $\pm$ 7.3 & 19.5 $\pm$ 6.2 & 15.2 $\pm$ 5.3 & 25.0 $\pm$ 6.8 & 14.9 $\pm$ 5.2 & 15.6 $\pm$ 5.4 \\
deepseek/deepseek-v3.2 & 5.3 $\pm$ 3.6 & 6.7 $\pm$ 3.7 & 16.7 $\pm$ 6.8 & 14.6 $\pm$ 5.5 & 10.9 $\pm$ 4.6 & 15.0 $\pm$ 5.6 & 6.4 $\pm$ 3.6 & 13.3 $\pm$ 5.1 \\
Qwen/Qwen2.5-7B-Instruct & 21.2 $\pm$ 1.0 & 21.7 $\pm$ 1.4 & 23.9 $\pm$ 1.1 & 22.0 $\pm$ 1.0 & 23.4 $\pm$ 0.8 & 20.7 $\pm$ 1.1 & 22.6 $\pm$ 1.5 & 21.0 $\pm$ 0.7 \\
Qwen/Qwen2-7B-Instruct & 25.4 $\pm$ 1.1 & 25.7 $\pm$ 1.5 & 26.4 $\pm$ 1.1 & 25.9 $\pm$ 1.1 & 25.3 $\pm$ 0.8 & 25.9 $\pm$ 1.2 & 25.0 $\pm$ 1.5 & 25.4 $\pm$ 0.7 \\
Qwen/Qwen3-8B & 24.0 $\pm$ 1.1 & 22.7 $\pm$ 1.4 & 26.2 $\pm$ 1.1 & 23.8 $\pm$ 1.1 & 23.8 $\pm$ 0.8 & 23.7 $\pm$ 1.2 & 26.7 $\pm$ 1.5 & 23.8 $\pm$ 0.7 \\
Qwen/Qwen2.5-7B & 25.3 $\pm$ 1.1 & 26.7 $\pm$ 1.5 & 29.8 $\pm$ 1.1 & 24.8 $\pm$ 1.1 & 27.3 $\pm$ 0.9 & 27.3 $\pm$ 1.2 & 27.7 $\pm$ 1.6 & 26.1 $\pm$ 0.7 \\
meta-llama/Llama-3.1-8B-Instruct & 26.1 $\pm$ 1.1 & 23.0 $\pm$ 1.4 & 27.4 $\pm$ 1.1 & 25.9 $\pm$ 1.1 & 26.8 $\pm$ 0.9 & 25.1 $\pm$ 1.2 & 24.8 $\pm$ 1.5 & 26.5 $\pm$ 0.7 \\
Qwen/Qwen3-4B & 24.6 $\pm$ 3.6 & 21.7 $\pm$ 3.8 & 21.6 $\pm$ 2.1 & 19.2 $\pm$ 2.6 & 21.6 $\pm$ 2.0 & 18.8 $\pm$ 3.3 & 20.3 $\pm$ 3.7 & 19.5 $\pm$ 2.0 \\
mistralai/Mistral-7B-Instruct-v0.3 & 30.1 $\pm$ 1.1 & 24.8 $\pm$ 1.5 & 28.6 $\pm$ 1.1 & 29.0 $\pm$ 1.1 & 28.2 $\pm$ 0.9 & 28.6 $\pm$ 1.2 & 28.1 $\pm$ 1.6 & 29.2 $\pm$ 0.7 \\
mistralai/Mistral-7B-Instruct-v0.1 & 28.3 $\pm$ 1.1 & 29.5 $\pm$ 1.5 & 27.6 $\pm$ 1.1 & 26.7 $\pm$ 1.1 & 27.6 $\pm$ 0.9 & 28.8 $\pm$ 1.2 & 25.0 $\pm$ 1.5 & 27.1 $\pm$ 0.7 \\
tiiuae/Falcon3-10B-Instruct & 16.9 $\pm$ 0.9 & 19.1 $\pm$ 1.3 & 19.4 $\pm$ 1.0 & 17.4 $\pm$ 1.0 & 18.1 $\pm$ 0.8 & 16.8 $\pm$ 1.0 & 17.1 $\pm$ 1.3 & 18.1 $\pm$ 0.6 \\
mistralai/Mistral-7B-Instruct-v0.2 & 21.9 $\pm$ 1.0 & 20.8 $\pm$ 1.4 & 23.7 $\pm$ 1.1 & 23.5 $\pm$ 1.1 & 22.4 $\pm$ 0.8 & 24.2 $\pm$ 1.2 & 21.0 $\pm$ 1.4 & 23.0 $\pm$ 0.7 \\
meta-llama/Llama-3.2-3B & 24.2 $\pm$ 1.1 & 22.9 $\pm$ 1.4 & 26.9 $\pm$ 1.1 & 25.5 $\pm$ 1.1 & 25.7 $\pm$ 0.9 & 24.9 $\pm$ 1.2 & 24.6 $\pm$ 1.5 & 25.2 $\pm$ 0.7 \\
meta-llama/Llama-3.2-3B-Instruct & 30.0 $\pm$ 1.1 & 30.2 $\pm$ 1.5 & 30.7 $\pm$ 1.2 & 30.8 $\pm$ 1.2 & 29.4 $\pm$ 0.9 & 31.1 $\pm$ 1.2 & 30.1 $\pm$ 1.6 & 30.2 $\pm$ 0.8 \\
meta-llama/Llama-3.2-1B & 13.5 $\pm$ 0.8 & 13.1 $\pm$ 1.1 & 14.9 $\pm$ 0.9 & 10.7 $\pm$ 0.8 & 16.9 $\pm$ 0.7 & 14.6 $\pm$ 1.0 & 17.1 $\pm$ 1.3 & 15.3 $\pm$ 0.6 \\
meta-llama/Llama-3.1-8B & 14.0 $\pm$ 0.9 & 13.0 $\pm$ 1.1 & 18.9 $\pm$ 1.0 & 15.6 $\pm$ 0.9 & 15.7 $\pm$ 0.7 & 15.7 $\pm$ 1.0 & 13.5 $\pm$ 1.2 & 15.5 $\pm$ 0.6 \\
Qwen/Qwen2.5-Math-7B & 23.8 $\pm$ 1.0 & 22.2 $\pm$ 1.4 & 25.4 $\pm$ 1.1 & 24.6 $\pm$ 1.1 & 24.9 $\pm$ 0.8 & 25.0 $\pm$ 1.2 & 27.2 $\pm$ 1.5 & 24.5 $\pm$ 0.7 \\
microsoft/Phi-4-reasoning-plus & 21.6 $\pm$ 1.0 & 22.5 $\pm$ 1.4 & 26.0 $\pm$ 1.1 & 21.8 $\pm$ 1.0 & 23.1 $\pm$ 0.8 & 22.9 $\pm$ 1.1 & 24.2 $\pm$ 1.5 & 23.5 $\pm$ 0.7 \\
meta-llama/Llama-3.2-1B-Instruct & 3.5 $\pm$ 0.5 & 4.0 $\pm$ 0.7 & 3.4 $\pm$ 0.5 & 2.9 $\pm$ 0.4 & 4.4 $\pm$ 0.4 & 5.0 $\pm$ 0.6 & 4.0 $\pm$ 0.7 & 4.1 $\pm$ 0.3 \\
microsoft/Phi-4-reasoning & 18.2 $\pm$ 1.4 & 21.7 $\pm$ 1.7 & 23.5 $\pm$ 1.2 & 18.2 $\pm$ 1.2 & 20.7 $\pm$ 1.0 & 20.1 $\pm$ 1.4 & 23.6 $\pm$ 1.7 & 19.8 $\pm$ 0.9 \\
Qwen/Qwen2.5-Math-7B-Instruct & 1.9 $\pm$ 0.3 & 1.7 $\pm$ 0.4 & 2.2 $\pm$ 0.4 & 2.3 $\pm$ 0.4 & 1.3 $\pm$ 0.2 & 1.4 $\pm$ 0.3 & 1.4 $\pm$ 0.4 & 1.5 $\pm$ 0.2 \\
deepseek/deepseek-v3.2-speciale & 0.0 $\pm$ 0.0 & 0.0 $\pm$ 0.0 & 0.0 $\pm$ 0.0 & 0.0 $\pm$ 0.0 & 0.0 $\pm$ 0.0 & 0.0 $\pm$ 0.0 & 0.0 $\pm$ 0.0 & 0.0 $\pm$ 0.0 \\
gemini-3-flash-preview & -- & -- & -- & -- & -- & -- & -- & -- \\
gemini-3.1-pro-preview & -- & -- & -- & -- & -- & -- & -- & -- \\
\Xhline{1.5pt}
\end{tabular}%
}
\caption{Detailed evaluation results of current popular models on ArxivRollBench2026a for cloze tasks. Scores are percentages and use the same valid-response scoring protocol as the overall ranking.}
\label{tab:appendix-2026a-c}
\end{table*}

\begin{table*}[t]
\centering
\resizebox{\textwidth}{!}{%
\begin{tabular}{c|rrrrrrrr}
\Xhline{1.5pt}
Model name & \multicolumn{8}{c}{ArxivRollBench-2026a (P)} \\ \hline
 & CS & Q-Fin. & Math & Phy. & Stat. & Bio. & Econ. & EESS \\ \hline
moonshotai/kimi-k2.6 & 100.0 $\pm$ 0.0 & 100.0 $\pm$ 0.0 & 83.3 $\pm$ 15.2 & 66.7 $\pm$ 27.2 & 100.0 $\pm$ 0.0 & 80.0 $\pm$ 17.9 & 83.3 $\pm$ 15.2 & 80.0 $\pm$ 17.9 \\
gemini-3.5-flash & 84.3 $\pm$ 5.1 & 74.5 $\pm$ 6.1 & 80.4 $\pm$ 5.6 & 60.0 $\pm$ 6.9 & 74.5 $\pm$ 6.1 & 75.0 $\pm$ 6.2 & 66.7 $\pm$ 6.6 & 74.5 $\pm$ 6.1 \\
gpt-5.5 & 72.5 $\pm$ 6.2 & 82.4 $\pm$ 5.3 & 72.5 $\pm$ 6.2 & 64.0 $\pm$ 6.8 & 62.7 $\pm$ 6.8 & 77.1 $\pm$ 6.1 & 78.4 $\pm$ 5.8 & 78.4 $\pm$ 5.8 \\
claude-opus-4-7 & 58.8 $\pm$ 6.9 & 72.5 $\pm$ 6.2 & 72.5 $\pm$ 6.2 & 67.3 $\pm$ 6.7 & 68.6 $\pm$ 6.5 & 64.4 $\pm$ 7.1 & 66.7 $\pm$ 6.6 & 72.0 $\pm$ 6.3 \\
deepseek/deepseek-v4-pro & 55.2 $\pm$ 9.2 & 83.3 $\pm$ 7.6 & 76.9 $\pm$ 8.3 & 61.9 $\pm$ 10.6 & 70.8 $\pm$ 9.3 & 57.7 $\pm$ 9.7 & 48.0 $\pm$ 10.0 & 67.7 $\pm$ 8.4 \\
x-ai/grok-4.3 & 60.0 $\pm$ 6.9 & 76.5 $\pm$ 5.9 & 68.0 $\pm$ 6.6 & 54.0 $\pm$ 7.0 & 62.7 $\pm$ 6.8 & 64.6 $\pm$ 6.9 & 56.9 $\pm$ 6.9 & 76.5 $\pm$ 5.9 \\
claude-sonnet-4-6 & 62.7 $\pm$ 6.8 & 76.5 $\pm$ 5.9 & 76.5 $\pm$ 5.9 & 66.0 $\pm$ 6.7 & 60.8 $\pm$ 6.8 & 66.7 $\pm$ 6.8 & 54.9 $\pm$ 7.0 & 68.6 $\pm$ 6.5 \\
moonshotai/kimi-k2.5 & 66.7 $\pm$ 15.7 & 63.2 $\pm$ 11.1 & 69.2 $\pm$ 12.8 & 29.4 $\pm$ 11.1 & 61.5 $\pm$ 13.5 & 72.7 $\pm$ 13.4 & 41.2 $\pm$ 11.9 & 71.4 $\pm$ 17.1 \\
deepseek/deepseek-v4-flash & 52.5 $\pm$ 7.9 & 59.5 $\pm$ 8.1 & 56.8 $\pm$ 8.1 & 38.7 $\pm$ 8.7 & 39.5 $\pm$ 7.9 & 64.7 $\pm$ 8.2 & 44.4 $\pm$ 8.3 & 47.4 $\pm$ 8.1 \\
deepseek/deepseek-r1-0528 & 44.4 $\pm$ 11.7 & 80.0 $\pm$ 10.3 & 70.0 $\pm$ 14.5 & 27.8 $\pm$ 10.6 & 57.9 $\pm$ 11.3 & 54.5 $\pm$ 10.6 & 42.9 $\pm$ 13.2 & 36.8 $\pm$ 11.1 \\
claude-opus-4-6 & 60.0 $\pm$ 7.7 & 0.0 $\pm$ 0.0 & 0.0 $\pm$ 0.0 & 0.0 $\pm$ 0.0 & 0.0 $\pm$ 0.0 & 0.0 $\pm$ 0.0 & 75.0 $\pm$ 21.7 & 0.0 $\pm$ 0.0 \\
gpt-5.3-codex & 56.2 $\pm$ 7.2 & 69.4 $\pm$ 6.6 & 68.0 $\pm$ 6.6 & 58.3 $\pm$ 7.1 & 61.2 $\pm$ 7.0 & 66.7 $\pm$ 6.8 & 52.0 $\pm$ 7.1 & 66.7 $\pm$ 6.6 \\
claude-haiku-4-5-20251001 & 50.0 $\pm$ 7.2 & 64.7 $\pm$ 6.7 & 54.9 $\pm$ 7.0 & 50.0 $\pm$ 7.1 & 55.6 $\pm$ 7.4 & 43.8 $\pm$ 7.2 & 41.3 $\pm$ 7.3 & 51.0 $\pm$ 7.0 \\
gpt-5.4 & 41.2 $\pm$ 6.9 & 45.1 $\pm$ 7.0 & 56.9 $\pm$ 6.9 & 56.0 $\pm$ 7.0 & 58.8 $\pm$ 6.9 & 64.6 $\pm$ 6.9 & 37.3 $\pm$ 6.8 & 56.9 $\pm$ 6.9 \\
Qwen/Qwen3-14B & 47.3 $\pm$ 2.6 & 41.6 $\pm$ 3.3 & 45.5 $\pm$ 1.7 & 45.1 $\pm$ 2.3 & 50.0 $\pm$ 1.7 & 51.1 $\pm$ 2.9 & 45.1 $\pm$ 3.2 & 51.2 $\pm$ 1.6 \\
deepseek/deepseek-chat-v3-0324 & 39.2 $\pm$ 6.8 & 52.9 $\pm$ 7.0 & 45.1 $\pm$ 7.0 & 44.0 $\pm$ 7.0 & 52.9 $\pm$ 7.0 & 47.9 $\pm$ 7.2 & 33.3 $\pm$ 6.6 & 54.9 $\pm$ 7.0 \\
gpt-5.4-mini & 43.1 $\pm$ 6.9 & 43.1 $\pm$ 6.9 & 45.1 $\pm$ 7.0 & 38.0 $\pm$ 6.9 & 47.1 $\pm$ 7.0 & 45.8 $\pm$ 7.2 & 35.3 $\pm$ 6.7 & 49.0 $\pm$ 7.0 \\
x-ai/grok-4.20 & 39.2 $\pm$ 6.8 & 49.0 $\pm$ 7.0 & 41.2 $\pm$ 6.9 & 46.0 $\pm$ 7.0 & 43.1 $\pm$ 6.9 & 54.2 $\pm$ 7.2 & 37.3 $\pm$ 6.8 & 62.7 $\pm$ 6.8 \\
deepseek/deepseek-chat-v3.1 & 33.3 $\pm$ 6.6 & 45.1 $\pm$ 7.0 & 39.2 $\pm$ 6.8 & 38.0 $\pm$ 6.9 & 52.9 $\pm$ 7.0 & 47.9 $\pm$ 7.2 & 37.3 $\pm$ 6.8 & 49.0 $\pm$ 7.0 \\
deepseek/deepseek-v3.2 & 31.4 $\pm$ 6.5 & 45.1 $\pm$ 7.0 & 37.3 $\pm$ 6.8 & 38.0 $\pm$ 6.9 & 39.2 $\pm$ 6.8 & 47.9 $\pm$ 7.2 & 33.3 $\pm$ 6.6 & 51.0 $\pm$ 7.0 \\
Qwen/Qwen2.5-7B-Instruct & 37.3 $\pm$ 1.1 & 38.2 $\pm$ 1.5 & 38.2 $\pm$ 0.9 & 38.8 $\pm$ 1.1 & 37.8 $\pm$ 0.8 & 38.4 $\pm$ 1.2 & 37.2 $\pm$ 1.6 & 39.1 $\pm$ 0.7 \\
Qwen/Qwen2-7B-Instruct & 36.4 $\pm$ 1.1 & 32.6 $\pm$ 1.5 & 32.4 $\pm$ 0.9 & 34.9 $\pm$ 1.1 & 34.8 $\pm$ 0.8 & 34.1 $\pm$ 1.2 & 34.1 $\pm$ 1.5 & 35.6 $\pm$ 0.7 \\
Qwen/Qwen3-8B & 32.3 $\pm$ 1.1 & 33.4 $\pm$ 1.5 & 30.4 $\pm$ 0.9 & 33.0 $\pm$ 1.0 & 33.0 $\pm$ 0.8 & 33.2 $\pm$ 1.2 & 32.3 $\pm$ 1.5 & 31.9 $\pm$ 0.7 \\
Qwen/Qwen2.5-7B & 31.1 $\pm$ 1.1 & 34.8 $\pm$ 1.5 & 32.4 $\pm$ 0.9 & 34.9 $\pm$ 1.1 & 33.4 $\pm$ 0.8 & 32.3 $\pm$ 1.2 & 33.3 $\pm$ 1.5 & 33.9 $\pm$ 0.7 \\
meta-llama/Llama-3.1-8B-Instruct & 29.8 $\pm$ 1.0 & 27.8 $\pm$ 1.4 & 28.0 $\pm$ 0.9 & 29.5 $\pm$ 1.0 & 30.3 $\pm$ 0.8 & 31.6 $\pm$ 1.2 & 29.8 $\pm$ 1.5 & 28.2 $\pm$ 0.7 \\
Qwen/Qwen3-4B & 27.4 $\pm$ 1.0 & 27.5 $\pm$ 1.4 & 26.9 $\pm$ 0.8 & 26.7 $\pm$ 1.0 & 28.2 $\pm$ 0.8 & 29.8 $\pm$ 1.2 & 29.2 $\pm$ 1.5 & 27.7 $\pm$ 0.7 \\
mistralai/Mistral-7B-Instruct-v0.3 & 24.9 $\pm$ 1.0 & 26.1 $\pm$ 1.4 & 26.0 $\pm$ 0.8 & 25.0 $\pm$ 1.0 & 26.5 $\pm$ 0.8 & 27.0 $\pm$ 1.1 & 25.6 $\pm$ 1.4 & 25.4 $\pm$ 0.7 \\
mistralai/Mistral-7B-Instruct-v0.1 & 23.7 $\pm$ 1.0 & 23.9 $\pm$ 1.3 & 26.1 $\pm$ 0.8 & 23.2 $\pm$ 0.9 & 25.9 $\pm$ 0.8 & 25.7 $\pm$ 1.1 & 23.7 $\pm$ 1.4 & 25.6 $\pm$ 0.7 \\
tiiuae/Falcon3-10B-Instruct & 36.9 $\pm$ 1.1 & 37.7 $\pm$ 1.5 & 33.8 $\pm$ 0.9 & 36.9 $\pm$ 1.1 & 34.9 $\pm$ 0.8 & 36.1 $\pm$ 1.2 & 36.7 $\pm$ 1.5 & 37.4 $\pm$ 0.7 \\
mistralai/Mistral-7B-Instruct-v0.2 & 24.2 $\pm$ 1.0 & 25.1 $\pm$ 1.4 & 23.3 $\pm$ 0.9 & 22.0 $\pm$ 1.0 & 25.1 $\pm$ 0.8 & 22.2 $\pm$ 1.1 & 23.1 $\pm$ 1.4 & 23.8 $\pm$ 0.7 \\
meta-llama/Llama-3.2-3B & 21.5 $\pm$ 0.9 & 22.6 $\pm$ 1.3 & 22.9 $\pm$ 0.8 & 21.0 $\pm$ 0.9 & 24.3 $\pm$ 0.7 & 23.8 $\pm$ 1.1 & 22.9 $\pm$ 1.3 & 22.3 $\pm$ 0.6 \\
meta-llama/Llama-3.2-3B-Instruct & 26.3 $\pm$ 1.0 & 24.7 $\pm$ 1.3 & 22.7 $\pm$ 0.8 & 23.5 $\pm$ 0.9 & 24.1 $\pm$ 0.7 & 25.1 $\pm$ 1.1 & 23.3 $\pm$ 1.4 & 23.9 $\pm$ 0.6 \\
meta-llama/Llama-3.2-1B & 24.5 $\pm$ 1.0 & 23.9 $\pm$ 1.3 & 25.6 $\pm$ 0.8 & 23.7 $\pm$ 0.9 & 25.9 $\pm$ 0.8 & 25.8 $\pm$ 1.1 & 23.5 $\pm$ 1.4 & 26.3 $\pm$ 0.7 \\
meta-llama/Llama-3.1-8B & 23.8 $\pm$ 1.0 & 25.7 $\pm$ 1.4 & 24.4 $\pm$ 0.8 & 22.7 $\pm$ 0.9 & 24.6 $\pm$ 0.7 & 24.3 $\pm$ 1.1 & 25.5 $\pm$ 1.4 & 22.0 $\pm$ 0.6 \\
Qwen/Qwen2.5-Math-7B & 22.6 $\pm$ 1.0 & 21.9 $\pm$ 1.3 & 23.7 $\pm$ 0.8 & 23.1 $\pm$ 1.0 & 23.6 $\pm$ 0.8 & 22.9 $\pm$ 1.1 & 22.9 $\pm$ 1.4 & 24.0 $\pm$ 0.7 \\
microsoft/Phi-4-reasoning-plus & 29.5 $\pm$ 1.1 & 30.8 $\pm$ 1.5 & 28.4 $\pm$ 0.9 & 31.3 $\pm$ 1.1 & 31.9 $\pm$ 0.9 & 32.2 $\pm$ 1.3 & 30.6 $\pm$ 1.5 & 31.1 $\pm$ 0.7 \\
meta-llama/Llama-3.2-1B-Instruct & 24.3 $\pm$ 1.0 & 23.8 $\pm$ 1.3 & 24.7 $\pm$ 0.8 & 22.6 $\pm$ 0.9 & 24.4 $\pm$ 0.7 & 26.0 $\pm$ 1.1 & 22.7 $\pm$ 1.3 & 24.4 $\pm$ 0.6 \\
microsoft/Phi-4-reasoning & 5.2 $\pm$ 1.6 & 7.9 $\pm$ 2.4 & 3.4 $\pm$ 0.7 & 7.4 $\pm$ 1.4 & 7.0 $\pm$ 1.2 & 6.5 $\pm$ 1.7 & 9.4 $\pm$ 2.5 & 4.2 $\pm$ 0.9 \\
Qwen/Qwen2.5-Math-7B-Instruct & 14.3 $\pm$ 0.8 & 13.9 $\pm$ 1.1 & 10.8 $\pm$ 0.6 & 13.1 $\pm$ 0.8 & 14.5 $\pm$ 0.6 & 16.0 $\pm$ 0.9 & 16.1 $\pm$ 1.2 & 14.4 $\pm$ 0.5 \\
deepseek/deepseek-v3.2-speciale & 0.0 $\pm$ 0.0 & 0.0 $\pm$ 0.0 & 0.0 $\pm$ 0.0 & 0.0 $\pm$ 0.0 & 0.0 $\pm$ 0.0 & 0.0 $\pm$ 0.0 & 0.0 $\pm$ 0.0 & 0.0 $\pm$ 0.0 \\
gemini-3-flash-preview & -- & -- & -- & -- & -- & -- & -- & -- \\
gemini-3.1-pro-preview & -- & -- & -- & -- & -- & -- & -- & -- \\
\Xhline{1.5pt}
\end{tabular}%
}
\caption{Detailed evaluation results of current popular models on ArxivRollBench2026a for prediction tasks. Scores are percentages and use the same valid-response scoring protocol as the overall ranking.}
\label{tab:appendix-2026a-p}
\end{table*}

\section{Supplemental Related Work}
\label{sec:rel}

\noindent
\textbf{Overestimation of LLMs}
The overestimation observed during
the evaluation of large language models (LLMs) typically stem from two
primary sources: \emph{contamination} of test samples and \emph{biased
  overtraining} on the evaluated tasks.

Contamination in LLMs, as referenced in prior research~\cite{contamination1,contamination2,contamination3,contamination4,contamination5}, occurs
when samples in the test dataset have already been included in the
pre-training or fine-tuning dataset for a specific LLM. Consequently,
the model can achieve superior performance on such a test set not only
through task generalization, but also by memorizing the samples. This
leads to an unfair and biased comparison among other uncontaminated
LLMs, and raises doubts about the actual capabilities of LLMs
on specific tasks. Previous studies have demonstrated that test
set contamination is a widespread phenomenon in LLMs~\cite{contamination2,contamination3}.
Furthermore, some methods~\cite{attack-contamination} deliberately induce \emph{intentional}
contamination to manipulate the benchmark results.

While contamination focuses on cheating at the sample level, another
category known as biased overtraining targets the task level to
deceive the evaluation. Specifically, when trainers possess prior
knowledge about the domains in which their models will be evaluated,
they may strategically enhance the performance of their LLMs on these
specific domains during pre-training, while neglecting other
domains. By doing so, they can manipulate the results of a limited and
task-sampled benchmark using their biased and over-trained
LLMs. Unlike contamination, biased overtraining has not garnered
sufficient attention.
% In our paper, we aim to shed light on this
% phenomenon and reveal its implications.

\noindent
\textbf{Robust LLM Evaluation}. To ensure a fair comparison among LLMs
and minimize the impact of test sample memorization on a given task,
various hand-crafted benchmarks and leaderboards have been
proposed. Examples include MMLU-Pro~\cite{mmlupro}, SCI-Bench~\cite{scibench}, and MMMU~\cite{mmmu},
which are designed to re-rank the performance of LLMs.

Inspired by these efforts, \emph{private benchmarks} utilizing trusted
third-party platforms have emerged as a potential solution to
prevent overestimation. However, the assumption of a fully trusted
third-party platform is often unrealistic, and the transparency and
reproducibility of the evaluation process cannot be
guaranteed. Consequently, adversaries may attempt to bribe the
platform to improve their ranking or leak test cases for cheating
without facing penalties.

Another approach to robust evaluation is the \emph{1-versus-1 arena}~\cite{arena,olympicarena,arena-hard,arena-hard-auto},
such as Chatbot Arena~\cite{arena}. In this setup, given the same
instruction from a user, the system randomly selects two models, A and
B, to answer the question. The user then provides feedback on which
model is better. With an infinite number of duels, the ranking of LLMs
stabilizes under an elo-based mechanism~\cite{elo}. While this method is
effective, it requires a significant amount of evaluation among models
and lacks interpretability in terms of why model A is better than
model B, both in terms of transparency and
reproducibility. Additionally, this leaderboard may be susceptible to
malicious annotators who could intentionally provide incorrect
feedback.

The third type of robust evaluation focuses on \emph{symbolic
  formatting}~\cite{gsm-symbolic,math-symbolic,livebench,dyval,darg,dempa}. Specifically, for certain tasks, we can design
numerous \emph{templates} with placeholders. By flexibly combining
these templates and filling in the entities, we can generate an
infinite number of test samples. However, this method is only suitable
for specific tasks, such as mathematical reasoning, and may be
challenging to apply to others, such as commonsense QA and
translation.

In summary, while various robust evaluation methods have been proposed
to address the challenges of evaluating LLMs, each has its own
limitations. Therefore, it is crucial to continue exploring new and
innovative approaches to ensure a fair, transparent, and reproducible
evaluation of LLMs.

%%% Local Variables:
%%% mode: latex
%%% TeX-master: "main"
%%% End::

% \begin{table*}[t]
% \centering
% \begin{tabular}{ll}
% \hline
% Type         &{URL}                                                                                                                         \\
% \hline
% Source Code       & \url{https://anonymous.4open.science/r/RoBench/}\\
% Datasets & \url{https://huggingface.co/liangzid?search_datasets=robench2024b}\\

% \hline                                                    
% \end{tabular}
% \caption{Resources of the paper.}\label{tab:resource}
% \end{table*}

%%% Local Variables:
%%% mode: latex
%%% TeX-master: "main"
%%% End:

\end{document}